\documentclass[twoside]{article}
\usepackage[accepted]{aistats2025}
\usepackage{lmodern}  
\usepackage[T1]{fontenc}  
\usepackage{amsmath, amsthm, amscd, amsfonts, amssymb, graphicx, multirow, centernot, dsfont, pifont}

\usepackage{titlesec}
\usepackage{listings}
\usepackage{color}

\usepackage{appendix}
\usepackage{tcolorbox}
\usepackage{booktabs}
\usepackage{graphicx}
\usepackage{hyperref}
\usepackage{bookmark}
\usepackage{algorithm}
\usepackage{algpseudocode}
\usepackage{stmaryrd}
\usepackage{xltabular}
\usepackage{ltablex}
\usepackage{longtable}
\usepackage{forest}

\usepackage[capitalize,noabbrev]{cleveref}

\definecolor{my_link}{rgb}{0.64,0.16,0.16}
\usepackage{hyperref}
\hypersetup{
    linkcolor=my_link,citecolor=my_link,colorlinks=true,urlcolor=my_link,%
}
\usepackage{url}
\newtheorem*{property*}{Property}
\newtheorem*{theorem*}{Theorem}
\newtheorem{condition}{Condition}

\newtheorem{definition}{Definition}
\newcommand{\Xprime}{{X'}}
\usepackage{subcaption}

\newcommand{\V}{\mathbb{V}}
\newcommand{\E}{\mathbb{E}}

\newcommand{\cov}[2]{\text{Cov}\left(#1, #2\right)}
\newcommand{\aCov}[2]{\text{Cov}^{\infty}\left(#1, #2\right)}
\newcommand{\Pb}{\mathbb{P}}
\newcommand{\aVar}{\mathbb{V}^{\infty}}
\newcommand{\indep}{\perp \!\!\! \perp}
\DeclareMathOperator*{\argmin}{arg\,min} 

\newtheorem{proposition}{Proposition}
\newtheorem{theorem}{Theorem}

\newtheorem{remark}{Remark}

\tcolorboxenvironment{defs}{left=1mm, sharp corners, colback={green}}

\usetikzlibrary{positioning, shapes.geometric, shadows, quotes, arrows}

\usepackage[round]{natbib}

\begin{document}
\runningtitle{Multi-Study ATE Estimation beyond Meta-Analysis}
\twocolumn[

\aistatstitle{Federated Causal Inference:\\Multi-Study ATE Estimation beyond Meta-Analysis}

\aistatsauthor{ R\'emi Khellaf \And Aur\'elien Bellet \And  Julie Josse }

\aistatsaddress{Inria, PreMeDICaL Team, Université de Montpellier} ]
\begin{abstract}
    We study Federated Causal Inference, an approach to estimate treatment effects from decentralized data. We compare three classes of Average Treatment Effect (ATE) estimators derived from the Plug-in G-Formula, ranging from simple meta-analysis to one-shot and multi-shot federated learning, the latter leveraging the full data to learn the outcome model (albeit requiring more communication). Focusing on Randomized Controlled Trial studies (RCTs), we derive the asymptotic variance of these estimators for linear models. Our results provide practical guidance on selecting the appropriate estimator for various scenarios, including heterogeneity in sample sizes, covariate distributions, treatment assignment schemes, and study-effects. We validate these findings through experiments on simulated and semi-synthetic data.
\end{abstract}

\section{INTRODUCTION}
\looseness=-1 In modern evidence-based medicine, Randomized Controlled Trials (RCT) are considered the gold standard for estimating the Average Treatment Effect (ATE) because they effectively isolate the treatment effect from confounding factors \citep{Cook2014UserGuide}. The most widely used estimator of the ATE, when expressed as a risk difference, is the difference-in-means (DM) estimator.
Recently, however, the \citet{FDA2023} has recommended to adjust for covariates using linear models for the outcome, as this approach consistently yields more precise ATE estimates than the DM estimator \citep{EMA2024, tsiatis2008covariate, benkeser2021improving} even when the assumption of linearity does not hold~\citep{lin2013agnostic, wager2020stats, lei2021regression, van2024covariate}.

Nevertheless, concerns have been raised about the limited scope of RCTs, including their stringent eligibility criteria, short timeframes,
limited sample size, etc. 
Consequently, regulatory agencies tasked with making high-stakes decisions on drug approvals---decisions that directly impact public health and for which the reimbursement of the drug is often tied to its efficacy \citep{SanteGouv}---frequently turn to meta-analysis to guide their choices. Meta-analysis, which aggregates estimated effects from multiple studies conducted across various studies \citep{hunter2004methods, borenstein2021introduction}, represents the pinnacle of evidence in clinical research \citep{blunt2015hierarchies}.
They can lead to increased statistical power and more precise estimates, while also offering valuable insights into rare adverse events.

\looseness=-1 Despite extensive guidelines on conducting meta-analyses \citep{moher1999improving, liberati2009g0tzsche, higgins2019cochrane}, multi-study 
approaches still face significant challenges. These primarily arise from heterogeneity caused by imbalances in datasets, variations in populations across studies, and study-effects on the outcome due to differing practices across studies
\citep{berlin2014meta}.
Moreover, simply aggregating local estimates is not the only approach to conducting meta-analyses. However, implementing ``one-stage'' meta-analyses \citep{morris2018meta} that pool individual patient data from all studies is practically challenging due to data silos and personal data regulations.

\looseness=-1 \emph{Federated causal inference}, an emerging field combining federated learning \citep{kairouz2021advances} and causal inference \citep{imbens2015causal, hernan2020book} to estimate causal effects from decentralized data sources, offers a compelling alternative to traditional meta-analysis. Federated Learning (FL) enables multiple studies to collaboratively train a model without sharing raw individual data, instead exchanging only model updates that are iteratively aggregated by an orchestrating server. This decentralized approach is especially valuable in fields like medicine \citep{sheller2020federated,che2022federated, prosperi2020causal}, where strong incentives exist to keep data on-site---whether to comply with data protection regulations \citep{koga2024differentially}, maintain ownership and control over the data, or avoid unwanted knowledge transfer. However, traditional FL algorithms are designed to learn predictive models \citep{kairouz2021advances}, rather than estimating causal effects.


\looseness=-1 \textbf{Contributions.} In this paper, we aim to estimate the ATE for a population represented by multiple RCT 
studies conducted over potentially heterogeneous populations,
using a federated approach. We study and compare three classes of federated estimators of the ATE:
(i) \emph{meta-analysis estimators}, which aggregate ATE estimates computed independently at each study; (ii) \emph{one-shot federated estimators}, where outcome model parameters are estimated at each study, aggregated, and shared back for studies to compute and aggregate ATE estimates; and (iii) \emph{gradient-based federated estimators}, where outcome model parameters are learned on joint data using federated gradient descent before each study computes and aggregates its ATE estimate. These estimators entail different communication costs, which often act as the bottleneck in real-world systems \citep{kairouz2021advances}.

\looseness=-1 Our primary contribution is the derivation of asymptotic variances for these estimators under a linear outcome model. This modeling choice is known to provide a variance reduction compared to the classic DM estimator, even when the underlying model is not linear \citep{lin2013agnostic, wager2020stats, lei2021regression, van2024covariate}, and aligns with recent recommendations from regulatory agencies \citep{FDA2023}. We specifically address scenarios involving heterogeneity, including distributional shifts (varying covariate distributions across studies) and study-effects on the outcome.  
Our results shed light on the trade-offs between the statistical efficiency, communication costs, and underlying modeling assumptions  of the considered estimators. We find that, despite their simplicity, low communication overhead and minimal assumptions, meta-analysis estimators can achieve statistical efficiency comparable to pooled data analysis when sufficient data is available at each study, while naturally accommodating study-effects. In contrast, when local datasets are small, gradient-based federated estimators stand out as the only viable option. One-shot estimators offer an interesting middle ground in some cases: they can recover the same ATE estimate as pooled data analysis while being robust to distributional shifts in covariates and differences in treatment assignments, but suffer from increased variance when study-effects are present. These conclusions are supported by experiments on (i) simulated data, illustrating the behavior of the estimators under the different scenarios; and (ii) semi-synthetic data, where we use the real-world Traumatrix database \citep{mayerdoubly} with synthetic outcomes. Ultimately, our work provides clear guidelines on selecting the most suitable estimator for different scenarios, as summarized by a decision diagram designed for practitioners (Figure~\ref{fig:diagramdec} in Appendix~\ref{app:diagram}).

\textbf{Related work.}
The work closest to ours is \citet{xiong2021federated}, which adapts estimators of the ATE 
through a one-shot federated estimation of the outcome/propensity score model parameters. However, their work does not compare the efficiency of these one-shot federated estimators with traditional meta-analysis and pooled dataset estimators, nor does it consider gradient-based federated alternatives. Our results offer clear guidelines on when the One-Shot estimators proposed by \citet{xiong2021federated} should be preferred over other methods.

\looseness=-1 Other work on federated causal inference (see \citet{brantner2023methods} and \citet{edmondson2023statistical} for an overview) consider different settings and objectives than the ones considered in our work. \citet{vo2021federated} employ a Bayesian framework using Gaussian processes to estimate the ATE under uniform data distributions across studies. \citet{terrail2023fedeca} focus on federating an external control arm for time-to-event outcomes, adapting a gradient-based algorithm for Cox hazard model parameters. While our work aims to estimate the causal effect of treatment across the joint population of studies, other studies \citep{vo2022adaptive,han2021federated,DBLP:conf/nips/HanSZ23,DBLP:journals/corr/abs-2402-17705,DBLP:journals/corr/abs-2404-15746} aim at transferring causal estimates from a source study to a target population. 

\looseness=-1 The meta-analysis literature on combining estimates from multiple studies is extensive. One can mention the work of \citet{morris2018meta} who discuss the differences and advantages of conducting meta-analysis with individual patient data on stratified (``two-stage'') versus pooled data (``one-stage''). However, they require sharing raw data and do not explore any federated strategy. 
Meta-analysis provides considerable flexibility in the choice of local models \citep{seo2021comparing, tan2022tree}, but also comes with many subtle statistical challenges. These include the ecological fallacy bias \citep{piantadosi1988ecological}, which occurs when incorrect conclusions about individuals are drawn from subgroup characteristics, as well as situations when ignoring study sizes can lead to biased ATE estimates \citep{kahan2023estimands}. 

\section{GENERAL FRAMEWORK} 
\label{sec:framework}


\textbf{Notations.}
We consider a set of $K$ studies, with $H$ denoting the random variable with values in $\{1,\dots,K\}$ indicating membership to a study.
Let $\mathcal{Z}=\{Z_i\}_{i=1}^n$ be a sample of $n$ independent and identically distributed (i.i.d.) realizations of the quadruplet $Z=(X, W, Y, H)$, where $X$ denotes a $d$-dimensional vector of covariates that belongs to a covariate space $\mathcal{X} \subset \mathbb{R}^d$, $W \in \{0, 1\}$ denote the binary treatment, and $Y \in \mathbb{R}$ is the observed outcome of interest. 

We denote by $\mathcal{Z}_k$ the local dataset of study $k$ with $n_k=\sum_{i=1}^n \mathds{1}_{\{H_i=k\}}$ observations, and by $\mathcal{Z}_{k}^{(w)}$ the $n_k^{(w)}=\sum_{i=1}^n \mathds{1}_{\{H_i=k,W_i=w\}}$ observations in study $k$ under treatment arm $w$. We further denote by $X_k^{(w)} = \{X_{i} \mid W_{i} = w, H_i=k\}_{i=1}^n \in\mathbb{R}^{n_k^{(w)}\times d}$ (resp. $Y_k(w) = \{Y_{i} \mid W_{i} = w, H_i=k\}_{i=1}^n \in\mathbb{R}^{n_k^{(w)}}$) the design matrix of the covariates (resp. the outcome vector) for treatment arm $w$ in study $k$. Similarly, we denote by $\mathcal{Z}^{(w)}$ the $n^{(w)}=\sum_{k=1}^n n_k^{(w)}$ observations under treatment arm $w$ in the pooled dataset $\mathcal{Z}$, and by $X^{(w)}\in\mathbb{R}^{n^{(w)}\times d}$  and $Y^{(w)}\in\mathbb{R}^{n^{(w)}}$ the corresponding design matrix and outcome vector.

\looseness=-1 \textbf{Average treatment effect in $K$ RCTs.}
We consider the setting of $K$ Bernoulli RCT trials, where each participant \(i\) in study $k$ has a fixed probability \(\mathbb{P}\left(W_i = 1 \mid H_i=k\right) = p_k\) of being assigned to the treatment group, which does not depend on $X$ in this design. We denote $\rho_k=\Pb(H_i=k)$ the probability that an observation belongs to study $k$ ($0<\rho_k<1$). Note that the probability $p$ of being treated within the pooled dataset $\mathcal{Z}$ is then given by $p=\sum_{k=1}^K \rho_k p_k$.

\looseness=-1 We consider the potential outcomes framework \citep{RubinDonaldB1974Eceo} and we aim to estimate the ATE $\tau\in \mathbb{R}$ defined as the Risk Difference over the $K$ studies as $\tau=\E\left(\E(Y_i(1) - Y_i(0)|H_i)\right)$, where $Y(w)$ is the outcome had the subject received treatment $w$. We denote the local ATE in study $k$ by $\tau_k = \E(Y_i(1) - Y_i(0)|H_i=k)$.

\looseness=-1 We assume that the classical identifiability assumptions for a RCT design hold locally at every study: \label{par:causal_assumptions_and_rct_def}(a) \textit{SUTVA} (Stable Unit Treatment Value Assumption): \(Y = W Y{(1)} + (1-W)Y{(0)} \), (b)
   \textit{Positivity}:  \( \exists \eta_1 >0 \) such that, almost surely, $\eta_1 \le \mathbb{P}(W = 1\mid H)  \le 1-\eta_1$ and \hypertarget{as:unconfoundedness}{(c)} \textit{Ignorability}: \(W\mkern-7mu \indep\mkern-7mu (Y(0), Y(1))\). Under these conditions, 
   the ATE is identifiable and the simple  ``Difference-in-Means'' \citep{Neyman} estimator defined as $\hat{\tau}_\mathrm{DM} = \frac{1}{n^{(1)}} \sum_{i=1}^{n^{(1)}} Y_i \, W_i - \frac{1}{n^{(0)}}\sum_{i=1}^{n^{(0)}} Y_i \,(1\mkern-4mu -\mkern-4mu W_i)$ 
   is an unbiased estimator of the ATE. 
   However, variance reduction can be obtained by adjusting from covariates and considering 
   the ``Plug-in G-formula'' or ``outcome-based regression'' estimator. 

\begin{definition}[\citealp{robins1986new}]\label{def:g-formula}
    \looseness=-1 The plug-in G-formula estimator is defined as
     $\hat{\tau}=\frac{1}{n} \sum_{i=1}^{n}\left(\hat \mu_1(X_i) - \hat \mu_0(X_i)\right)$, 
    where $\hat \mu_{w}(x)$ is an estimator of the surface response $\mu_{w}(x)=\mathbb{E}\left[Y| W=w, X=x\right]$.
\end{definition}

Throughout the paper, we will consider two different regimes for the sample sizes. 
\begin{condition}[Local Full Rank]\label{cond:local_large_sample_size}
    $\forall k \in \{1,\ldots, K \}$ and for $w\in \{0,1\}$, we have $\mathrm{rank}({X_k^{(w)}}^\top {X_k^{(w)}}) = d$, which implies that $\forall (k,w), n_k^{(w)} \geq d$.
\end{condition}

\begin{condition}[Federated Full Rank]\label{cond:federated_large_sample_size}
    For $w \in \{0,1\}$, we have $\mathrm{rank}({X^{(w)}}^\top {X^{(w)}}) = d$, which implies that $\forall w, \sum_{k=1}^K n_k^{(w)} \geq d$.
\end{condition}




\section{HOMOGENEOUS POPULATION}\label{sec:homo}
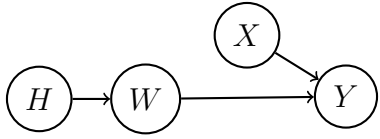
\begin{figure}[t]
    \centering
    \begin{tikzpicture}[node distance={15mm}, thick, main/.style = {draw, circle}]
    \node[main] (1) {$W$};
    \node[main] (2) [above right=2mm and 7 mm of 1] {$X$};
    \node[main] (3) [below right=2mm and 7 mm of 2] {$Y$};
    \node[main] (4) [left =5mm of 1] {$H$};
    \draw[->] (2) -- (3);
    \draw[->] (1) -- (3);
    \draw[->] (4) -- (1);
    \coordinate (midpoint) at ($(1)!0.5!(3)$);
    \end{tikzpicture}
    \caption{Graphical Model for Homogeneous Settings}\label{graph:homog}
\end{figure}

In this section, we focus on estimating the ATE over $K$ RCTs studying the same population. We assume that the joint distribution of $Z=(X,W,Y,H)$ decomposes as
\begin{equation}
\label{eq:decompo_homo}
\Pb(Z)=\Pb(H)\Pb(W|H)\Pb(X)\Pb(Y|X,W).
\end{equation}
This corresponds to the graphical model shown in \cref{graph:homog}.
We refer to this setting as \textit{homogeneous} because $X$ and $Y|X,W$ are independent of $H$: in other words, there is no distributional shift for the covariates and the conditional outcomes  across studies. \textit{Heterogeneous} settings will be addressed in Section~\ref{sec:hetero}. 

We consider a linear model for the potential outcomes
\begin{equation}\label{model:model1}
    Y_{k,i}(w) = c^{(w)} + X_{k,i} \beta^{(w)} + \varepsilon_{k,i} (w),
\end{equation}
with $c^{(w)}\in \mathbb{R}$ the intercept, $\beta^{(w)}\in \mathbb{R}^d$ the coefficients,  $\E(\varepsilon_{k,i}(w) \mid X_{k,i}) =0$ and $\V(\varepsilon_{k,i}(w)\mid X_{k,i}) = \sigma^2$. 

Note that $\beta^{(1)}$ and $\beta^{(0)}$ can be different, so that the treatment effect can be heterogeneous (\textit{i.e.} depends on the covariates).
We denote $\theta^{(w)} = (c^{(w)}, \beta^{(w)}) \in \mathbb{R}^{d+1}$
and $\Xprime = (1, X) \in \mathbb{R}^{n,d+1}$ the covariate matrix augmented with a column of ones. 
The model parameters $\theta^{(w)}$ are equal in every study, in accordance with the homogeneous setting defined by the decomposition in Eq.~\ref{eq:decompo_homo}. We assume that $X$ has finite first two moments $\E(X)=\mu$ and $\V(X)=\Sigma =\E((X-\mu)^\top (X-\mu)$.

Under the above model, the ATE can be written as 
\begin{equation}
\label{eq:tau}
    \tau =\E(X'_i)(\theta^{(1)} - \theta^{(0)})
\end{equation}
and the ATEs per study $\tau_k = \E(X'_{k,i})(\theta^{(1)} - \theta^{(0)})$ are homogeneous across studies, i.e., $\tau_1 = \dots = \tau_K = \tau$, as the covariate distribution is the same across studies.


The G-formula estimator on the pooled data $\mathcal{Z}$ can be used to estimate the ATE over the $K$ studies
\begin{equation}
    \hat{\tau}_{\mathrm{pool}} = \textstyle\frac{1}{n} \sum_{i=1}^{n}\big(X'_i\hat\theta_{\mathrm{pool}}^{(1)} - X'_i\hat\theta_{\mathrm{pool}}^{(0)} \big),
    \label{eq:pool}
\end{equation} 
where $\hat \theta_{\mathrm{pool}}^{(1)}$ and $\hat \theta_{\mathrm{pool}}^{(0)}$ are the regression coefficients estimated  by fitting two OLS regressions over $\mathcal{Z}^{(1)}$ and $\mathcal{Z}^{(0)}$ respectively. This \textit{pooled} estimator satisfies (see \cref{proof:asymp_ols} for a proof extending the standard result in \citet{wager2020stats} to non-centered covariates)
\begin{align}\label{eq:asymp_ols}
    \root \of n (\hat\tau_{\mathrm{pool}} - \tau) \xrightarrow{d} \mathcal{N}(0, V_\mathrm{pool}),
\end{align}
with $V_\mathrm{pool} = \frac{\sigma^2}{p(1-p)} + \Vert \beta^{(1)} - \beta^{(0)} \Vert^2_\Sigma$.
However, computing $\hat{\tau}_\mathrm{pool}$ requires access to the \textit{pooled} dataset $\mathcal{Z}$, which is not accessible in the decentralized setting we consider.
Under Condition~\ref{cond:local_large_sample_size}, each study in isolation can only estimate the ATE using its local dataset $\mathcal{Z}_k$
\begin{equation}
    \hat{\tau}_k = \textstyle\frac{1}{n_k} \sum_{i=1}^{n_k}\big(X'_{k,i}\hat{\theta}_k^{(1)} - X'_{k,i}\hat{\theta}_k^{(0)}\big), 
    \label{eq:tauk}
\end{equation} 
where $\hat\theta_k^{(w)}\mkern-3mu=\mkern-3mu({{{X'_k}^{(w)}}}^\top {{X'_k}^{(w)}})^{-1} {{{X'_k}^{(w)}}}^\top Y_k^{(w)}$ is the OLS estimator computed over $\mathcal{Z}_k^{(w)}$. 
To improve upon this baseline, we now introduce several estimation strategies of the ATE over the $K$ studies, with the aim of obtaining estimates as if one had access to $\mathcal{Z}$. 



\subsection{Definition of the Estimators}
\label{sec:estimators}

\subsubsection{Meta-Analysis Estimators}\label{subsec:meta}

A first strategy under \cref{cond:local_large_sample_size} is to aggregate the local ATE estimates $\hat{\tau}_k$ in Eq.~\ref{eq:tauk}.
The studies then send their local ATE estimates to the server, which aggregates them using non-negative weights $\omega_k$ that sum to 1 over the $K$ studies to obtain a global ATE estimate. For selecting the weights $\omega_k$, \citet{hunter2004methods} describes two common methods for absolute measures like the risk difference: sample size weighting (SW) and inverse variance weighting (IVW). These meta-analysis estimators involve a single round of communication: each study $k$ sends $\hat{\tau}_k$ to the server.


\begin{definition}[Meta-Analysis - SW Aggregation]\label{def:meta_sw}
\begin{equation}
\label{eq:meta_sw}
   \textstyle \hat{\tau}_{\mathrm{Meta\text{-}SW}} = \sum_{k=1}^{K} \frac{n_k}{n} \hat{\tau}_k.
\end{equation}
\end{definition}

\begin{definition}[Meta-Analysis - IVW Aggregation]\label{def:meta_ivw}
\begin{equation}
    \label{eq:meta_ivw}
    \textstyle \hat{\tau}_{\mathrm{Meta\text{-}IVW}} = \frac{\sum_{k=1}^{K}\V(\hat{\tau}_k)^{-1} \hat{\tau}_k}{\sum_{k=1}^{K}\V(\hat{\tau}_k)^{-1}}.
\end{equation}
\end{definition}

\begin{proposition}[proof in \cref{proof:meta_ivw_min_var}]\label{prop:meta_ivw_min_var}
    $\hat{\tau}_{\mathrm{Meta\text{-}IVW}}$ is the minimum-variance estimator of $\tau$ among the class of aggregation-based estimators.
\end{proposition}
\begin{remark}\label{rmk:meta_ivw}
In practice, $\V(\hat{\tau}_k)$ is often unknown and must be estimated, leading to an approximation of $\hat{\tau}_{\mathrm{Meta\text{-}IVW}}$. In contrast, $\hat{\tau}_{\mathrm{Meta\text{-}SW}}$ only requires knowledge of the local sample sizes.
\end{remark}


\subsubsection{One-Shot Federated Estimators}\label{sec:one_shot_def}

To go beyond the mere aggregation of local ATEs,
we can follow \citet{xiong2021federated} and aggregate the local outcome model parameters using a single round of communication (hence the term ``one-shot'' federated) to build better local ATE estimates, before aggregating them.


\textbf{Step 1. Local estimation of outcome parameters:} Under Condition~\ref{cond:local_large_sample_size}, each study $k$ estimates $\hat\theta_k^{(w)}$ locally with an OLS regression.

\textbf{Step 2. One-shot federation of parameters:}\label{subsec:one_shot_fed} We perform a meta-analysis (local estimation then weighted aggregation) of the local outcome model parameters $\hat{\theta}_k^{(w)}$. Specifically, the studies send their local estimates to the server, which computes $\hat\theta^{(w)}_\mathrm{1S} = \sum_{k=1}^K \omega_k^{(\theta)}\hat\theta^{(w)}_k$ (where ‘‘1S'' stands for ‘‘one-shot''), with $\omega_k^{(\theta)}$ some federation weights (summing to 1 over the $K$ studies) like SW or IVW. The server sends back the obtained $\hat\theta^{(w)}_\mathrm{1S}$  to all the studies. 

\begin{definition}[SW Federation of $\hat\theta_k^{(w)}$]
    \begin{align}
    \label{eq:theta_1s_sw}
        \textstyle\hat\theta_\mathrm{1S\text{-}SW}^{(w)} = \sum_{k=1}^K \frac{n_k^{(w)}}{n^{(w)}} \hat\theta_k^{(w)}.
    \end{align}
\end{definition}

\begin{definition}[IVW Federation of $\hat\theta_k^{(w)}$]
    \begin{align}
    \label{eq:theta_1s_ivw}
        \textstyle\hat{\theta}_{\mathrm{1S\text{-}IVW}}^{(w)} = V^{-1} \sum_{k=1}^K \big(\V(\hat{{\theta}}^{(w)}_k)^{-1} \hat{\theta}_k^{(w)}\big).
    \end{align}
    where $V=\sum_{k=1}^K \V(\hat{{\theta}}^{(w)}_k)^{-1} =\sum_{k=1}^K \frac{1}{\sigma^2}{{{X'_k}^{(w)}}}^\top {{{X'_k}^{(w)}}}$.
\end{definition}

\begin{theorem}[proof in \cref{proof:min_var_ivw_thetas}]\label{prop:1S_ivw_thetas_equal_pool}
    Under \cref{cond:local_large_sample_size}, ${\hat\theta_{\mathrm{1S\text{-}IVW}}^{(w)} = \hat\theta_{\mathrm{pool}}^{(w)}}$.
\end{theorem}


Remarkably, \cref{prop:1S_ivw_thetas_equal_pool} shows that one can obtain similar estimates as   $\hat\theta_\mathrm{pool}$ (which has the lowest variance among the class of linear unbiased estimators \citep{giraud2012high}) by federating the local estimates with a one-shot IVW procedure, even in finite sample sizes, whenever the local datasets are of full rank. This gives a very strong argument in favor of this approach in comparison to the One-Shot SW which thus necessarily has higher variance than $\hat\theta_\mathrm{1S\text{-}IVW}$ and $\hat\theta_\mathrm{pool}$.
However, note that the communication cost of computing  $\hat\theta_{\mathrm{1S\text{-}IVW}}^{(w)}$ is $O(d)$ times larger than for $\hat\theta_\mathrm{1S\text{-}SW}^{(w)}$, as each study must send to the server its $(d+1)\times (d+1)$ local variance matrix ${{{X'_k}^{(w)}}}^\top {{{X'_k}^{(w)}}}$.



\textbf{Step 3. Aggregation of the ATEs:}\label{subsec:ate_agg}
Each study estimates its local ATE using the federated outcome model parameters:  
\begin{equation}
    \label{eq:theta_1S}
    \textstyle\hat\tau_k^\mathrm{1S-agg} = \frac{1}{n_k} \sum_{i=1}^{n_k} (X'_{k,i}\hat\theta^{(1)}_\mathrm{1S-agg} - X'_{k,i}\hat\theta^{(0)}_\mathrm{1S-agg})
\end{equation}

with $\mathrm{agg} \in\{\text{SW, IVW}\}$. Finally, a second communication round is used where studies each send their $\hat\tau_k^\mathrm{1S-agg}$ to the server for aggregation with weights $\omega^{(\tau)}$:
\begin{align*}
\hat\tau_\mathrm{1S-agg} &= \textstyle\sum_{k=1}^K \omega_k^{(\tau)} \hat\tau_k^\mathrm{1S-agg}.
\end{align*}
It turns out that using SW or IVW for $\omega^{(\tau)}$ is asymptotically equivalent (see Appendix~\ref{proof:sw_ivw_same_weights}), so in the following we focus on sample size aggregation weights.\label{prop:sw_ivw_same_weights}

\begin{definition}[1S SW Federation - SW Aggregation]\label{def:1SSSSS}
\begin{equation}\label{eq:1SSSSS}
\textstyle\hat{\tau}_\mathrm{1S\text{-}SW} = \sum_{k=1}^{K} \frac{n_k}{n} \hat{\tau}_k^{\mathrm{1S\text{-}SW}}
\end{equation}
\end{definition}
 
\begin{definition}[1S IVW Federation - SW Aggregation]\label{def:1SIVWSS}
    \begin{equation}
    \label{eq:1SIVWSS}
\textstyle\hat{\tau}_\mathrm{1S\text{-}IVW} =  \sum_{k=1}^{K} \frac{n_k}{n} \hat{\tau}_k^{\mathrm{1S\text{-}IVW}} 
    \end{equation}
\end{definition}

\subsubsection{Gradient-based Federated Estimators}
\label{sec:fed_est}

Neither the meta nor the one-shot estimators can be used when \cref{cond:local_large_sample_size} does not hold (e.g., as soon as one study has its sample size $n_k^{(w)} < d$), since $\hat\theta_k^{(w)}$ is not defined. In such cases, we propose to leverage gradient-based federated estimators.

\textbf{Step 1. Multi-shot federation of parameters:} Studies jointly estimate $\hat\theta_\mathrm{pool}^{(w)}$ by solving the underlying OLS problem in a federated fashion. 
To the best of our knowledge, we are the first to propose this approach to estimate the ATE in Federated Causal Inference.
Finding $\hat\theta_\mathrm{pool}^{(w)}$ amounts to minimizing the Mean Squared Error loss function, defined as $\ell(\theta^{(w)}, \bigcup_{k=1}^K \mathcal{Z}_k^{(w)}) \mkern-7mu=\mkern-7mu \frac{1}{n^{(w)}} \sum_{i=1}^{n^{(w)}} (Y_i^{(w)}\mkern-5mu -\mkern-5mu {X'}_i^{(w)} \theta^{(w)})^2$. This optimization problem can be solved by a federated gradient descent-based algorithm. We propose to use the FedAvg algorithm \citep{mcmahan2017communication} for its intuitive simplicity, strong convergence guarantees, and good empirical performance on both homogeneous \citep{stich_localsgd,DBLP:conf/aistats/0001MR20} and heterogeneous data \citep{fedavggoodhetero}.
FedAvg alternates for $T$ rounds between performing $E$ local gradient steps in each study and aggregating the parameters at the server, see \cref{app:choice_of_T} for the detailed algorithm. The output of this procedure is an estimate $\hat\theta_\mathrm{GD}^{(w)}$ (“GD” stands for Gradient Descent).


\looseness=-1 \label{par:eta_gd} Let $\lambda_{\mathrm{max},k}$ be the largest eigenvalue of the covariance matrix in study $k$. If we set $T=1$ (a single communication round) and the local learning rate for study $k$ to $2/\lambda_{\mathrm{max},k}$, then $\hat\theta_\mathrm{GD}^{(w)}$ is guaranteed to converge to the one-shot estimate as $\hat\theta_\mathrm{SW}^{(w)}$ as the number of local steps $E\rightarrow\infty$. Conversely, if we set the number of local steps $E=1$ and the global learning rate to $\frac{2}{\sum_{k=1}^K\lambda_{\mathrm{max},k}}$, then $\hat\theta_\mathrm{GD}^{(w)}$ is guaranteed to converge to the pooled estimate $\theta^{(w)}_\mathrm{pool}$ as $T\rightarrow\infty$.
More details are provided in \cref{app:choice_of_learning_rate}. FedAvg thus allows to learn a better estimate of the outcome model parameters than one-shot approaches (in particular when Condition~\ref{cond:local_large_sample_size} does not hold) at the cost of more communication. In our homogeneous case, an extremely accurate estimate of $\hat\theta_\mathrm{pool}^{(w)}$ can be obtained with a small number of communication rounds $T$ (see \cref{app:choice_of_T}).
Then, the estimation error on the ATE with $\hat\theta^{(w)}_\mathrm{GD}$ compared to  $\hat\theta^{(w)}_\mathrm{pool}$ satisfies $(\hat\tau_{\mathrm{pool}} - \hat\tau_\mathrm{GD})^2 \leq \Vert\frac{2}{n}\sum_{i=1}^n{X_i'}\Vert^2 (\epsilon^{(1)} + \epsilon^{(0)})$, with 
$\epsilon^{(w)}=\Vert \hat\theta_{\mathrm{pool}}^{(w)} - \hat\theta_\mathrm{GD}^{(w)}\Vert^2_2$ the error on the parameters.

In the following, we consider that the parameters of FedAvg
are chosen such that $\hat\theta_\mathrm{GD}^{(w)} = \hat\theta_\mathrm{pool}^{(w)}$.
\textbf{Step 2. Aggregation of the ATEs:} 
Each study $k$ then computes a local estimate of the ATE using $\hat\theta_\mathrm{GD}^{(w)}$
\begin{equation}
    \label{eq:tauGDk}
    \textstyle\hat{\tau}_k^{\mathrm{GD}}=\frac{1}{n_k}\sum_{i=1}^{n_k}\big({X'}_{k,i}\hat\theta_\mathrm{GD}^{(1)} - {X'}_{k,i}\hat\theta_\mathrm{GD}^{(0)}\big)
\end{equation}
 Finally, these estimates are aggregated in a last round of communication with sample weighting
 (IVW is asymptotically equivalent, see \cref{sec:one_shot_def}), which yields the following GD estimator of the global ATE.

\begin{definition}[GD Federation - SW Aggregation]\label{def:GDSW}
    \begin{equation}\label{eq:GDSW}
\textstyle\hat{\tau}_\mathrm{GD} = \sum_{k=1}^{K} \frac{n_k}{n} \hat{\tau}_k^{\mathrm{GD}} 
    \end{equation}
\end{definition}




\subsection{Comparison of the Federated Estimators}\label{subsec:additional_prop}

\begin{table*}[!ht]
    \caption{Properties of the (unbiased) estimators of the ATE in the homogeneous setting: asymptotic variance, number of communication rounds and total communication cost (in number of floats per study).}
    \label{tab:taus_var}
    \centering
    \footnotesize
        \begin{tabular}{l l l l l l}
        \toprule
       Estimator & Notation & Condition & $\aVar$ &  Com. rounds & Com. cost\\
        \midrule
        \addlinespace[0.3em]
        \text{Local} & $\hat{\tau}_k$ (Eq.~\ref{eq:tauk})& Cond.~\ref{cond:local_large_sample_size} & $\frac{\sigma^2}{n_k} \frac{1}{p_k(1-p_k)} + \frac{1}{n_k}\Vert \beta^{(1)} - \beta^{(0)} \Vert^2_\Sigma$ & 0 & 0\\

        \addlinespace[0.1em]
        \midrule 
        \addlinespace[0.3em]

        \text{Meta\text{-}SW} & $\hat{\tau}_{\text{Meta\text{-}SW}}$ (Eq.~\ref{eq:meta_sw})& Cond.~\ref{cond:local_large_sample_size}  & {\tiny $\frac{\sigma^2}{n} \displaystyle\sum_{k=1}^{K} \frac{\rho_k}{p_k(1-p_k)}\! +\! \frac{1}{n} \Vert \beta^{(1)} \!\!-\! \beta^{(0)} \Vert^2_\Sigma$}    & 1 & $O(1)$
        \\
        \addlinespace[0.1em]
        \text{Meta\text{-}IVW} & $\hat{\tau}_{\text{Meta\text{-}IVW}}$ (Eq.~\ref{eq:meta_ivw})& Cond.~\ref{cond:local_large_sample_size}  & {\tiny $\Big(\displaystyle\sum_{k=1}^{K} \bigl(\sigma^2\frac{n \rho_k}{p_k(1-p_k)} \!+\! \frac{1}{n_k}\Vert \beta^{(1)}\!\! -\! \beta^{(0)} \Vert^2_\Sigma\bigr)^{-1}\Big)^{-1}$} & 1 & $O(1)$ \\
        \addlinespace[0.1em]

        \text{1S\text{-}SW} & $\hat{\tau}_{\text{1S\text{-}SW}}$ (Eq.~\ref{eq:1SSSSS})& Cond.~\ref{cond:local_large_sample_size}
        & $\frac{\sigma^2}{n} \frac{1}{p(1-p)} + \frac{1}{n}\Vert \beta^{(1)} - \beta^{(0)} \Vert^2_\Sigma$ & 2 & $O(d)$\\
        \addlinespace[0.5em]

        \text{1S\text{-}IVW} & $\hat{\tau}_{\text{1S\text{-}IVW}}$ (Eq.~\ref{eq:1SIVWSS})& Cond.~\ref{cond:local_large_sample_size}  & $\frac{\sigma^2}{n} \frac{1}{p(1-p)} + \frac{1}{n}\Vert \beta^{(1)} - \beta^{(0)} \Vert^2_\Sigma$  & 2 & $O(d^2)$ \\
        \addlinespace[0.5em]
        \text{GD} & $\hat{\tau}_\mathrm{GD}$  (Eq.~\ref{eq:GDSW}) & Cond.~\ref{cond:federated_large_sample_size} & $\frac{\sigma^2}{n} \frac{1}{p(1-p)} + \frac{1}{n}\Vert \beta^{(1)} - \beta^{(0)} \Vert^2_\Sigma$ & $T+1$ & $O(Td)$ \\
        \addlinespace[0.3em]
        \midrule
        \text{Pool} & $\hat{\tau}_\text{pool}$ (Eq.~\ref{eq:pool}) & Cond.~\ref{cond:federated_large_sample_size} & $\frac{\sigma^2}{n} \frac{1}{p(1-p)} + \frac{1}{n}\Vert \beta^{(1)} - \beta^{(0)} \Vert^2_\Sigma$& --- & --- \\
        \bottomrule
        \end{tabular}
\end{table*}
\textbf{General case.} 
Under the graphical model in Figure~\ref{graph:homog} 
and \cref{cond:local_large_sample_size}, all ATE estimators presented so far 
are unbiased (as proved in \cref{proof:no_bias_estimators}).

For each estimator, we report in Table~\ref{tab:taus_var} its asymptotic variance $\aVar$ (with proofs in \cref{proof:avar_ate_random_design_v1}), the sample size required (as per Condition~\ref{cond:local_large_sample_size} or the weaker Condition~\ref{cond:federated_large_sample_size}), the number of communication rounds needed between the studies and the server, and the total communication cost per study (in number of floats). We observe that the one-shot and gradient-based federated estimators achieve the same variance of the pooled-data estimator. The differences lie in the sample size conditions and communication costs. While one-shot estimators require two communication rounds (and generally lower total communication costs), they require that the \emph{local} sample size $n_k^{(w)}$ at each study $k$ and arm $w$ be larger than the dimension $d$. In contrast, the GD estimator requires this only for the \emph{pooled} sample size $n^{(w)}=\sum_{k=1}^K n_k^{(w)}$.

\begin{theorem}[Proof in \cref{proof:comparison_variances}]\label{prop:metas_larger_var_than_pool}
Under graphical model in Fig.~\ref{graph:homog}, the estimators compare as:
    \begin{align*}
        \aVar(\hat{\tau}_\mathrm{pool}) 
        \!=\!\! \aVar(\hat{\tau}_\mathrm{fed}) 
        \!\leq\! \aVar(\hat{\tau}_\mathrm{Meta\text{-}IVW}) 
        \!\leq\! \aVar(\hat{\tau}_\mathrm{Meta\text{-}SW})
    \end{align*}
\end{theorem}

\begin{figure}[t]
    \centering
    \begin{tikzpicture}[node distance={15mm}, thick, main/.style = {draw, circle}]
    \node[main] (1) {$W$};
    \node[main] (2) [above right=2mm and 20 mm of 1] {$X$};
    \node[main] (3) [below right=2mm and 13 mm of 2] {$Y$};
    \node[main] (4) [left =7mm of 2] {$H$};
    \draw[->] (2) -- (3);
    \draw[->] (1) -- (3);
    \draw[->] (4) -- (1);
    \draw[->] (4) -- (2);
    \coordinate (midpoint) at ($(1)!0.5!(3)$);
    \end{tikzpicture}
    \caption{Graphical Model in the Heterogeneous Distributions Setting.} 
    \label{graph:diff_distribs}
\end{figure}
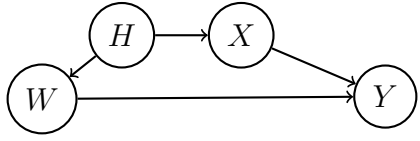

\looseness=-1 where $\text{fed} \in \{\text{GD, 1S\text{-}IVW, 1S\text{-}SW}\}$. \cref{prop:metas_larger_var_than_pool} shows that meta-analysis estimators typically exhibit larger variance. This happens as soon as treatment probabilities are not equal across studies, and the variance difference increases as the treatment probabilities $\{p_k\}_k$ become more distinct. Moreover, $\aVar(\hat{\tau}_\mathrm{Meta\text{-}IVW}) < \aVar(\hat{\tau}_{\mathrm{Meta\text{-}SW}})$ whenever $\{p_k(1-p_k)\}_k$ differ across studies, and the difference increase as the difference between these quantities increase. We provide examples in \cref{proof:comparison_variances}.

\textbf{Special case: one RCT conducted across $K$ studies.}\label{subsec:one_rct_special_case}
Consider the special case where $W \indep H$, i.e., there is no edge from $H$ to $W$ in the graphical model of Figure~\ref{graph:homog}. This corresponds to a single Bernoulli RCT with treatment probability $p$ implemented across multiple studies. 
Then, under \cref{cond:local_large_sample_size}, all estimators are asymptotically equivalent (see \cref{proof:metas_equal_var_to_pool}) and thus meta-analysis estimators should be used as they require a single round of communication.

\section{HETEROGENEOUS SCENARIOS}
\label{sec:hetero}

\subsection{Distributional Shifts in Covariates}\label{subsec:diff_means}

\looseness=-1 In model~\eqref{model:model1}, we do not assume $H \indep X$ and consider the graphical model in Figure~\ref{graph:diff_distribs}, depicting \textit{Heterogeneous} Bernoulli Trials in the distribution of $X|H$.

Here, $\Pb(Z)\mkern-7mu =\mkern-7mu \Pb(Y|X,W) \Pb(X|H) \Pb(W|H) \Pb(H)$, hence the observations $(X_{k,i}, W_{k,i}, Y_{k,i})_i$ are i.i.d. within study $k$ but not necessarily across studies. We denote $\E(X_{k,i}) \!\!=\! \mu_k$, $\V(X_{k,i}) \!=\! \Sigma_k$ and we have 
$\Sigma \!=\! \V(X) \!=\! \sum_{k=1}^{K} \rho_k \Sigma_k$ and $\mu \!=\! \E(X) \!=\! \sum_{k=1}^{K} \rho_k \mu_k$. Furthermore, the local ATEs  $\tau_k = \E(Y_k(1) - Y_k(0)) =\E(X'_k)(\theta^{(1)} - \theta^{(0)})$ generally differ from each other and from the global ATE $\tau\! =\!\E(Y(1)\! -\! Y(0))\! =\! \E(X')(\theta^{(1)}\!  -\! \theta^{(0)})$. 



\textbf{Meta estimators.} The ATE can be written as $\tau = \sum_{k=1}^{K} \rho_k \tau_k$, so the Meta-SW estimator remains unbiased (as $n_k/n$ is an unbiased estimate of $\rho_k$) and has the same variance as in Table~\ref{tab:taus_var} with $\Sigma=\sum_{k=1}^{K} \rho_k \Sigma_k$  (proof in \cref{proof:meta_sw_diff_means}). In contrast, the Meta-IVW estimator becomes unsuitable under distributional shifts: IVW weights give biased estimates of the $\rho_k$'s, leading to a biased estimate of $\tau$.

\textbf{Pool and GD estimators.} The pooled and GD estimators are robust to covariate shifts, leading them to be also unbiased with same variance as in Table~\ref{tab:taus_var} (as proved in \cref{proof:gd_pool_diff_means_var}).


\textbf{One-shot estimators.} Although the One-Shot IVW outcome parameters estimator still enjoys \cref{prop:1S_ivw_thetas_equal_pool}, the variance of the One-Shot SW one is impacted by the difference in population means at each study.
\begin{proposition}[Larger Variance of 1S\text{-}SW, proof in \ref{proof:1s_diff_means_var}]\label{prop:1s_diff_means_var}
    Under \cref{cond:local_large_sample_size}, the one-shot federated estimators are unbiased and \vspace*{-1mm}
    \begin{align*}
         \V(\hat \theta_{\mathrm{pool}}) =  \V(\hat \theta_{\mathrm{GD}}) = \V(\hat \theta_{\mathrm{1S\text{-}IVW}}) \leq \V(\hat \theta_{\mathrm{1S\text{-}SW}})
    \end{align*}
    which yields\vspace*{-1mm} 
    \begin{align*}
         \aVar(\hat \tau_{\mathrm{pool}}) = \aVar(\hat \tau_{\mathrm{GD}}) =  \aVar(\hat \tau_{\mathrm{1S\text{-}IVW}})  \leq \aVar(\hat \tau_{\mathrm{1S\text{-}SW}})
    \end{align*}
\end{proposition}

\begin{theorem}[Comparison of asymptotic variances under distributional shift]
    \begin{align*}
    \aVar(\hat \tau_{\mathrm{pool}}) \mkern-3mu=\mkern-3mu \aVar(\hat \tau_{\mathrm{GD}}) \mkern-3mu=\mkern-3mu  \aVar(\hat \tau_{\mathrm{1S\text{-}IVW}}) \mkern-3mu\leq\mkern-3mu \aVar(\hat \tau_{\mathrm{Meta\text{-}SW}})
\end{align*}
\end{theorem}

Note that in this setting the DM estimators are no longer unbiased as there is a path between $W$ and $Y$ in Figure~\ref{graph:diff_distribs}. Here, $X$ is a sufficient adjustment set and adding $H$ could be harmful, particularly when the
association between $X$ and $Y$ is weak and the association between $H$ and $W$ is strong \citep[Lemma 2 therein]{JMLR:v21:19-1026}. 

\subsection{Study-Effects}\label{subsec:study_effects_intercepts}
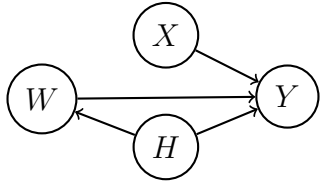
\begin{figure}[t]
    \centering
    \begin{tikzpicture}[node distance={10mm}, thick, main/.style = {draw, circle}]
    \node[main] (1) {$W$};
    \node[main] (2) [above right=2mm and 10mm of 1] {$X$};
    \node[main] (3) [below right=2mm and 10mm of 2] {$Y$};
    \node[main] (4) [below=6mm of 2] {$H$};
    \draw[->] (2) -- (3);
    \draw[->] (1) -- (3);
    \draw[->] (4) -- (1);
    \draw[->] (4) -- (3);
    \coordinate (midpoint) at ($(1)!0.5!(3)$);
    \end{tikzpicture}
    \caption{Graphical model for the heterogeneous study-effects setting.} 
    \label{graph:diff_intercepts}
\end{figure}
\looseness=-1 We now consider the graphical model shown in Figure~\ref{graph:diff_intercepts}, exhibiting an effect of study $H$ onto the outcome $Y$. The distribution of $Z$ decomposes as $\Pb(Z)=\Pb(Y|X,W,H) \Pb(X|H) \Pb(W|H) \Pb(H)$. We modify model~(\ref{model:model1}) to account for a constant study effect on individual outcomes by adding a term 
$h_k \in \mathbb{R}$:
\begin{equation}\label{model:model2}
    Y_{k,i}(w) = c^{(w)} + h_k + X_{k,i} \beta^{(w)} + \varepsilon_i (w)
\end{equation}
Model \eqref{model:model2} accounts for the possibility that studies may have different baselines in individual outcomes resulting from varying practices or organizational contexts. Here, $\tau$ and $\tau_k$ are still defined as $\E(Y(1) - Y(0)) = c^{(1)} - c^{(0)} + \E(X)(\beta^{(1)} - \beta^{(0)})$ and $\E(Y_k(1) - Y_k(0)) = c^{(1)} - c^{(0)} + \E(X_k)(\beta^{(1)} - \beta^{(0)})$ respectively since the $h_k$ terms cancel out in the differences. In other words, this modeling assumes that the Conditional Average Treatment Effect (CATE) remains consistent across studies, while allowing for an additive shift in the outcomes at each study.
In this setting, aggregating multiple RCTs into the pooled data is not itself an RCT because $H$ is now a confounder, affecting both the outcome variable $Y_{k,i}(w)$ through $h_k$ and the treatment variable $W_{k,i}$ through the treatment probability $p_k$, thereby violating the ignorability assumption (\hyperlink{as:unconfoundedness}{c}).
Therefore, the (unadjusted) pooled OLS estimator \eqref{eq:pool} is biased (see proof in Appendix~\ref{proof:bias_ols_study_effect_intercepts}). An unbiased estimator can be obtained by adjusting the model to incorporate the study effect.

\textbf{Meta estimators.} While most estimators need to be adjusted and thus require prior knowledge on the underlying model, Meta-analysis estimators can be applied directly without such modifications. We prove in \cref{proof:meta_study_effects_intercepts}\label{prop:meta_study_effects_intercepts} that under Graphical Model~\ref{graph:diff_intercepts} and model \eqref{model:model2}, the Meta-IVW (which is relevant here as $H \indep X$) and Meta-SW estimators remain unbiased, with asymptotic variances as in \cref{tab:taus_var}.

\textbf{Adjusted gradient-based estimator.} We augment the design matrix $X$ with $K-1$ dummy variables $H=\{H_2, \dots, H_K\}$ and note $\tilde X'_{k,i} = (1, X_{k,i}, H_{2,i}, \ldots, H_{K,i})$. 
Then, the method is the same as in \cref{sec:fed_est}:  $\hat \theta_{\mathrm{GD}\circ}^{(w)} \mkern-7mu=\mkern-7mu \big(\hat c^{(w)}, \hat \beta^{(w)}, \hat h_2, \dots, \hat h_K\big) \in \mathbb{R}^{d+K}$ is obtained with FedAvg and used to compute the local ATEs before final aggregation. 

\begin{definition}[Adjusted GD estimator]\label{def:GD_adj}
    \begin{align*}
        \textstyle\hat \tau_{\mathrm{GD}\circ} &= \frac{1}{n}\sum_{k=1}^{K} \sum_{i=1}^{n_k} \big(\tilde X'_{k,i}(\hat \theta_\mathrm{GD\circ}^{(1)} - \hat \theta_\mathrm{GD\circ}^{(0)})\big)
    \end{align*}
\end{definition}
This estimator is unbiased, and we do not pay a price in terms of asymptotic variance in adjusting the variables $H$. Indeed, its asymptotic variance is equal to the unadjusted $\hat \tau_\mathrm{GD}$'s one in \cref{tab:taus_var} (proof in Appendix~\ref{proof:gd_intercepts}) since $h_k$ is equal in both treatment arms and cancel out in the difference of the true parameters.


\textbf{Adjusted one-shot federated estimators.} Like the vanilla pooled and GD estimators, the one-shot estimators are biased in the presence of study-effects. However, their adjustment procedure is different because the one-shot procedure does not allow the inclusion of membership variables, as the variance matrices of the local dummy-augmented datasets are not full rank, violating \cref{cond:local_large_sample_size}. Instead, we compute the OLS $\hat\theta_k$ at each study, then share and aggregate only the coefficients $\hat \beta_k^{(w)}$, without federating the locally estimated intercepts $\hat a_k^{(w)} = \hat c^{(w)} + \hat h_k$. 

\begin{definition}[Adjusted one-shot ATE estimators]\label{def:1S_adj}
    \begin{align*}
        \hat \tau_{\mathrm{1S\text{-}SW}\circ} &= \frac{1}{n} \sum_{k=1}^{K} \sum_{i=1}^{n_k} \bigl(\hat a_k^{(1)} \mkern-3mu-\mkern-3mu \hat a_k^{(0)} \mkern-3mu+\mkern-3mu X_{k,i}(\hat \beta^{(1)}_{\mathrm{SW}\circ} \mkern-3mu-\mkern-3mu \hat \beta^{(0)}_{\mathrm{SW}\circ})\bigr)\\
        \hat \tau_{\mathrm{1S\text{-}IVW}\circ} &= \frac{1}{n} \sum_{k=1}^{K}\sum_{i=1}^{n_k} \bigl(\hat a_k^{(1)} \mkern-3mu-\mkern-3mu \hat a_k^{(0)} \mkern-3mu+\mkern-3mu {X}_{k,i}(\hat \beta^{(1)}_{\mathrm{IVW}\circ} \mkern-3mu-\mkern-3mu \hat \beta^{(0)}_{\mathrm{IVW}\circ})\bigr)
    \end{align*}
    with ${\scriptstyle \hat \beta_{\mathrm{IVW}\circ}^{(w)}} \!=\! \frac{\sum_{k=1}^K \!\V(\hat\beta_k^{(w)})^{-1}\hat \beta_k^{(w)}}{\sum_{l=1}^K \V(\hat\beta_l^{(w)})^{-1}},\, {\scriptstyle \hat \beta_{\mathrm{SW}\circ}^{(w)}} \!=\! \sum_{k=1}^K \!\frac{n_k^{(w)}}{n^{(w)}}{\scriptstyle \hat \beta_k^{(w)}}.$
\end{definition}

Note that, in general, $\hat \beta_{\mathrm{IVW}\circ}^{(w)} \neq \hat \beta_{\mathrm{IVW}}^{(w)}$ as the aggregation weights $\V(\hat\beta_k^{(w)})^{-1}$ and $\V(\hat\theta_k^{(w)})^{-1}$ are different.

\looseness=-1 The variances of the adjusted one-shot estimators are affected by the lack of federation of the local intercepts $\{\hat a_k^{(w)}\}_k$, which converge at a rate of $1/n_k^{(w)}$ rather than $1/n^{(w)}$. Additionally, they estimate the study-effects twice in each study on independent data, whereas GD$\circ$ and pool$\circ$ estimate them on $n^{(w)}$ observations. As a result, adjusted one-shot estimators generally underperform compared to other methods (see simulations in \cref{sec:simulation_figures_appendix}).
\begin{figure*}[!ht]
    \centering
    \begin{minipage}[t]{0.9\textwidth}
        \centering
        \includegraphics[width=\linewidth]{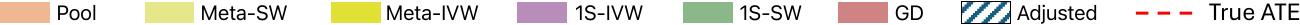}
        \vspace*{-.5cm}
    \end{minipage}
    \begin{subfigure}[t]{0.39\textwidth}
        \centering
        \begin{subfigure}[t]{0.5\textwidth}
            \centering
            \includegraphics[height=4cm, width=3.5cm, trim={0.2cm 1cm 0 1.1cm}, clip]{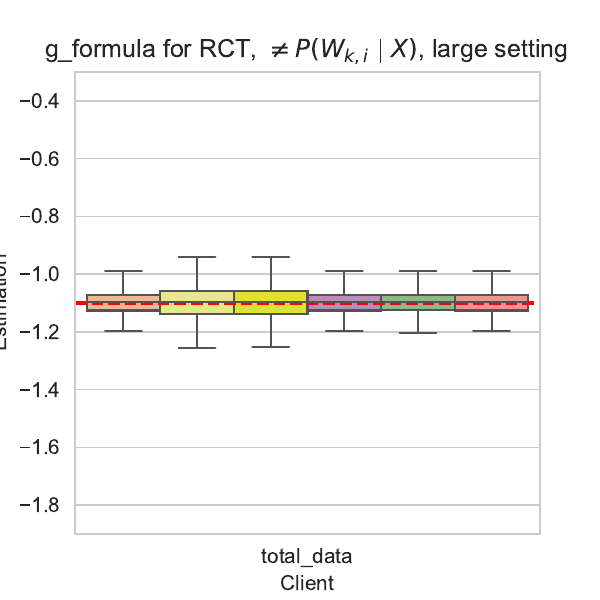}
        \end{subfigure}%
        \begin{subfigure}[t]{0.5\textwidth}
            \centering
            \includegraphics[height=4cm, width=3.5cm, trim={0.2cm 1cm 0 1.1cm}, clip]{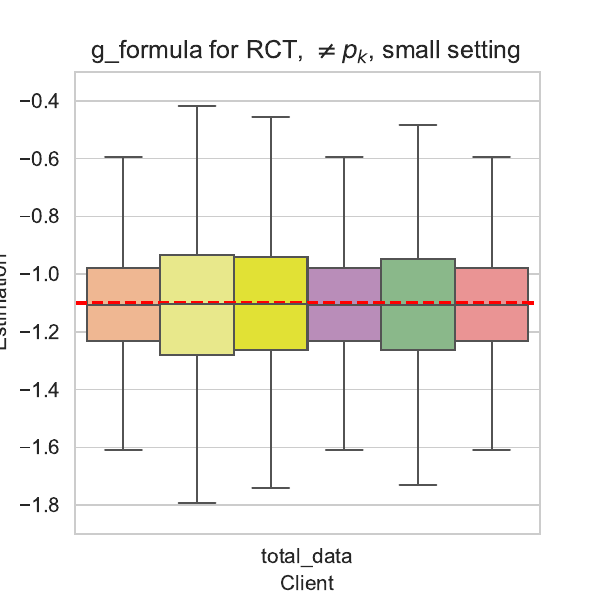}
        \end{subfigure}
        \subcaption{\centering Homogeneous Case (Figure~\ref{graph:homog}): Large (left) and Small (right) regimes}\label{fig:simu_homog}
    \end{subfigure}
    \begin{subfigure}[t]{0.6\textwidth}
        \begin{subfigure}[t]{0.33\textwidth}
            \centering
            \includegraphics[height=4cm, width=3.5cm, trim={0.4cm 1cm 0 1.1cm}, clip]{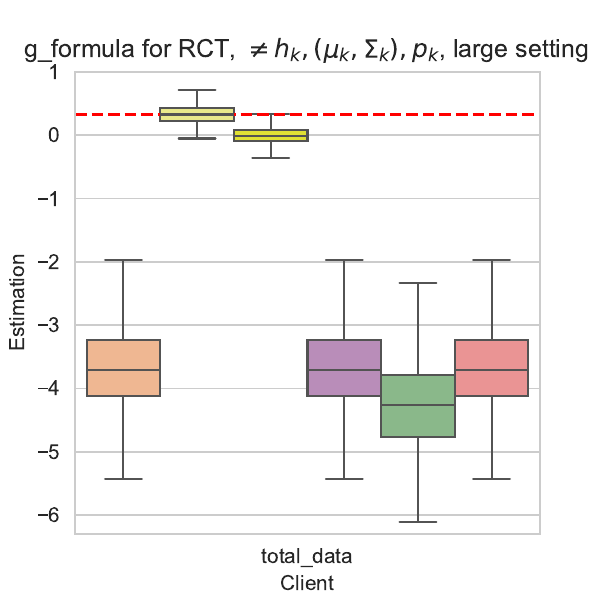}
        \end{subfigure}%
        \begin{subfigure}[t]{0.33\textwidth}
            \centering
            \includegraphics[height=4cm, width=3.3cm, trim={0cm 0cm 0 0cm}, clip]{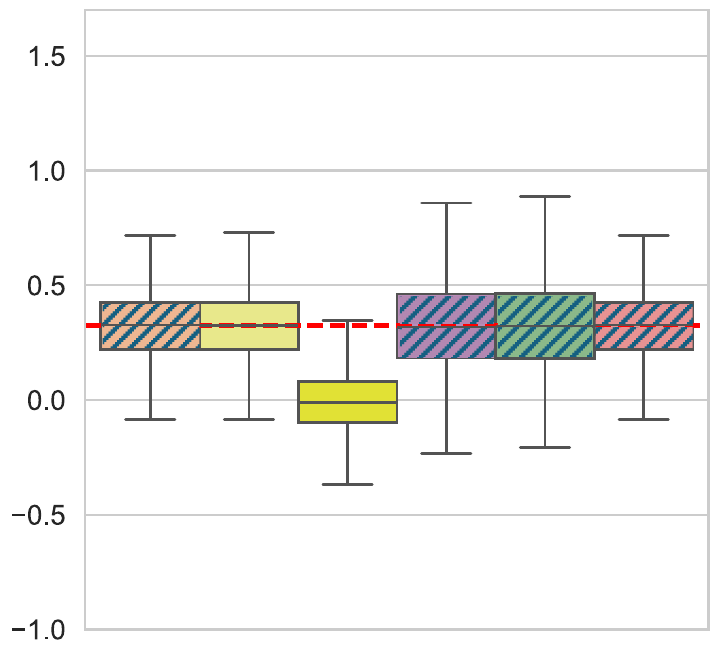}
        \end{subfigure}%
        \begin{subfigure}[t]{0.33\textwidth}
            \centering
            \includegraphics[height=4cm, width=3.3cm, trim={0cm 0cm 0cm 0cm}, clip]{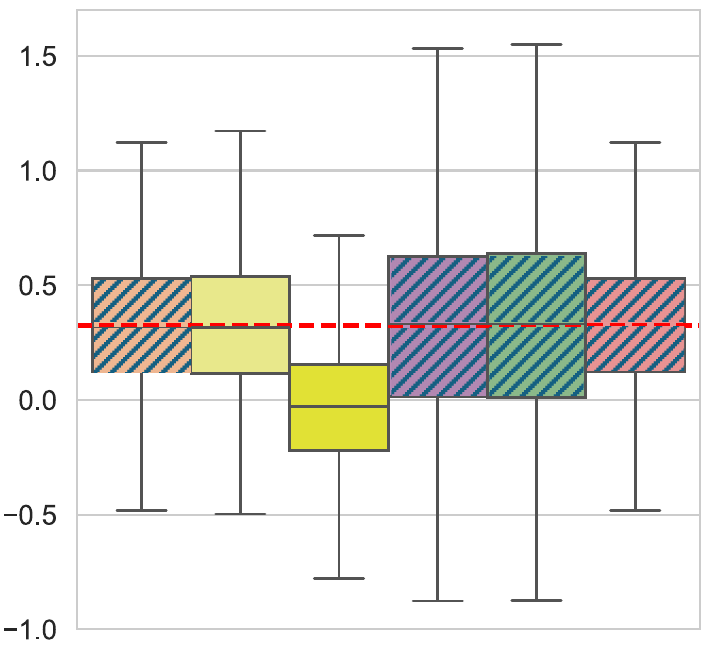}
        \end{subfigure}
        \subcaption{\centering Fully Heterogeneous Case (Figure~\ref{graph:full_hetero}): Large regime, unadjusted (left); Large regime, adjusted (middle); Small regime, adjusted (right)}
        \label{fig:simu_full_hetero}
    \end{subfigure}
    \caption{Multi-study ATE Estimation under Homogeneous and Heterogeneity Scenarios}
    \label{fig:five_scenarios}
\end{figure*}
\section{EXPERIMENTS}
\label{sec:simu}

We now present some numerical experiments on simulated and semi-synthetic data.

We compare the following estimators: Meta-SW (\cref{def:meta_sw}), Meta-IVW (\cref{def:meta_ivw}), 1S\text{-}SW (\cref{def:1SSSSS}), 1S\text{-}IVW (\cref{def:1SIVWSS}) and GD (\cref{{def:GDSW}}), and the adjusted estimators GD$\circ$ (\cref{def:GD_adj}), 1S\text{-}SW$\circ$ and 1S\text{-}IVW$\circ$ (\cref{def:1S_adj}). We also include the pooled estimator (Pool, Eq.~\ref{eq:pool}) as a baseline.
For the Meta-IVW estimator, we use empirical estimates of $\aVar(\hat{\tau}_k)^{-1}$ for the aggregation weights.

\subsection{Synthetic Data}
\looseness=-1 We generate data according to the graphical models in Figures~\eqref{graph:homog} and \eqref{model:model1}, with $K=5$ studies and $d=10$ covariates. We consider two magnitudes of sample sizes, referred to as \textit{Large} ($\forall k, n_k=20^*d$) and \textit{Small} ($\forall k, n_k=6^*d$) local sample sizes. 
We consider the following treatment assignment probabilities for the \textit{Large} setting:  $p_1 \!=\! p_2 \!=\! p_3 \!=\! 0.9, p_4 \!=\! p_5 \!=\! 0.1$. In the \textit{Small} regime we choose less extreme probabilities in order to guarantee \cref{cond:local_large_sample_size}: $p_1 \!=\! p_2 \!=\! p_3 \!=\! 0.65$, $p_4 \!=\! p_5 \!=\! 0.35$. For each scenario considered, we perform 2000 simulations and display the distribution of the global estimates of the ATE given by each estimator. More details about the simulations can be found in \cref{subsec:simulation_parameters}.

\textbf{Homogeneous case.} The results displayed in Figure~\ref{fig:simu_homog} (left) are in agreement with \cref{prop:metas_larger_var_than_pool}: the variances of the meta estimators are larger than the pooled, one-shot and gradient-based estimators when the assignment probabilities $p_k$ differ from one study to another, with the variance of the Meta-IVW being smaller than that of Meta-SW (\cref{prop:meta_ivw_min_var}). 
In the \textit{Small} regime (\cref{fig:simu_homog}, right),
the variance of One-Shot SW is larger than compared to One-Shot IVW and GD, but we still have $\V(\hat\tau_\mathrm{pool}) =\V(\hat\tau_\mathrm{GD}) = \V(\hat\tau_\mathrm{1S\text{-}IVW})$, in line with \cref{prop:1S_ivw_thetas_equal_pool}.

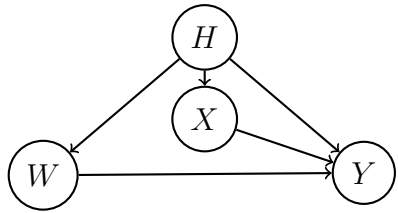
\begin{figure}[!htb]
    \centering
    \begin{tikzpicture}[node distance={7mm}, thick, main/.style = {draw, circle}, scale=0.5]
    \node[main] (1) {$W$};
    \node[main] (2) [above right=1mm and 15mm of 1] {$X$};
    \node[main] (3) [below right=1mm and 15mm of 2] {$Y$};
    \node[main] (4) [above=2mm of 2] {$H$};
    \draw[->] (2) -- (3);
    \draw[->] (1) -- (3);
    \draw[->] (4) -- (1);
    \draw[->] (4) -- (3);
    \draw[->] (4) -- (2);
    \coordinate (midpoint) at ($(1)!0.5!(3)$);
    \end{tikzpicture}
    \vspace{-0.5em}
    \caption{Graphical model in fully heterogeneous case.}
    \label{graph:full_hetero}
\end{figure}

\textbf{Heterogeneous case.}\label{subsec:full_hetero_simus}
We now consider a ``fully heterogeneous'' setting combining three sources of heterogeneity (see graphical model in Figure~\ref{graph:full_hetero}): different covariate distributions across studies, presence of study-effects, and different $p_k$ (see \cref{tab:params_full_hetero} for the chosen values of $\{(\mu_k, \Sigma_k, h_k, p_k)\}_k$). 
\cref{fig:simu_full_hetero} (left) shows that the pooled, one-shot and gradient-based estimators are biased when we do not adjust for the study-effects.  
Here, only the Meta SW estimator is unbiased.


\looseness=-1 \cref{fig:simu_full_hetero} (middle, right) shows the results of adjusted estimators, as described in \cref{subsec:study_effects_intercepts}.  
Adjusting removes their bias but comes at a cost for the one-shot estimators, whose variances are impacted by study-effects, unlike the meta and gradient-based estimators. 
In the \textit{Small} regime, the one-shot estimators further suffer from the combination of differences in covariate means and unshared local intercepts, which inflate $\aVar(\hat a_k^{(w)})$. On the other hand, the variance of the Meta-SW estimator is nearly equal to that of the adjusted pooled and adjusted GD estimators.

\subsection{Semi-Synthetic Data}
We now provide results on the Traumatrix database \citep{mayerdoubly} to test our theory on real-world data with both linear and non-linear synthetic outcomes. We focus on brain trauma patients and consider 8,097 patients scattered across 13 sites, suffering from traumatic brain injury (TBI), with 15 covariates such as systolic and diastolic blood pressure, heart rate, oxygen saturation, and information on interventions like catecholamine administration. We consider each site as an independent study, with different treatment probabilities and potential covariate shifts, corresponding to the graphical model in \cref{graph:diff_distribs}.

We regenerate the treatment and outcome variables while keeping the covariates as is. The synthetic treatment, independent of the covariates as in an RCT setting, is generated as a Bernoulli variable that has probability $p_k$ at site $k$, with $p_k$ ranging from 0.2 to 0.8. Then, in the linear setting, we generate two continuous outcomes $Y^{(1)}$ and $Y^{(0)}$, respectively as $X' \theta^{(1)} + \varepsilon$ and $X' \theta^{(0)} + \varepsilon$, with $X'\in\mathbb{R}^{8,097 \times 16}$, $\theta^{(1)},\theta^{(0)}\in\mathbb{R}^{16}$ the parameters generated (drawn from $U([-1,1]^d)$) once with each coordinate $\theta^{(1)}_l = \theta^{(0)}_l + 0.05$ so that the true ATE $\tau = \mathbb{E}(X)(\theta^{(1)} - \theta^{(0)}) \approx \overline{X}(\theta^{(1)} - \theta^{(0)}) \approx 0.37$, and $\varepsilon \sim N(0, 2)$. Finally, we bootstrapped 3,000 times the dataset and computed the estimators on these resampled data to estimate their means and variances.

\looseness=-1 We also consider non-linear outcomes, repeating the same steps as above but with polynomial non-linear outcome model equations with interaction terms $\mu_{w}(X) \!\!=\!\!\theta_0^{(w)} \!+\! \sum_{j=1}^3 X_j^j \theta_j^{(w)}  \!\!+\!\! \sum_{j=4}^{16} X_j \theta_j^{(w)}  \!+\! \theta_\mathrm{int}^{(w)} X_\mathrm{int}  \!+\! \varepsilon$, where the interaction terms are given by $X_\mathrm{int}\!\!=\!\!\left(-X_2 \!\times\! X_3, X_1\!\times\! X_4 \right)$. This leads to a true ATE $\tau \approx 1.07$. For these non-linear outcomes, the estimators still rely on linear regressions to model the outcomes. As stated above, even if the model is misspecified in this case, adjusting for covariates is still recommended in RCTs to reduce variance.

\looseness=-1 Note that we chose $p_k$ and $n_\mathrm{bootstrap}=4,000$ to ensure that \cref{cond:local_large_sample_size} holds, as some sites have very small datasets. 
In this regard, the semi-synthetic simulation does not match the asymptotic regime on which our results rely, but it echoes the empirical results on the finite-sample regime.
\begin{table}[h]
    \centering
    \begin{tabular}{lcc}
        \toprule
        Estimator & Squared Bias & RMSE \\
        \midrule
        Meta SW & 0.000 & 1.238 \\
        Meta IVW & 0.002 & 0.116 \\
        One-Shot SW, SW Agg & 0.004 & 1.161 \\
        One-Shot IVW, SW Agg & 0.005 & 0.117 \\
        GD & 0.001 & 0.073 \\
        Pool & 0.001 & 0.073 \\
        \bottomrule
    \end{tabular}
    \caption{ATE estimation on semi-synthetic data with linear outcome}
    \label{tab:linear_synthetic_sim}
\end{table}

As expected, most estimators appear to achieve a low (empirical) bias, but differ significantly in terms of Root Mean Square Error (RMSE). Additionally, as the $p_k$ are different, the meta estimators struggle more.
In the linear setting (\cref{tab:linear_synthetic_sim}), the variance ranking of the estimators aligns with our theory: the Pool and GD-based estimators yield the same results, with the lowest variance: we have $\mathbb{V}(\hat\tau_\mathrm{pool}) = \mathbb{V}(\hat\tau_\mathrm{GD}) \leq \mathbb{V}(\hat\tau_\mathrm{meta-IVW}) \leq \mathbb{V}(\hat\tau_\mathrm{meta-SW})$.

\begin{table}[h]
    \centering
    \begin{tabular}{lcc}
        \toprule
        Estimator & Squared Bias & RMSE \\
        \midrule
        Meta SW & 0.004 & 1.923 \\
        Meta IVW & 0.007 & 0.136 \\
        One-Shot SW, SW Agg & 0.010 & 1.064 \\
        One-Shot IVW, SW Agg & 0.002 & 0.120 \\
        GD & 0.001 & 0.078 \\
        Pool & 0.001 & 0.078 \\
        \bottomrule
    \end{tabular}
    \caption{ATE estimation on semi-synthetic data with nonlinear outcome}
    \label{tab:nonlinear_synthetic_sim}
\end{table}

In the non-linear setting (\cref{tab:nonlinear_synthetic_sim}), the same conclusions seem to hold, although a more thorough analysis should be conducted to theoretically corroborate these results.

\section{CONCLUSION}

\looseness=-1 After clearly defining the population and estimand of interest, our findings can be turned into clear guidelines for practitioners that we summarize as a decision diagram (Figure~\ref{fig:diagramdec} in Appendix~\ref{app:diagram}). We recommend the one-shot IVW estimator when each study can perform a local OLS regression (\cref{cond:local_large_sample_size}) and there are no study-effects. However, this estimator requires studies to share sample covariance matrices, which can be impractical in high-dimensional settings or when privacy is a concern.
On the other hand, meta estimators are preferable when study-effects are present or when there is limited prior knowledge about the underlying model, provided \cref{cond:local_large_sample_size} holds.
Caution is required with Meta-IVW, which has lower variance than Meta-SW but is biased when studies analyze different populations.
Finally, our (adjusted) GD estimator allows us to estimate the ATE under the weaker \cref{cond:federated_large_sample_size} with the same precision as if the data were pooled, regardless of the setting.
    
\looseness=-1 Several challenges remain for the deployment of these methods, particularly in medical contexts. Future work includes addressing covariate mismatch, where some features are missing in specific studies, estimating non-collapsible causal measures (e.g., odds ratio), and 
extensions to observational studies.
Finally, a key practical challenge in multi-study settings is to ensure consistent data encoding, especially for outcomes.

\newpage
\clearpage
\bibliographystyle{plainnat}
\bibliography{bib}

\begin{thebibliography}{50}
\providecommand{\natexlab}[1]{#1}
\providecommand{\url}[1]{\texttt{#1}}
\expandafter\ifx\csname urlstyle\endcsname\relax
  \providecommand{\doi}[1]{doi: #1}\else
  \providecommand{\doi}{doi: \begingroup \urlstyle{rm}\Url}\fi

\bibitem[Benkeser et~al.(2021)Benkeser, D{\'\i}az, Luedtke, Segal, Scharfstein, and Rosenblum]{benkeser2021improving}
David Benkeser, Iv{\'a}n D{\'\i}az, Alex Luedtke, Jodi Segal, Daniel Scharfstein, and Michael Rosenblum.
\newblock Improving precision and power in randomized trials for covid-19 treatments using covariate adjustment, for binary, ordinal, and time-to-event outcomes.
\newblock \emph{Biometrics}, 77\penalty0 (4):\penalty0 1467--1481, 2021.

\bibitem[Berlin and Golub(2014)]{berlin2014meta}
Jesse~A. Berlin and Robert~M. Golub.
\newblock Meta-analysis as evidence: Building a better pyramid.
\newblock \emph{JAMA}, 312\penalty0 (6):\penalty0 603--606, 2014.
\newblock \doi{10.1001/jama.2014.8167}.

\bibitem[Blunt(2015)]{blunt2015hierarchies}
Christopher Blunt.
\newblock \emph{Hierarchies of evidence in evidence-based medicine}.
\newblock PhD thesis, London School of Economics and Political Science, 2015.

\bibitem[Borenstein et~al.(2021)Borenstein, Hedges, Higgins, and Rothstein]{borenstein2021introduction}
Michael Borenstein, Larry~V Hedges, Julian~PT Higgins, and Hannah~R Rothstein.
\newblock \emph{Introduction to meta-analysis}.
\newblock John Wiley \& Sons, 2021.

\bibitem[Brantner et~al.(2023)Brantner, Chang, Nguyen, Hong, Di~Stefano, and Stuart]{brantner2023methods}
Carly~Lupton Brantner, Ting-Hsuan Chang, Trang~Quynh Nguyen, Hwanhee Hong, Leon Di~Stefano, and Elizabeth~A Stuart.
\newblock Methods for integrating trials and non-experimental data to examine treatment effect heterogeneity.
\newblock \emph{Statistical Science}, 38\penalty0 (4):\penalty0 640--654, 2023.

\bibitem[Che et~al.(2022)Che, Kong, Peng, Sun, Leow, Chen, and He]{che2022federated}
Sicong Che, Zhaoming Kong, Hao Peng, Lichao Sun, Alex Leow, Yong Chen, and Lifang He.
\newblock Federated multi-view learning for private medical data integration and analysis.
\newblock \emph{ACM Transactions on Intelligent Systems and Technology (TIST)}, 13\penalty0 (4):\penalty0 1--23, 2022.

\bibitem[Edmondson et~al.(2023)Edmondson, Luo, and Chen]{edmondson2023statistical}
Mackenzie~J Edmondson, Chongliang Luo, and Yong Chen.
\newblock Statistical analysis—meta-analysis/reproducibility.
\newblock In \emph{Clinical Applications of Artificial Intelligence in Real-World Data}, pages 125--139. Springer, 2023.

\bibitem[{European Medicines Agency}(2024)]{EMA2024}
{European Medicines Agency}.
\newblock {ICH E9 Statistical Principles for Clinical Trials: Scientific Guideline}, 2024.

\bibitem[{French Health Authority}(2024)]{SanteGouv}
{French Health Authority}.
\newblock Pricing \& reimbursement of drugs and hta policies in france, 2024.
\newblock URL \url{https://www.has-sante.fr/upload/docs/application/pdf/2014-03/pricing_reimbursement_of_drugs_and_hta_policies_in_france.pdf}.

\bibitem[Giraud et~al.(2012)Giraud, Huet, and Verzelen]{giraud2012high}
Christophe Giraud, Sylvie Huet, and Nicolas Verzelen.
\newblock High-dimensional regression with unknown variance.
\newblock 2012.

\bibitem[Guo et~al.(2024)Guo, Karimireddy, and Jordan]{DBLP:journals/corr/abs-2404-15746}
Tianyu Guo, Sai~Praneeth Karimireddy, and Michael~I. Jordan.
\newblock Collaborative heterogeneous causal inference beyond meta-analysis.
\newblock \emph{arXiv preprint arXiv:2404.15746}, 2024.

\bibitem[Guyatt et~al.(2015)Guyatt, Rennie, Meade, and Cook]{Cook2014UserGuide}
Gordon Guyatt, Drummond Rennie, Maureen~O. Meade, and Deborah~J. Cook.
\newblock \emph{Users' Guides to the Medical Literature : A Manual for Evidence-Based Clinical Practice.}
\newblock McGraw-Hill Education, New York, 2015.

\bibitem[Han et~al.(2021)Han, Hou, Cho, Duan, and Cai]{han2021federated}
Larry Han, Jue Hou, Kelly Cho, Rui Duan, and Tianxi Cai.
\newblock Federated adaptive causal estimation (face) of target treatment effects.
\newblock \emph{arXiv preprint arXiv:2112.09313}, 2021.

\bibitem[Han et~al.(2023)Han, Shen, and Zubizarreta]{DBLP:conf/nips/HanSZ23}
Larry Han, Zhu Shen, and Jos{\'{e}}~R. Zubizarreta.
\newblock Multiply robust federated estimation of targeted average treatment effects.
\newblock In Alice Oh, Tristan Naumann, Amir Globerson, Kate Saenko, Moritz Hardt, and Sergey Levine, editors, \emph{Advances in Neural Information Processing Systems 36: Annual Conference on Neural Information Processing Systems 2023, NeurIPS 2023, New Orleans, LA, USA, December 10 - 16, 2023}, 2023.

\bibitem[Hernan(2020)]{hernan2020book}
JM~Hernan, MA~Robins.
\newblock \emph{Causal Inference: What If.}
\newblock Boca Raton: Chapman \& Hall/CRC., 2020.

\bibitem[Higgins et~al.(2019)Higgins, Thomas, Chandler, Cumpston, Li, Page, and Welch]{higgins2019cochrane}
Julian~PT Higgins, James Thomas, Jacqueline Chandler, Miranda Cumpston, Tianjing Li, Matthew~J Page, and Vivian~A Welch, editors.
\newblock \emph{Cochrane Handbook for Systematic Reviews of Interventions}.
\newblock John Wiley \& Sons, Chichester, UK, 2nd edition, 2019.

\bibitem[Hunter and Schmidt(2004)]{hunter2004methods}
John~E Hunter and Frank~L Schmidt.
\newblock \emph{Methods of meta-analysis: Correcting error and bias in research findings}.
\newblock Sage, 2004.

\bibitem[Imbens and Rubin(2015)]{imbens2015causal}
Guido~W Imbens and Donald~B Rubin.
\newblock \emph{{Causal Inference in Statistics, Social, and Biomedical Sciences}}.
\newblock Cambridge University Press, Cambridge UK, 2015.

\bibitem[Kahan et~al.(2023)Kahan, Li, Copas, and Harhay]{kahan2023estimands}
Brennan~C Kahan, Fan Li, Andrew~J Copas, and Michael~O Harhay.
\newblock Estimands in cluster-randomized trials: choosing analyses that answer the right question.
\newblock \emph{International Journal of Epidemiology}, 52\penalty0 (1):\penalty0 107--118, 2023.

\bibitem[Kairouz et~al.(2021)Kairouz, McMahan, Avent, Bellet, Bennis, Bhagoji, Bonawitz, Charles, Cormode, Cummings, et~al.]{kairouz2021advances}
Peter Kairouz, H~Brendan McMahan, Brendan Avent, Aur{\'e}lien Bellet, Mehdi Bennis, Arjun~Nitin Bhagoji, Kallista Bonawitz, Zachary Charles, Graham Cormode, Rachel Cummings, et~al.
\newblock Advances and open problems in federated learning.
\newblock \emph{Foundations and trends{\textregistered} in machine learning}, 14\penalty0 (1--2):\penalty0 1--210, 2021.

\bibitem[Khaled et~al.(2020)Khaled, Mishchenko, and Richt{\'{a}}rik]{DBLP:conf/aistats/0001MR20}
Ahmed Khaled, Konstantin Mishchenko, and Peter Richt{\'{a}}rik.
\newblock Tighter theory for local {SGD} on identical and heterogeneous data.
\newblock In \emph{AISTATS}, 2020.

\bibitem[Koga et~al.(2024)Koga, Chaudhuri, and Page]{koga2024differentially}
Tatsuki Koga, Kamalika Chaudhuri, and David Page.
\newblock Differentially private multi-site treatment effect estimation.
\newblock In \emph{2024 IEEE Conference on Secure and Trustworthy Machine Learning (SaTML)}, pages 472--489. IEEE, 2024.

\bibitem[Koloskova et~al.(2020)Koloskova, Loizou, Boreiri, Jaggi, and Stich]{pmlr-v119-koloskova20a}
Anastasia Koloskova, Nicolas Loizou, Sadra Boreiri, Martin Jaggi, and Sebastian Stich.
\newblock A unified theory of decentralized {SGD} with changing topology and local updates.
\newblock In Hal~Daumé III and Aarti Singh, editors, \emph{Proceedings of the 37th International Conference on Machine Learning}, volume 119 of \emph{Proceedings of Machine Learning Research}, pages 5381--5393. PMLR, 13--18 Jul 2020.
\newblock URL \url{https://proceedings.mlr.press/v119/koloskova20a.html}.

\bibitem[Lei and Ding(2021)]{lei2021regression}
Lihua Lei and Peng Ding.
\newblock Regression adjustment in completely randomized experiments with a diverging number of covariates.
\newblock \emph{Biometrika}, 108\penalty0 (4):\penalty0 815--828, 2021.

\bibitem[Liberati et~al.(2009)Liberati, Altman, Tetzlaff, and Mulrow]{liberati2009g0tzsche}
A~Liberati, DG~Altman, J~Tetzlaff, and C~Mulrow.
\newblock G0tzsche pc, ioannidis jp, et al. the prisma statement for reporting systematic reviews and meta-analyses of studies that evaluate health care interventions: explanation and elaboration.
\newblock \emph{J Clin Epidemiol}, 62\penalty0 (10):\penalty0 e1--34, 2009.

\bibitem[Lin(2013)]{lin2013agnostic}
Winston Lin.
\newblock Agnostic notes on regression adjustments to experimental data: Reexamining freedman’s critique.
\newblock 2013.

\bibitem[Makhija et~al.(2024)Makhija, Ghosh, and Kim]{DBLP:journals/corr/abs-2402-17705}
Disha Makhija, Joydeep Ghosh, and Yejin Kim.
\newblock Federated learning for estimating heterogeneous treatment effects.
\newblock \emph{CoRR}, abs/2402.17705, 2024.
\newblock \doi{10.48550/ARXIV.2402.17705}.
\newblock URL \url{https://doi.org/10.48550/arXiv.2402.17705}.

\bibitem[Mayer et~al.(2020)Mayer, Sverdrup, Gauss, Moyer, Wager, and Josse]{mayerdoubly}
Imke Mayer, Erik Sverdrup, Tobias Gauss, Jean-Denis Moyer, Stefan Wager, and Julie Josse.
\newblock {Doubly robust treatment effect estimation with missing attributes}.
\newblock \emph{The Annals of Applied Statistics}, 14\penalty0 (3):\penalty0 1409 -- 1431, 2020.

\bibitem[McMahan et~al.(2017)McMahan, Moore, Ramage, Hampson, and y~Arcas]{mcmahan2017communication}
Brendan McMahan, Eider Moore, Daniel Ramage, Seth Hampson, and Blaise~Aguera y~Arcas.
\newblock Communication-efficient learning of deep networks from decentralized data.
\newblock In \emph{Artificial intelligence and statistics}, pages 1273--1282. PMLR, 2017.

\bibitem[Moher et~al.(1999)Moher, Cook, Eastwood, Olkin, Rennie, and Stroup]{moher1999improving}
David Moher, Deborah~J Cook, Susan Eastwood, Ingram Olkin, Drummond Rennie, and Donna~F Stroup.
\newblock Improving the quality of reports of meta-analyses of randomised controlled trials: the quorom statement.
\newblock \emph{The Lancet}, 354\penalty0 (9193):\penalty0 1896--1900, 1999.

\bibitem[Morris et~al.(2018)Morris, Fisher, Kenward, and Carpenter]{morris2018meta}
Tim~P Morris, David~J Fisher, Michael~G Kenward, and James~R Carpenter.
\newblock Meta-analysis of gaussian individual patient data: Two-stage or not two-stage?
\newblock \emph{Statistics in medicine}, 37\penalty0 (9):\penalty0 1419--1438, 2018.

\bibitem[Piantadosi et~al.(1988)Piantadosi, Byar, and Green]{piantadosi1988ecological}
Steven Piantadosi, David~P Byar, and Sylvan~B Green.
\newblock The ecological fallacy.
\newblock \emph{American journal of epidemiology}, 127\penalty0 (5):\penalty0 893--904, 1988.

\bibitem[Prosperi et~al.(2020)Prosperi, Guo, Sperrin, Koopman, Min, He, Rich, Wang, Buchan, and Bian]{prosperi2020causal}
Mattia Prosperi, Yi~Guo, Matt Sperrin, James~S Koopman, Jae~S Min, Xing He, Shannan Rich, Mo~Wang, Iain~E Buchan, and Jiang Bian.
\newblock Causal inference and counterfactual prediction in machine learning for actionable healthcare.
\newblock \emph{Nature Machine Intelligence}, 2\penalty0 (7):\penalty0 369--375, 2020.

\bibitem[Robins(1986)]{robins1986new}
James Robins.
\newblock A new approach to causal inference in mortality studies with a sustained exposure period—application to control of the healthy worker survivor effect.
\newblock \emph{Mathematical modelling}, 7\penalty0 (9-12):\penalty0 1393--1512, 1986.

\bibitem[Rotnitzky and Smucler(2020)]{JMLR:v21:19-1026}
Andrea Rotnitzky and Ezequiel Smucler.
\newblock Efficient adjustment sets for population average causal treatment effect estimation in graphical models.
\newblock \emph{Journal of Machine Learning Research}, 21\penalty0 (188):\penalty0 1--86, 2020.
\newblock URL \url{http://jmlr.org/papers/v21/19-1026.html}.

\bibitem[Rubin(1974)]{RubinDonaldB1974Eceo}
Donald~B Rubin.
\newblock Estimating causal effects of treatments in randomized and nonrandomized studies.
\newblock \emph{Journal of educational psychology}, 66\penalty0 (5):\penalty0 688--701, 1974.
\newblock ISSN 0022-0663.

\bibitem[Seo et~al.(2021)Seo, White, Furukawa, Imai, Valgimigli, Egger, Zwahlen, and Efthimiou]{seo2021comparing}
Michael Seo, Ian~R White, Toshi~A Furukawa, Hissei Imai, Marco Valgimigli, Matthias Egger, Marcel Zwahlen, and Orestis Efthimiou.
\newblock Comparing methods for estimating patient-specific treatment effects in individual patient data meta-analysis.
\newblock \emph{Statistics in medicine}, 40\penalty0 (6):\penalty0 1553--1573, 2021.

\bibitem[Sheller et~al.(2020)Sheller, Edwards, Reina, Martin, Pati, Kotrotsou, Milchenko, Xu, Marcus, Colen, et~al.]{sheller2020federated}
Micah~J Sheller, Brandon Edwards, G~Anthony Reina, Jason Martin, Sarthak Pati, Aikaterini Kotrotsou, Mikhail Milchenko, Weilin Xu, Daniel Marcus, Rivka~R Colen, et~al.
\newblock Federated learning in medicine: facilitating multi-institutional collaborations without sharing patient data.
\newblock \emph{Scientific reports}, 10\penalty0 (1):\penalty0 12598, 2020.

\bibitem[Splawa-Neyman et~al.(1990)Splawa-Neyman, Dabrowska, and Speed]{Neyman}
Jerzy Splawa-Neyman, D.~M. Dabrowska, and T.~P. Speed.
\newblock {On the Application of Probability Theory to Agricultural Experiments. Essay on Principles. Section 9}.
\newblock \emph{Statistical Science}, 5\penalty0 (4):\penalty0 465 -- 472, 1990.
\newblock \doi{10.1214/ss/1177012031}.
\newblock URL \url{https://doi.org/10.1214/ss/1177012031}.

\bibitem[Stich(2019)]{stich_localsgd}
Sebastian~U. Stich.
\newblock Local sgd converges fast and communicates little.
\newblock In \emph{ICLR}, 2019.

\bibitem[Tan et~al.(2022)Tan, Chang, Zhou, and Tang]{tan2022tree}
Xiaoqing Tan, Chung-Chou~H Chang, Ling Zhou, and Lu~Tang.
\newblock A tree-based model averaging approach for personalized treatment effect estimation from heterogeneous data sources.
\newblock In \emph{International Conference on Machine Learning}, pages 21013--21036. PMLR, 2022.

\bibitem[Terrail et~al.(2023)Terrail, Klopfenstein, Li, Mayer, Loiseau, Hallal, Balazard, and Andreux]{terrail2023fedeca}
Jean Ogier~du Terrail, Quentin Klopfenstein, Honghao Li, Imke Mayer, Nicolas Loiseau, Mohammad Hallal, F{\'e}lix Balazard, and Mathieu Andreux.
\newblock Fedeca: A federated external control arm method for causal inference with time-to-event data in distributed settings.
\newblock \emph{arXiv preprint arXiv:2311.16984}, 2023.

\bibitem[Tsiatis et~al.(2008)Tsiatis, Davidian, Zhang, and Lu]{tsiatis2008covariate}
Anastasios~A Tsiatis, Marie Davidian, Min Zhang, and Xiaomin Lu.
\newblock Covariate adjustment for two-sample treatment comparisons in randomized clinical trials: a principled yet flexible approach.
\newblock \emph{Statistics in medicine}, 27\penalty0 (23):\penalty0 4658--4677, 2008.

\bibitem[{U.S. Food and Drug Administration}(2023)]{FDA2023}
{U.S. Food and Drug Administration}.
\newblock Adjusting for covariates in randomized clinical trials for drugs and biological products, 2023.

\bibitem[Van~Lancker et~al.(2024)Van~Lancker, Bretz, and Dukes]{van2024covariate}
Kelly Van~Lancker, Frank Bretz, and Oliver Dukes.
\newblock Covariate adjustment in randomized controlled trials: General concepts and practical considerations.
\newblock \emph{Clinical Trials}, page 17407745241251568, 2024.

\bibitem[Vo et~al.(2022{\natexlab{a}})Vo, Bhattacharyya, Lee, and Leong]{vo2022adaptive}
Thanh~Vinh Vo, Arnab Bhattacharyya, Young Lee, and Tze-Yun Leong.
\newblock An adaptive kernel approach to federated learning of heterogeneous causal effects.
\newblock \emph{Advances in Neural Information Processing Systems}, 35:\penalty0 24459--24473, 2022{\natexlab{a}}.

\bibitem[Vo et~al.(2022{\natexlab{b}})Vo, Lee, Hoang, and Leong]{vo2021federated}
Thanh~Vinh Vo, Young Lee, Trong~Nghia Hoang, and Tze-Yun Leong.
\newblock Bayesian federated estimation of causal effects from observational data.
\newblock In \emph{UAI}, 2022{\natexlab{b}}.

\bibitem[Wager(2020)]{wager2020stats}
Stefan Wager.
\newblock Stats 361: Causal inference.
\newblock Technical report, 2020.

\bibitem[Wang et~al.(2024)Wang, Wang, Chen, and Ji]{fedavggoodhetero}
Jiayi Wang, Shiqiang Wang, Rong-Rong Chen, and Mingyue Ji.
\newblock A new theoretical perspective on data heterogeneity in federated optimization.
\newblock In \emph{International Conference on Machine Learning (ICML)}, 2024.

\bibitem[Xiong et~al.(2023)Xiong, Koenecke, Powell, Shen, Vogelstein, and Athey]{xiong2021federated}
Ruoxuan Xiong, Allison Koenecke, Michael Powell, Zhu Shen, Joshua~T Vogelstein, and Susan Athey.
\newblock Federated causal inference in heterogeneous observational data.
\newblock \emph{Statistics in Medicine}, 42\penalty0 (24):\penalty0 4418--4439, 2023.

\end{thebibliography}



\newpage

\begin{appendices}
%
%




%

%


\onecolumn
\aistatstitle{Supplementary Materials}
\appendix

\section{DECISION DIAGRAM}
\label{app:diagram}

\tikzstyle{block} = [rectangle, draw, fill=blue!20, text width=8em, text centered, rounded corners, minimum height=4em, drop shadow]
\tikzstyle{line} = [draw, -latex']
\tikzstyle{cloud} = [draw, rectangle, fill=red!20, node distance=3cm, minimum height=2em, minimum width=2em, text width=8em,text centered, drop shadow]
\begin{figure*}[!h]
    \begin{center}
        \begin{tikzpicture}[
            node distance=1.5cm and 1.5cm,
        ]
    
        \node [block, text width=10em, ](local_full_rank) {Local Full Rank \\ (\cref{cond:local_large_sample_size})?};
        \node [block, below right=of local_full_rank, xshift=-2cm,text width=10em] (federated_full_rank) {
            Federated Full Rank (\cref{cond:federated_large_sample_size})? 
        };
        \node [block, below left=of federated_full_rank, xshift=5cm] (federated_full_rank_yes_study_effect) {Study-effects?};
        \node [cloud, below of= federated_full_rank_yes_study_effect, xshift=-1cm, yshift=-.5cm] (federated_full_rank_study_effect_yes) {Use Adjusted GD};
        \node [right of=federated_full_rank_study_effect_yes, xshift=-0.15cm, yshift=-.2cm] {\ding{72}};
        \node [cloud, below right=of federated_full_rank_yes_study_effect, xshift=-3.5cm, yshift=1cm] (federated_full_rank_study_effect_no) {Use GD};
        \node [cloud, above right=of federated_full_rank, xshift=-3.5cm, yshift=-1cm] (more_data) {Gather more data or add studies};
    
        \node [block, below left=of local_full_rank, xshift=2.5cm, text width=15em](same_distribution) {Same distribution of $X$ across studies?};
        \node[block, below left=of same_distribution, xshift=4cm] (same_d_yes_study_effects) {Study-effects?};
        \node[cloud, below left=of same_d_yes_study_effects, xshift=2cm, yshift=1cm] (same_d_yes_study_effects_yes) {Use Adjusted GD, or Meta-IVW (\cref{fig:diff_intercepts_adjusted})};
        \node [left of=same_d_yes_study_effects_yes, xshift=.25cm, yshift=0cm] {\ding{72}};
        \node[block, below of=same_d_yes_study_effects, xshift=.25cm, yshift=-1.5cm] (same_p) {Same treatment probabilities?\\ ($H\indep W$)};
        \node [cloud, below left=of same_p, xshift=2cm, yshift=1cm] (same_p_yes) {Use 1S-IVW, or Meta-IVW (\cref{fig:homog})};
        \node [cloud, below of= same_p, xshift=0.5cm, yshift=.5cm] (same_p_no) {Use 1S-IVW (\cref{fig:simu_homog})};
        \node [block, below right=of same_distribution, xshift=-4cm] (same_d_no_study_effects) {Study-effects?};
        \node [cloud, below right=of same_d_no_study_effects, xshift=-3cm, yshift=1.5cm] (same_d_no_study_effects_no) {Use 1S-IVW (\cref{fig:diff_means})};
        \node [left of=same_d_no_study_effects_no, xshift=0.25cm, yshift=0cm] {\ding{72}};
        \node [cloud, below of=same_d_no_study_effects, xshift=0cm, yshift=-.5cm] (not_same_d_study_effects) {Use Adjusted GD, or Meta-SW (\cref{fig:simu_full_hetero})};
        \node [right of=not_same_d_study_effects, xshift=-0.25cm, yshift=-.4cm] {\ding{72}};
    
        \path [line] (local_full_rank) -- node[midway, left] {no} (federated_full_rank);
        \path [line] (local_full_rank) -- node[midway, left] {yes} (same_distribution);
        \path [line] (federated_full_rank) -- node[midway, left] {yes} (federated_full_rank_yes_study_effect);
        \path [line] (federated_full_rank_yes_study_effect) -- node[midway, left] {yes} (federated_full_rank_study_effect_yes);
        \path [line] (federated_full_rank_yes_study_effect) -- node[midway, right] {no} (federated_full_rank_study_effect_no);
        \path [line] (federated_full_rank) -- node[midway, right] {no} (more_data);
    
        \path [line] (same_distribution) -- node[midway, left] {yes} (same_d_yes_study_effects);
        \path [line] (same_d_yes_study_effects) -- node[midway, left] {yes} (same_d_yes_study_effects_yes);
        \path [line] (same_d_yes_study_effects) -- node[midway, right] {no} (same_p);
        \path [line] (same_distribution) -- node[midway, right] {no} (same_d_no_study_effects);
        \path [line] (same_d_no_study_effects) -- node[midway, right] {no} (same_d_no_study_effects_no);
        \path [line] (same_d_no_study_effects) -- node[midway, left] {yes} (not_same_d_study_effects);
        \path [line] (same_p) -- node[midway, left] {no} (same_p_no);
        \path [line] (same_p) -- node[midway, right] {yes} (same_p_yes);
    
        \path [line] (more_data.west) to [out=90, in=360, looseness=0] (local_full_rank.east);
        \end{tikzpicture}
    \end{center}
      \caption{Decision Diagram for Practitionners. The sign \ding{72} denotes scenarios where the DM estimator is biased.}
        \label{fig:diagramdec}
    \end{figure*}
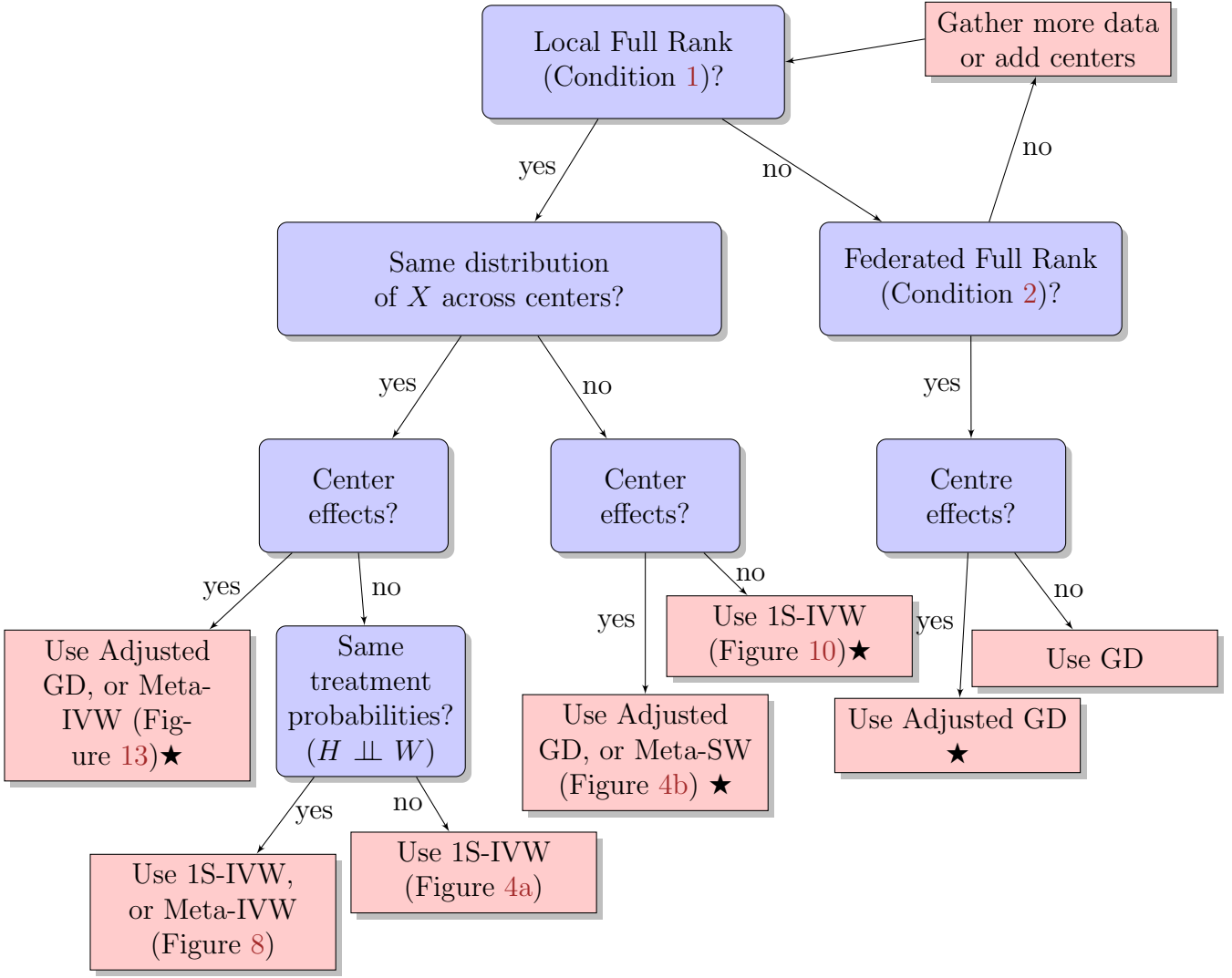

Our results yield clear guidelines for practitioners to select the most suitable estimator for different scenarios, which we present as a decision diagram in Figure~\ref{fig:diagramdec}.
Note that each time we mention the use of a meta estimator in second position, it means that this meta estimator yields a valid unbiased estimator, but with a higher variance than the estimator in first position.

It is also worth mentioning that the Difference-in-Means estimator on the pooled individual data is biased whenever $H$ acts as a confounder between $W$ and $Y$, which happens when studies have distinct treatment probabilities and study effects. The DM estimator is also biased in the graphical model in \cref{graph:diff_distribs}, although $H$ is not technically a confounder in this setting.

\section{PROOFS}
\subsection{Weighting methods}
\subsubsection{Probability of treatment in pooled dataset}\label{proof:p_pool_is_weighted_avg_of_pk}
We consider that a study is included in the federated study if its sample size is strictly larger than zero, i.e. $\forall k, n_k>0$. Since $n_k$ is a binomial random variable of parameters $n$ and $\rho_k$, we have $\E(n_k) = n \rho_k$ which yields $\rho_k = \E(\frac{n_k}{n})$.

\subsubsection{Meta-IVW has minimum variance among unbiased aggregation-based estimators}\label{proof:meta_ivw_min_var} 
Proof of \cref{prop:meta_ivw_min_var}:
Let $\hat{\tau}_k \sim \mathcal{N}(\tau, \V(\hat{\tau}_k))$ and $K$ independent studies. We denote as $\hat{\tau} = \frac{\sum_{k=1}^{K} w_k \hat{\tau}_k}{\sum_{k=1}^{K} w_k}$ a $w$-weighted average of the local estimators of $\tau$. We have:
    \begin{align*}
        \V(\hat{\tau}) &= \V\left(\frac{\sum_{k=1}^{K} w_k \hat{\tau}_k}{\sum_{k=1}^{K} w_k}\right) \\
        &= \frac{1}{\left(\sum_{k=1}^{K} w_k\right)^2} \sum_{k=1}^{K} w_k^2 \V\left(\hat{\tau}_k\right) \\
        &= \sum_{k=1}^{K} \left(\frac{w_k}{\sum_{k=1}^{K} w_k}\right)^2 \V\left(\hat{\tau}_k\right) \\
        &= \sum_{k=1}^{K} u_k^2 \V\left(\hat{\tau}_k\right) \\
    \end{align*}
    \quad \quad \quad with $u_k = \frac{w_k}{\sum_{k=1}^{K} w_k}$ and $\sum_{k=1}^{K} u_k = 1$\\

    We want to minimize $\V(\hat{\tau})$ under the constraint $\sum_{k=1}^{K} u_k = 1$. We can use the Lagrange multiplier method to find the minimum of $\V(\hat{\tau})$ under this constraint. We define the Lagrangian function:
    $$\mathcal{L}(u_1, \dots, u_K, \lambda) = \sum_{k=1}^{K} u_k^2 \V\left(\hat{\tau}_k\right) + \lambda \left(1 - \sum_{k=1}^{K} u_k\right)$$
    Then we have:
    \begin{align*}
        \frac{\partial \mathcal{L}}{\partial u_k} &= 2 u_k \V\left(\hat{\tau}_k\right) - \lambda \\
        \frac{\partial \mathcal{L}}{\partial \lambda} &= 1 - \sum_{k=1}^{K} u_k \\
    \end{align*}
    To cancel out the two derivatives above, we get that $\forall k \in [1;K], u_k = \frac{\lambda}{2 \V\left(\hat{\tau}_k\right)}$. Then starting back from the constraint we have:\\
    $$\begin{cases}
        \sum_{k=1}^{K} u_k = 1 \\
        u_k = \frac{\lambda}{2 \V\left(\hat{\tau}_k\right)}
    \end{cases}
    \Rightarrow
    \begin{cases}
        u_k = \frac{\V\left(\hat{\tau}_k\right)^{-1}}{\sum_{k=1}^{K} \V\left(\hat{\tau}_k\right)^{-1}} \\
    \end{cases}$$
    Injecting this result in $\V(\hat{\tau})$, we get:
    \begin{align*}
        \V(\hat{\tau}) &= \sum_{k=1}^{K} u_k^2 \V\left(\hat{\tau}_k\right) \\
        &= \sum_{k=1}^{K} \left(\frac{\V\left(\hat{\tau}_k\right)^{-1}}{\sum_{k=1}^{K} \V\left(\hat{\tau}_k\right)^{-1}}\right)^2 \V\left(\hat{\tau}_k\right) \\
        &= \frac{1}{\sum_{k=1}^{K} \V\left(\hat{\tau}_k\right)^{-1}}
    \end{align*}
    Finally, we get that $\forall k \in [1;K], u_k = \frac{{\V\left(\hat{\tau}_k\right)}^{-1}}{{\sum_{k=1}^{K} \V\left(\hat{\tau}_k\right)}^{-1}}$.\\
    Therefore, $\hat{\tau}_{\text{Meta-IVW}}$ is the minimum-variance unbiased estimator of $\tau$ among the class of aggregation-based estimators.

\subsection{Homogeneous setting}
We prove in this section the results in \cref{sec:homo}, assuming \cref{cond:local_large_sample_size} and the graphical model in Figure~\ref{graph:homog}. 
\subsubsection{Properties of federated outcome estimators}
Proof of \cref{prop:1S_ivw_thetas_equal_pool}: \label{proof:min_var_ivw_thetas}
    \begin{align*}
        \hat\theta_\mathrm{IVW}^{(w)} &= \frac{\sum_{k=1}^K \V(\hat{{\theta}}^{(w)}_k)^{-1} \hat{\theta}_k^{(w)}}{\sum_{k=1}^K \V(\hat{{\theta}}^{(w)}_k)^{-1}} &\\
        &= \frac{\sum_{k=1}^K \left(\frac{1}{\sigma^2} \Xprime_k^{{(w)}^\top}\Xprime_k^{{(w)}}\right) \left({{{X'_k}^{(w)}}}^\top {{X'_k}^{(w)}}\right)^{-1}\Xprime_k^{{(w)}^\top}y_k^{(w)}}{\sum_{k=1}^K \left(\frac{1}{\sigma^2} \Xprime_k^{{(w)}^\top}\Xprime_k^{{(w)}}\right)} &\\
        &= \frac{\sum_{k=1}^K \Xprime_k^{{(w)}^\top} y_k^{(w)}}{\sum_{k=1}^K \Xprime_k^{{(w)}^\top} {{X'_k}^{(w)}}} &\\
        &= \frac{\Xprime_\mathrm{pool}^{{(w)}^\top} y_\mathrm{pool}^{(w)}}{\Xprime_\mathrm{pool}^{{(w)}^\top} \Xprime_\mathrm{pool}^{(w)}} &&\text{by sum of matrix product} &\\
        &= \hat\theta_\mathrm{pool}^{(w)}
    \end{align*}




\subsubsection{Bias of the outcome model estimators}
\label{proof:no_bias_pool}
Unbiasedness of $\hat\theta_\mathrm{pool}$:
\begin{align*}
    \E\left(\hat{\theta}_{\text{pool}}^{(w)}\right) &= \E\left(\left({\Xprime^{(w)}}^\top {\Xprime^{(w)}}\right)^{-1} {\Xprime^{(w)}}^\top y\right)\\
    &= \E\left(\E\left(\left({\Xprime^{(w)}}^\top \Xprime^{(w)}\right)^{-1} {\Xprime^{(w)}}^\top y \mid \Xprime^{(w)}\right)\right) \\
    &= \E\left(\left({\Xprime^{(w)}}^\top \Xprime^{(w)}\right)^{-1} {\Xprime^{(w)}}^\top\E\left(\Xprime^{(w)}\theta + \varepsilon \mid \Xprime^{(w)}\right)\right) \\
    &= \E\left(\left({\Xprime^{(w)}}^\top \Xprime^{(w)}\right)^{-1} {\Xprime^{(w)}}^\top \Xprime^{(w)}\theta + 0 \right) \\
    &= \theta
\end{align*}
Unbiasedness of $\hat\theta_\mathrm{GD}$: under convergence of Algorithm~\ref{alg:FedAvg}, $\hat{\theta}_{\text{GD}}^{(w)} = \hat{\theta}_{\text{pool}}^{(w)}$ which implies $\E\left(\hat{\theta}_{\text{GD}}^{(w)}\right) = \E\left(\hat{\theta}_{\text{pool}}^{(w)}\right) = \theta$.

Unbiasedness of $\hat\theta_\mathrm{1S\text{-}IVW}$: 
\begin{align*}
    \E\left(\hat{\theta}_{\text{IVW}}^{(w)}\right) &= \E\left(\hat{\theta}_{\text{pool}}^{(w)}\right) && \text{using \cref{prop:1S_ivw_thetas_equal_pool}}\\
    &= \theta
\end{align*}
Unbiasedness of $\hat\theta_\mathrm{1S\text{-}SW}$:\\
We condition on the realization of $H$ to account for the variability in the $\{n_k^{(w)}\}_k$ terms.
\begin{align*}
    \E\left(\E\left(\hat{\theta}_\text{1S-SW}^{(w)}\mid H \right)\right) &= \E\left(\E\Big(\sum_{k=1}^K \frac{n_k^{(w)}}{n^{(w)}} \hat{\theta}_k^{(w)}\mid H\Big)\right) \\
    &= \E\left(\sum_{k=1}^K \frac{n_k^{(w)}}{n^{(w)}} \E\left(\hat{\theta}_k^{(w)}\mid H\right)\right)\\
    &= \sum_{k=1}^K \frac{\E(n_k^{(w)})}{n^{(w)}} \theta  \\
    &= \theta
\end{align*}

\subsubsection{(Non-asymptotic) variance comparison of the outcome model parameters}
Variance of the local outcome model OLS estimators:
\begin{align*}
    \V\left(\hat\theta_k^{(w)}\right) &= \E\left(\V\left(\left(\hat\theta_k^{(w)} \mid {{X'_k}^{(w)}}\right)\right)\right) + \V\left(\E\left(\hat\theta_k^{(w)} \mid {{X'_k}^{(w)}}\right)\right) \\
    &= \E\left(\V\left(\left(\Xprime_k^{(w)\top} {{X'_k}^{(w)}}\right)^{-1} \Xprime_k^{(w)\top} y_k^{(w)} \mid {{X'_k}^{(w)}}\right)\right) + 0 \\
    &= \E\left(\left(\Xprime_k^{(w)\top} {{X'_k}^{(w)}}\right)^{-1} \Xprime_k^{(w)\top} \V\left(y_k^{(w)} \mid {{X'_k}^{(w)}}\right) {{X'_k}^{(w)}} \left(\Xprime_k^{(w)\top} {{X'_k}^{(w)}}\right)^{-1}\right) \\
    &= \E\left(\left(\Xprime_k^{(w)\top} {{X'_k}^{(w)}}\right)^{-1} \Xprime_k^{(w)\top} \sigma^2 I_p {{X'_k}^{(w)}} \left(\Xprime_k^{(w)\top} {{X'_k}^{(w)}}\right)^{-1}\right) \\
    &= \sigma^2 \E\left(\left(\Xprime_k^{(w)\top} {{X'_k}^{(w)}}\right)^{-1}\right)
\end{align*}

Similarly we get $\V\left(\hat\theta_\mathrm{IVW}^{(w)}\right) = \V\left(\hat\theta_\mathrm{pool}^{(w)}\right) = \V\left(\hat\theta_\mathrm{GD}^{(w)}\right) = \sigma^2 \E\left(\left(\Xprime^{(w)\top} {{\Xprime^{(w)}}}\right)^{-1}\right)$.

The variance of the One-Shot SW outcome model parameters estimator is obtained using the law of total variance over $H$:
\begin{align*}
    \V\left(\hat\theta_\mathrm{SW}^{(w)}\right) &= \E\left(\V(\hat\theta_\mathrm{SW}^{(w)}\mid H)\right) + \V\left(\E(\hat\theta_\mathrm{SW}^{(w)}\mid H)\right)\\
    &= \sum_{k=1}^K \E\left(\frac{n_k^2}{n^2}\right) \V\left(\hat\theta_k^{(w)}\right) + 0 \\
    &= \sigma^2 \sum_{k=1}^K \frac{\E(n_k^2)}{n^2} \E\left(\Xprime_k^{(w)\top} {{X'_k}^{(w)}}^{-1}\right)
\end{align*}
 by independence of the studies for the last equality.

We now compare the variances above. First notice that:
\begin{align*}
    \E\left(\left(\Xprime_\mathrm{pool}^{(w)\top} {{X^{(w)}_\mathrm{pool}}}\right)^{-1}\right) = \frac{1}{n} \E\left(\frac{1}{\frac{1}{n}\Xprime_\mathrm{pool}^{(w)\top} {{X^{(w)}_\mathrm{pool}}}}\right) = \frac{1}{n} \E\left(\frac{1}{\sum_{k=1}^K \frac{\E(n_k)}{n} \left(\frac{1}{\E(n_k)}\Xprime_k^{(w)\top} {{X^{(w)}_k}}\right)}\right)
\end{align*}
By applying Jensen's inequality on the inverse function over the space of semi positive definite matrices and with weights summing to 1 $\{\E(n_k)/n\}_{k\in\llbracket1,K\rrbracket}$:
\begin{align*}
    \E\left(\left(\Xprime_\mathrm{pool}^{(w)\top} {{X^{(w)}_\mathrm{pool}}}\right)^{-1}\right) &\preceq \frac{1}{n} \sum_{k=1}^K \frac{\E(n_k)}{n} \E\left(\left(\frac{1}{\E(n_k)} \Xprime_k^{(w)\top} {{X^{(w)}_k}}\right)^{-1}\right)\\
    &\preceq \sum_{k=1}^K \frac{\E(n_k)^2}{n^2} \E\left(\left(\Xprime_k^{(w)\top} {{X'_k}^{(w)}}\right)^{-1}\right) \\
    &\preceq \sum_{k=1}^K \frac{\E(n_k^2)}{n^2} \E\left(\left(\Xprime_k^{(w)\top} {{X'_k}^{(w)}}\right)^{-1}\right) && \text{as } \E(n_k)^2 \leq \E(n_k^2)
\end{align*}
which leads to the conclusion of \cref{prop:1s_diff_means_var}:         $\V(\hat\theta_\mathrm{pool}^{(w)})= \V(\hat\theta_\mathrm{GD}^{(w)}) = \V(\hat\theta_\mathrm{IVW}^{(w)}) \preceq \V(\hat\theta_\mathrm{SW}^{(w)})$.

\subsubsection{Asymptotic variances of the outcome model parameters}\label{proof:asymp_variances_thetas}
Asymptotically, we have
\begin{align*}
    {{X'_k}^{(w)}}^\top {{X'_k}^{(w)}} &= \frac{1}{n_k^{(w)}} \sum_{i=1}^{n_k^{(w)}} \Xprime_{k,i}^{(w)\top} \Xprime_{k,i}^{(w)} \\
    &\xrightarrow{n_k^{(w)} \rightarrow \infty} A_k \\
    \left({{X'_k}^{(w)}}^\top {{X'_k}^{(w)}} \right)^{-1} &\rightarrow A_k^{-1} \quad \text{by continuous mapping }
\end{align*}
where $A_k = \begin{pmatrix}
        1 & \mu_{k,1} & \dots & \mu_{k,p} \\
        \mu_{k,1} & \\
        \vdots & & \Sigma_k \\
        \mu_{k,p} &
    \end{pmatrix}$ and $
    \mu_{k,j}$ is the mean of the j-th covariate and $\Sigma_k = \E\left({X_k^{(w)}}^\top {X_k^{(w)}}\right)$ is the covariate matrix in study $k$.

Therefore we get the asymptotic variance of the local outcome model parameters:
\begin{align*}
    \aVar\left(\hat\theta_k^{(w)}\right) &= \frac{\sigma^2}{n_k^{(w)}} A_k^{-1}\\
    \aVar\left(\hat\beta_k^{(w)}\right) &= \frac{\sigma^2}{n_k^{(w)}} \Sigma_k^{-1}
\end{align*}

Under the graphical model in Figure~\ref{graph:homog}, $\Sigma_k= \Sigma$ and $A_k= A$ which yields:

\begin{align*}
    \text{Local OLS} & \quad \aVar\left(\hat \theta_k^{(w)}\right) = \sigma^2 \left({X'_k}^{{(w)}^\top} {X'_k}^{(w)}\right)^{-1} = \frac{\sigma^2}{n_k^{(w)}} A^{-1} &\\
    \text{Pool OLS} & \quad \aVar\left(\hat \theta_{\text{pool}}^{(w)}\right) = \sigma^2 \left(X'^{{(w)}^\top} X'^{(w)}\right)^{-1} = \frac{\sigma^2}{n^{(w)}} A^{-1} &\\
    \text{GD} & \quad \aVar\left(\hat \theta_{\text{GD}}^{(w)}\right) = \sigma^2 \left({X'}^{{(w)}^\top} {X'}^{(w)}\right)^{-1} = \frac{\sigma^2}{n^{(w)}} A^{-1} &\\
    \text{1S\text{-}SW} & \quad \aVar\left(\hat \theta_{\text{SW}}^{(w)}\right) = \E\left(\aVar\left(\sum_{k=1}^K \frac{n_k^{(w)}}{n^{(w)}} \hat \theta_k^{(w)}\mid H\right)\right) + 0 \\
    &\quad \quad \quad \quad\quad= \E\left(\sum_{k=1}^K \frac{{n_k^{(w)}}^2}{{n^{(w)}}^2} \aVar\left(\hat \theta_k^{(w)}\mid H \right)\right)\\
    &\quad \quad \quad \quad\quad= \E\left(\sum_{k=1}^K \left(\frac{n_k^{(w)}}{n^{(w)}}\right)^2 \frac{\sigma^2}{n_k^{(w)}} A^{-1}\right) \\
    &\quad \quad \quad \quad\quad= \frac{\sigma^2}{n^{{(w)}^2}} A^{-1} \sum_{k=1}^{K} \E(n_k^{(w)})\\
    &\quad \quad \quad \quad\quad= \frac{\sigma^2}{n^{(w)}} A^{-1}\\
    \text{1S\text{-}IVW} & \quad \aVar\left(\hat \theta_{\text{IVW}}^{(w)}\right) = \aVar\left(\frac{\sum_{k=1}^K \left(\aVar(\hat{{ \theta}}^{(w)}_k)^{-1} \hat \theta_k^{(w)}\right)}{\sum_{k=1}^K \aVar(\hat{{ \theta}}^{(w)}_k)^{-1}}\right)\\
    &\quad \quad \quad \quad\quad= \V\left(\frac{\sum_{k=1}^K \frac{1}{\sigma^2} {X'_k}^{(w)\top} {X'_k}^{(w)} \times \left({X'_k}^{(w)\top} {X'_k}^{(w)}\right)^{-1} {X'_k}^{(w)\top} y_k^{(w)}}{\sum_{k=1}^K \frac{1}{\sigma^2} {X'_k}^{(w)\top} {X'_k}^{(w)}}\right)\\
    &\quad \quad \quad \quad\quad= \V\left(\frac{\sum_{k=1}^K {X'_k}^{(w)\top} y_k^{(w)}}{\sum_{k=1}^K {X'_k}^{(w)\top} {X'_k}^{(w)}}\right)\\
    &\quad \quad \quad \quad\quad= \frac{\sum_{k=1}^{K} {X'_k}^{(w)\top} \V(y_k^{(w)}) {X'_k}^{(w)}}{\left(\sum_{k=1}^K {X'_k}^{(w)\top} {X'_k}^{(w)}\right)^2}\\
    &\quad \quad \quad \quad\quad= \frac{\sigma^2}{\sum_{k=1}^K {X'_k}^{(w)\top} {X'_k}^{(w)}}\\
    &\quad \quad \quad \quad\quad= \frac{\sigma^2}{n^{(w)}} A^{-1} &
\end{align*}

So under this model, for $w \in \{0, 1\}$, $\aVar\left(\hat \theta_{\text{GD}}^{(w)}\right) = \aVar\left(\hat \theta_{\text{SW}}^{(w)}\right) = \aVar\left(\hat \theta_{\text{IVW}}^{(w)}\right) = \frac{\sigma^2}{n^{(w)}} A^{-1}$. These results, along with the associated communication costs, are summarized in Table~\ref{table:summaryoutcomeparametersRCT}.
\begin{table*}[!h]
    \footnotesize
    \centering
        \begin{tabular}{l l l l l l}
        \toprule
        Estimator &  Notation & Condition & $\aVar$ &  Com. rounds & Com. cost \\
        \midrule
        {Local} & $\hat{\theta}_k^{(w)}$ & Condition~\ref{cond:local_large_sample_size} & $\frac{\sigma^2}{n^{(w)}_k} A^{-1}$ &  0 & 0 \\
        \midrule
        {One-Shot SW} & $\hat{\theta}_{\text{1S-SW}}^{(w)}$ (Eq.~\ref{eq:theta_1s_sw}) & Condition~\ref{cond:local_large_sample_size} & $\frac{\sigma^2}{n^{(w)}} A^{-1}$ & 1 & $O(d+1)$ \\
        {One-Shot IVW} & $\hat{\theta}_{\text{1S-IVW}}^{(w)}$ (Eq.~\ref{eq:theta_1s_ivw})&  Condition~\ref{cond:local_large_sample_size} & $\frac{\sigma^2}{n^{(w)}} A^{-1}$ & 1 & $O(d^2)$\\
        {GD-Federated} & $\hat{\theta}_{\text{GD}}^{(w)}$ (Alg.~\ref{alg:FedAvg}) & Condition~\ref{cond:federated_large_sample_size} & $\frac{\sigma^2}{n^{(w)}} A^{-1}$ & $T$ & $O(Td)$\\
        \midrule        
        {Pool} & $\hat{\theta}_\text{pool}^{(w)}$ & Condition~\ref{cond:federated_large_sample_size} & $\frac{\sigma^2}{n^{(w)}} A^{-1}$ &  --- & --- \\
        \bottomrule
        \end{tabular}
        \caption{Properties of the (unbiased) estimators of the outcome model parameters in the homogeneous setting: asymptotic variance, number of communication rounds and total communication cost (in number of floats per study).}
        \label{table:summaryoutcomeparametersRCT}
\end{table*}

\subsubsection{Bias of the ATE estimators}\label{proof:no_bias_estimators}
We now prove that under the graphical model in Figure~\ref{graph:homog} all the estimators are unbiased.

First, notice that for any random variable $U\in \mathbb{R}^{n\times d+1}, w\in\{0,1\}$, 
\begin{align*}
    \E(U\hat\theta^{(w)}) &= \E\left(U ({{X'}^{(w)}}^\top {{X'}^{(w)}})^{-1} {{X'}^{(w)}} Y^{(w)}\right)\\
    &=\E\left(U ({{X'}^{(w)}}^\top {{X'}^{(w)}})^{-1} {{X'}^{(w)}} ({{X'}^{(w)}} \theta^{(w)} + \varepsilon)\right) \\
    &= \E(U)\theta^{(w)}
\end{align*}
Then, 
\begin{align*}
    \E(\hat\tau_k) &= \E\left(\frac{1}{n_k} \sum_{i=1}^{n_k}\left(X'_{k,i}\hat{\theta}_k^{(1)} - X'_{k,i}\hat{\theta}_k^{(0)}\right)\right) && \text{Defined in \eqref{eq:tauk}}\\
    &= \frac{1}{n_k} \sum_{i=1}^{n_k}\left(\E\left(X'_{k,i}\hat{\theta}_k^{(1)}\right) - \E\left(X'_{k,i}\hat{\theta}_k^{(0)}\right)\right) \\
    &= \frac{1}{n_k} \sum_{i=1}^{n_k}\left(\E\left(X'_{k,i}\right)\left({\theta}^{(1)} - {\theta}^{(0)}\right)\right)\\
    &=\E\left(X'_{i}\right)\left({\theta}^{(1)} - {\theta}^{(0)}\right) = \tau && \text{Defined in \eqref{eq:tau}}
\end{align*}
Similarly, conditioning on $H$ to account for the variability in the $n_k$ random (binomial) terms:
\begin{align*}
    \E(\hat\tau_\mathrm{pool}\mid H) &= \E\left(\frac{1}{n} \sum_{i=1}^{n}\left(X'_i\hat\theta_{\mathrm{pool}}^{(1)} - X'_i\hat\theta_{\mathrm{pool}} \right)\mid H\right) && \text{Defined in \eqref{eq:pool}}\\
    &= \frac{1}{n} \sum_{i=1}^{n}\left(\E\left(X'_{i}\hat{\theta}_{\mathrm{pool}}^{(1)}\mid H\right) - \E\left(X'_{i}\hat{\theta}_{\mathrm{pool}}^{(0)}\mid H\right)\right) \\
    &= \frac{1}{n} \sum_{k=1}^{K} n_k \frac{1}{n_k} \sum_{i=1}^{n_k} (\E(X'_{k,i}\hat\theta^{(1)}_\mathrm{1S\text{-}IVW}\mid H) - \E(X'_{k,i}\hat\theta^{(0)}_\mathrm{1S\text{-}IVW}\mid H)) = \E(\hat\tau_\mathrm{1S\text{-}IVW}\mid H) && \text{Def.~\ref{def:1SIVWSS} +Th.~ \ref{prop:1S_ivw_thetas_equal_pool}}\\
    &= \frac{1}{n} \sum_{k=1}^{K} n_k \frac{1}{n_k} \sum_{i=1}^{n_k} (\E(X'_{k,i}\hat\theta^{(1)}_\mathrm{GD}\mid H) - \E(X'_{k,i}\hat\theta^{(0)}_\mathrm{GD}\mid H)) = \E(\hat\tau_\mathrm{GD}\mid H) && \text{\cref{def:GDSW}}\\
    &=\E\left(X'_{i}\mid H\right)\left({\theta}^{(1)} - {\theta}^{(0)}\right) = \tau
\end{align*}
And,
\begin{align*}
    \E(\hat\tau_\mathrm{1S\text{-}SW}\mid H) &= \frac{1}{n} \sum_{k=1}^{K} n_k \frac{1}{n_k} \sum_{i=1}^{n_k} (\E(X'_{k,i}\hat\theta^{(1)}_\mathrm{1S-SW}) - \E(X'_{k,i}\hat\theta^{(0)}_\mathrm{1S-SW})) && \text{\cref{def:1SSSSS}} \\
    &= \frac{1}{n} \sum_{i=1}^{n} \left(\E\left(X'_{k,i}\sum_{l=1}^K \frac{n_l}{n}\hat\theta^{(1)}_l\right) - \E\left(X'_{k,i}\sum_{l=1}^K \frac{n_l}{n}\hat\theta^{(0)}_l\right)\right)\\
    &= \frac{1}{n} \sum_{i=1}^{n} \left(\sum_{l=1}^K \frac{n_l}{n} \E\left(X'_{k,i}\hat\theta^{(1)}_l\mid H\right) - \sum_{l=1}^K \frac{n_l}{n} \E\left(X'_{k,i} \hat\theta^{(0)}_l\mid H\right)\right) \\
    &= \frac{1}{n} \sum_{i=1}^{n} \left(\sum_{l=1}^K \frac{n_l}{n} \E\left(X'_{k,i}\mid H\right)(\theta^{(1)} - \theta^{(0)})\mid H\right) \\
    &= \E\left(X'_{i}\mid H\right)(\theta^{(1)} - \theta^{(0)}) = \tau
\end{align*}
Noticing that none of the expectations above depend on the $n_k$ and that $X\indep H$, we can remove the conditioning over $H$, so $\E(\hat\tau_\mathrm{pool})=\E(\hat\tau_\mathrm{GD})=\E(\hat\tau_\mathrm{1S\text{-}IVW})=\E(\hat\tau_\mathrm{1S\text{-}SW})=\tau$.

\subsubsection{Asymptotic variances of the ATE estimators}\label{proof:avar_ate_random_design_v1}
We recall that $\rho_k = \Pb(H_i=k)=\E\left(\frac{n_k}{n}\right)$ and that $p = \Pb(W_i=1) =\sum_{k=1}^K \Pb(H_i=k) \Pb(W_i=1|H_i=k) = \sum_{k=1}^K \rho_k p_k$.

\vspace{2em}
\textbf{Proof of \cref{prop:sw_ivw_same_weights} }\label{proof:sw_ivw_same_weights}: we prove that the SW ($\omega_k^\mathrm{SW} = \frac{n_k}{\sum_{k=1}^{K}n_k}$) and IVW ($\omega_k^\mathrm{IVW} = \frac{{\aVar(\hat\tau_k^\mathrm{fed})}^{-1}}{\sum_{k=1}^{K}{\aVar(\hat\tau_k^\mathrm{fed})}^{-1}}$) weights are asymptotically equivalent for federated local estimators ($\hat\tau_k^\mathrm{1S\text{-}SW}$, $\hat\tau_k^\mathrm{1S\text{-}IVW}$ both defined in Eq.~\ref{eq:theta_1S}, and  $\hat\tau_k^\mathrm{GD}$ defined in Eq.~\ref{eq:tauGDk}). We denote by $\hat\tau_k^\mathrm{fed}$ any of these estimators.

Then:
\begin{align*}
    \omega_k^\mathrm{IVW} &= \frac{\left(\frac{\sigma^2}{n_k}\left(\frac{n}{n^{(1)}} + \frac{n}{n^{(0)}}\right) + \frac{1}{n_k}\Vert \beta^{(1)} - \beta^{(0)} \Vert^2_\Sigma\right)^{-1}}{\sum_{k=1}^{K}\left({\frac{\sigma^2}{n_k}\left(\frac{n}{n^{(1)}} + \frac{n}{n^{(0)}}\right) + \frac{1}{n_k}\Vert \beta^{(1)} - \beta^{(0)} \Vert^2_\Sigma}\right)^{-1}}\\
    &= \frac{n_k\left(\sigma^2\left(\frac{n}{n^{(1)}} + \frac{n}{n^{(0)}}\right) + \Vert \beta^{(1)} - \beta^{(0)} \Vert^2_\Sigma\right)^{-1}}{\sum_{k=1}^{K} n_k\left(\sigma^2\left(\frac{n}{n^{(1)}} + \frac{n}{n^{(0)}}\right) + \Vert \beta^{(1)} - \beta^{(0)} \Vert^2_\Sigma\right)^{-1}}\\
    &= \frac{n_k}{\sum_{k=1}^{K} n_k}\\
    &= \omega_k^\mathrm{SW}
\end{align*}
\vspace*{-1em}
Therefore, asymptotically, aggregating the local federated estimates of the ATE with SW or IVW is the same. 

\vspace*{2em}
\textbf{Asymptotic Variance of local ATE estimator.}

First, recall from \cref{proof:asymp_variances_thetas} that if $\E(X^\top \varepsilon\mid H) = 0$ and \cref{cond:local_large_sample_size} holds, the OLS estimator $\hat\theta_k^{(w)}$ is consistent and asymptotically normal:
    \begin{align}
        \hat\theta_k^{(w)} &\xrightarrow{P} \theta_k^{(w)} \nonumber\\
        \sqrt{n_k^{(w)}} (\hat\theta_k^{(w)} - \theta_k^{(w)}) &\xrightarrow{d} \mathcal{N}(0, \sigma^2 A^{-1})
    \end{align}
which gives:
\begin{align}
    \sqrt{n_k^{(w)}} (\hat c_k^{(w)} - c_k^{(w)}) &\xrightarrow{d} \mathcal{N}(0, \sigma^2)\label{eq:1}\\
    \sqrt{n_k^{(w)}} (\hat\beta_k^{(w)} - \beta_k^{(w)}) &\xrightarrow{d} \mathcal{N}(0, \sigma^2 \Sigma_k^{-1}) \label{eq:2}
\end{align}
In particular, we have that $\hat c_k^{(1)}, \hat c_k^{(0)}, \hat \beta_k^{(1)},\hat \beta_k^{(0)}, \overline{X_k} = \frac{1}{n_k} \sum_{i=1}^{n_k} X_{k,i}$ are all asymptotically independent.

Then we have that:
\begin{align}\label{eq:form_ols}
    \hat \tau_{k} - \tau &= \hat c_k^{(1)} - c^{(1)} - (\hat c_k^{(0)} - c^{(0)}) + \overline{X_k} (\hat\beta_k^{(1)} - \hat\beta^{(0)}) - \E(X_k)(\beta_k^{(1)} - \beta^{(0)}) \nonumber\\
    &= \hat c_k^{(1)} - c^{(1)} - (\hat c_k^{(0)} - c^{(0)}) + \overline{X_k} \left((\hat\beta_k^{(1)} - \beta^{(1)}) - (\hat\beta_k^{(0)} - \beta^{(0)})\right) \nonumber\\
    & \quad \quad - \E(X_k)(\beta_k^{(1)} - \beta^{(0)}) - \overline{X_k} (\beta_k^{(1)} - \beta^{(0)})\nonumber\\
    &= \underbrace{\hat c_k^{(1)} - c^{(1)}}_{A_1} - (\underbrace{\hat c_k^{(0)} - c^{(0)}}_{A_0}) + \underbrace{\overline{X_k} \left((\hat\beta_k^{(1)} - \beta^{(1)}) - (\hat\beta_k^{(0)} - \beta^{(0)})\right)}_{B} \nonumber\\
    & \quad \quad +\underbrace{(\overline{X_k} - \E(X_k))(\beta^{(1)} - \beta^{(0)})}_{C}
\end{align}

\begin{itemize}
    \item For $w \in \{0, 1\}$, $A_w \xrightarrow{d} \mathcal{N}(0, \frac{\sigma^2}{n_k^{(w)}})$ from \cref{eq:1}
    \item Let $M > 0$ be a real number. Then, \begin{align*}
        \Pb\left(\left|n_k B\right| > M\right) &= \Pb\Bigl(|\underbrace{\root \of n_k \left(\overline X_k - \E(X_k)\right)}_{\xrightarrow{d} \mathcal{N}(0, \Sigma_k)} (\underbrace{\root\of n_k(\hat\beta_k^{(1)} - \beta_k^{(1)})}_{\xrightarrow{d} \mathcal{N}(0, \frac{\sigma^2}{n_k^{(1)}})} - \underbrace{\sqrt{n_k}(\hat\beta_k^{(0)} + \beta^{(0)})}_{\xrightarrow{d} \mathcal{N}(0, \frac{\sigma^2}{n_k^{(0)}})})| > M\Bigr) \xrightarrow{n_k \to \infty} 0
    \end{align*}
    so that $\Pb(\left|\frac{B}{1/n_k} - 0 \right|>M) < \varepsilon $, meaning that $B = \mathcal{O}_P(1/n_k)$ by definition.
    \item $C$: from the central limit theorem, we have that $\sqrt{n_k} \left(\overline X_k - \E(X_k)\right) \xrightarrow{d} \mathcal{N}(0, \Sigma_k)$ and $\beta_k^{(1)} - \beta_k^{(0)}$ is a constant vector. However, for any p-multivariate $\sqrt{n} Z \sim \mathcal{N}(0, \Sigma_k)$ random variable and $D$ a constant vector of size $p\times 1$, we have $\root \of n Z D \sim \mathcal{N}(0, D^\top \Sigma D)$.\\
    Therefore, $\sqrt{n_k} C \xrightarrow{d} \mathcal{N}(0, \Vert \beta^{(1)} - \beta^{(0)}\vert^2_{\Sigma_k})$.
\end{itemize}

Finally, we have that:
\begin{align}
    \aVar(\hat\tau_k) &= \frac{1}{n_k} \aVar(\sqrt{n_k} (\hat\tau_k - \tau)) \nonumber \\
    &= \frac{1}{n_k} \left(\aVar(\sqrt{n_k} A_1) + \aVar(\sqrt{n_k} A_0) + \aVar(\sqrt{n_k} B) + \aVar(\sqrt{n_k} C)\right)\nonumber\\
    &= \frac{1}{n_k} \left(\frac{n_k}{n_k^{(1)}} \sigma^2 + \frac{n_k}{n_k^{(0)}} \sigma^2 + 0 + \Vert \beta^{(1)} - \beta^{(0)} \Vert^2_{\Sigma_k}\right)\nonumber\\
    &= \sigma^2 \left(\frac{1}{n_k^{(1)}} + \frac{1}{n_k^{(0)}}\right) + \frac{1}{n_k} \Vert \beta^{(1)} - \beta^{(0)} \Vert^2_{\Sigma_k}\label{eq:local_avar_w_nk} \\
    &= \frac{\sigma^2}{n_k} \left(\frac{1}{p_k} + \frac{1}{1-p_k}\right) + \frac{1}{n_k} \Vert \beta^{(1)} - \beta^{(0)} \Vert^2_{\Sigma_k} \nonumber
\end{align}
with $p_k=\Pb(W_i=1|H_i=k)$.

Using Central Limit Theorem: 
\begin{align}
    \boxed{\sqrt{n_k} (\hat{\tau}_k - \tau) \xrightarrow{d} \mathcal{N}\left(0, \frac{\sigma^2}{p_k(1-p_k)} + \Vert \beta^{(1)} - \beta^{(0)} \Vert^2_{{\Sigma_k}}\right)} \label{eq:avar_local}
\end{align}

\vspace*{2em}
\textbf{Proof of \cref{tab:taus_var}}:

From \cref{eq:avar_local} we have that in a Bernoulli trial (and denoting $[H=k]=\{H_i=k\}_{i=1}^n$):
\begin{align*}
    \textbf{Local ATE aVar} & \quad \aVar \left(\hat\tau_k|H=k\right) = \frac{\sigma^2}{n_k p_k (1-p_k)} + \frac{1}{n_k} \Vert \beta^{(1)} - \beta^{(0)} \Vert^2_{\Sigma_k}
\end{align*}

Let's apply this result to the pooled dataset $\mathcal{Z}$ and considering it a Bernoulli trial with treatment probability $p$, and denoting $H=\{H_i\}_{i=1}^n$: 
\begin{align*}
    \aVar \left(\hat\tau_{\mathrm{pool}}|H\right) = \frac{\sigma^2}{n p(1-p)} + \frac{1}{n} \Vert \beta^{(1)} - \beta^{(0)} \Vert^2_\Sigma \\
\end{align*}
Finally, because $H$ is not associated with the outcome, we have that $\aVar \left(\hat\tau_{\mathrm{pool}}\right) = \aVar \left(\hat\tau_{\mathrm{pool}}|H\right)$, which allows us to conclude:

\begin{align*}
    \textbf{Pooled ATE aVar} & \quad \aVar \left(\hat\tau_{\mathrm{pool}}\right) = \frac{\sigma^2}{n p(1-p)} + \frac{1}{n} \Vert \beta^{(1)} - \beta^{(0)} \Vert^2_\Sigma 
\end{align*}\label{proof:asymp_ols}
which proves Eq.~\ref{eq:asymp_ols} (even in the non-centered covariates case).


To compute the asymptotic variance of the local federated outcome parameters ATE estimator, we first compute the asymptotic variance of the federated-outcome model parameters estimated individual treatment effect $\hat\tau_{k,i}^\mathrm{fed}=X'_{k,i} \hat\theta^{(1)}_\mathrm{fed} - X'_{k,i} \hat\theta^{(0)}_\mathrm{fed}$. To do this, remark that:
\begin{align*}
    \aVar(\hat\tau_{\mathrm{pool}}) &= \aVar\left(\frac{1}{n} \sum_{k=1}^{K}\sum_{i=1}^{n_k} \hat\tau_{k,i}^\mathrm{pool}\right)\\ 
    &= \aVar\left(\frac{1}{n} \sum_{k=1}^{K}\sum_{i=1}^{n_k} \hat\tau_{k,i}^\mathrm{fed}\right) &&\text{since } \hat\theta_\mathrm{fed}^{{(w)}}\to \hat\theta_\mathrm{pool}^{{(w)}}\\
    &= \frac{1}{n^2} \sum_{k=1}^{K}\sum_{i=1}^{n_k} \aVar(\hat\tau_{k,i}^\mathrm{fed}) + \frac{1}{n^2} \sum_{k=1}^{K}\sum_{l=1}^{K} \sum_{i=1}^{n_k} \sum_{j \neq i}^{n_l} \aCov{\hat\tau_{k,i}^\mathrm{fed}}{\hat\tau_{l,j}^\mathrm{fed}}
\end{align*}
Let us show that the $\text{Cov}(\hat\tau_{k,i}^\mathrm{fed},\hat\tau_{l,j}^\mathrm{fed})$ are asymptotically null for all $(k,i) \neq (l,j)$. Denote by $\aCov{a}{b}$ the asymptotic covariance of $a$ and $b$ random variables. We further write $\hat Y_i^{(w)} = X'_i \hat\theta^{(w)}_\mathrm{fed}$, and remark that asymptotically, $\hat Y_i^{(w)} = X'_i \hat\theta^{(w)}_\mathrm{pool}$, then:
\begin{align*}
    \aCov{\hat\tau_{k,i}^\mathrm{fed}}{\hat\tau_{l,j}^\mathrm{fed}} &= \aCov{X'_{k,i} \left(\hat\theta_{\text{fed}}^{(1)} - \hat\theta_{\text{fed}}^{(0)}\right)}{X'_{l,j} \left(\hat\theta_{\text{fed}}^{(1)} - \hat\theta_{\text{fed}}^{(0)}\right)} \\
    &= \aCov{\hat Y_{k,i}(1) - \hat Y_{k,i}(0)}{\hat Y_{l,j}(1) - \hat Y_{l,j}(0)}\\
    &= {\aCov{\hat Y_{k,i}(1)}{\hat Y_{l,j}(1)}} - \aCov{\hat Y_{k,i}(1)}{\hat Y_{l,j}(0)} \\
    & \quad \quad - \aCov{\hat Y_{k,i}(0)}{\hat Y_{l,j}(1)} + \aCov{\hat Y_{k,i}(0)}{\hat Y_{l,j}(0)}\\
    \aCov{\hat Y_{k,i}^{(w_a)}}{\hat Y_{l,j}^{(w_b)}} &= \E\left[ \left(\hat Y_{k,i}^{(w_a)} - \E(\hat Y_{k,i}^{(w_a)})\right) \left(\hat Y_{l,j}^{(w_b)} - \E(\hat Y_{l,j}^{(w_b)})\right) \right] && \forall w_a, w_b \in \{0, 1\}\\
    &= \E\left[ \left(\hat Y_{k,i}^{(w_a)} - Y_{k,i}^{(w_a)}\right) \left(\hat Y_{l,j}^{(w_b)} - Y_{l,j}^{(w_b)}\right) \right] && \text{unbiased estimators} \\
    &= \E\left[ \varepsilon_{k,i}^{(w_a)} \varepsilon_{l,j}^{(w_b)} \right] && \text{residuals} \\
    &= \E\left[ \varepsilon_{k,i}^{(w_a)}\right] \E\left[ \varepsilon_{l,j}^{(w_b)} \right] && \text{independent errors} \\
    &= 0 && \text{centered noise}
\end{align*}
So that $\aCov{\hat\tau_{i}^\mathrm{fed}}{\hat\tau_{j}^\mathrm{fed}|H_i, H_j} = 0$, which yields that $\aVar(\hat\tau_{\mathrm{pool}}) = \frac{1}{n^2} \sum_{k=1}^{K}\sum_{i=1}^{n_k} \aVar(\hat\tau_{k,i}^\mathrm{fed}|H_i=k)$. Finally, since the individuals follow the same distribution across studies within the graphical model in Figure~\ref{graph:homog}, the $\hat\tau_{k,i}^\mathrm{fed}$ are \textit{i.i.d.} across studies, so that their asymptotic variances are equal:
\begin{align*}
    \aVar(\hat\tau_{k,i}^\mathrm{fed}|H_i=k) = n \aVar(\hat\tau_{\mathrm{pool}}) = \sigma^2 \left(\frac{1}{p(1-p)}\right) + \Vert \beta^{(1)} - \beta^{(0)} \Vert^2_\Sigma
\end{align*}

Therefore,
\begin{align*}
    \textbf{Federated Local ATE} & \quad \aVar(\hat\tau_k^\mathrm{fed}|H=k) = \aVar(\frac{1}{n_k} \sum_{i=1}^{n_k} \hat\tau_{k,i}^\mathrm{fed}|H=k) \\
    &= \frac{1}{n_k^2} \sum_{i=1}^{n_k} \aVar(\hat\tau_{k,i}^\mathrm{fed}|H=k) + \frac{1}{n_k^2} \sum_{i=1}^{n_k} \sum_{j \neq i} \underbrace{\aCov{\hat\tau_{k,i}^\mathrm{fed}}{\hat\tau_{k,j}^\mathrm{fed}|H=k}}_{=0}\\
    &= \frac{n}{n_k} \aVar(\hat\tau_{\mathrm{pool}}) \\
    &= \frac{\sigma^2}{n_k} \left(\frac{1}{p(1-p)}\right) + \frac{1}{n_k}\Vert \beta^{(1)} - \beta^{(0)} \Vert^2_{\Sigma}
\end{align*}

For the Meta estimators we use the law of total variance:
\begin{align*}
    \textbf{Meta-SW ATE} & \quad \aVar(\hat\tau_{\text{Meta-SW}}) = 
    \aVar(\sum_{k=1}^K \frac{n_k}{n} \hat\tau_k|H) \\
    &= \E\left(\sum_{k=1}^K \left(\frac{n_k}{n}\right)^2 \aVar(\hat\tau_k|H)\right) + \aVar\left(\E\left(\sum_{k=1}^K \frac{n_k}{n} \hat\tau_k|H\right)\right)\\
    &= \sum_{k=1}^K \E\left[\left(\frac{n_k}{n}\right)^2 \left(\frac{\sigma^2}{n_k p_k (1-p_k)} + \frac{1}{n_k} \Vert \beta^{(1)} - \beta^{(0)} \Vert^2_\Sigma\right)\right] +0\\
    &= \frac{\sigma^2}{n} \sum_{k=1}^{K} \E\left[\frac{n_k}{n}\right]  \frac{1}{p_k(1-p_k)} + \frac{1}{n} \Vert \beta^{(1)} - \beta^{(0)} \Vert^2_\Sigma\\
    &= \frac{\sigma^2}{n} \sum_{k=1}^{K} \frac{\rho_k}{p_k(1-p_k)} + \frac{1}{n} \Vert \beta^{(1)} - \beta^{(0)} \Vert^2_\Sigma\\
\end{align*}

\begin{align*}
    \textbf{Meta\text{-}IVW ATE} & \quad \aVar(\hat\tau_{\mathrm{Meta\text{-}IVW}}) = \E\left(\aVar\left(\frac{\sum_{k=1}^K \left(\aVar(\hat{{\tau}}_k|H=k)^{-1} \hat{\tau}_k\right)}{\sum_{k=1}^K \aVar(\hat{{\tau}}_k|H=k)^{-1}}|H\right)\right) + 0 \\
    &=  \E\left(\frac{\sum_{k=1}^{K} \aVar(\hat{{\tau}}_k|H=k)^{-2} \aVar(\hat{{\tau}}_k|H=k)}{\left(\sum_{k=1}^K \aVar(\hat{{\tau}}_k|H=k)^{-1}\right)^2}\right) && \hat\tau_k \indep \hat\tau_l\\
    &=  \E\left(\left(\sum_{k=1}^K \aVar(\hat{{\tau}}_k|H=k)^{-1}\right)^{-1}\right)\\
    &=  \E\left(\left(\sum_{k=1}^K \left(\frac{\sigma^2}{n_k p_k (1-p_k)} + \frac{1}{n_k} \Vert \beta^{(1)} - \beta^{(0)} \Vert^2_\Sigma\right)^{-1}\right)^{-1}\right)\\
    &= \mathbb{E}\left(\left(\sum_{k=1}^K \frac{n_k}{\frac{\sigma^2}{p_k (1-p_k)} + \|\beta^{(1)} - \beta^{(0)}\|^2_{\Sigma}}\right)^{-1}\right) \\
    &= \frac{1}{n} \left(\sum_{k=1}^{K} \frac{\rho_k}{\frac{\sigma^2}{p_k(1-p_k)} + \Vert \beta^{(1)} - \beta^{(0)} \Vert^2_\Sigma}\right)^{-1} && \text{(LLN)}\\
\end{align*}
where (LLN) refers to the law of large numbers, stating that given $n_k \sim \text{Binomial}(n, \rho_k)$ and $\mathbb{E}[n_k] = n \rho_k$, for large $n$, we have $\mathbb{E}\left(\frac{1}{n_k}\right) \approx \frac{1}{\mathbb{E}[n_k]} = \frac{1}{n \rho_k}$.


We denote by $\hat\tau_{\text{SW}}^\mathrm{fed}=\sum_{k=1}^K \frac{n_k}{n} \hat\tau_k^\mathrm{fed}$:
\begin{align*}
    \textbf{SS Weighted Fed. ATE} & \quad \aVar(\hat\tau_{\text{SW}}^\mathrm{fed}) = \E\left(\aVar\left(\sum_{k=1}^K \frac{n_k}{n} \hat\tau_k^\mathrm{fed}\mid H\right)\right) +  0 \\
    &= \E\left(\aVar\left(\sum_{k=1}^{K} \frac{n_k}{n} \frac{1}{n_k} \sum_{i=1}^{n_k} \left(X'_{k,i} \hat\theta_{\text{fed}}^{(1)} - X'_{k,i} \hat\theta_{\text{fed}}^{(0)}\right)\mid H\right)\right)\\
    &= \E\left(\aVar\left(\frac{1}{n}\sum_{i=1}^{n} \left(X'_i \hat\theta_{\text{fed}}^{(1)} - X'_i \hat\theta_{\text{fed}}^{(0)}\right)\mid H\right)\right)\\
    &= \E\left(\aVar\left(\frac{1}{n}\sum_{i=1}^{n} \left(X'_i \hat\theta_{\text{pool}}^{(1)} - X'_i \hat\theta_{\text{pool}}^{(0)}\right)\mid H\right)\right)\\
    &= \E\left(\aVar\left(\hat\tau_\mathrm{pool}\mid H\right)\right)\\
    &= \aVar\left(\hat\tau_{\text{pool}}\right)\\
    &= \frac{\sigma^2}{n} \frac{1}{p(1-p)} + \frac{1}{n}\Vert \beta^{(1)} - \beta^{(0)} \Vert^2_\Sigma
\end{align*}
We denote by $\hat\tau_{\text{IVW}}^\mathrm{fed}=\frac{\sum_{k=1}^K \left(\aVar(\hat{{\tau}}_k^\mathrm{fed})^{-1} \hat{\tau}_k^\mathrm{fed}\right)}{\sum_{k=1}^K \aVar(\hat{{\tau}}_k^\mathrm{fed})^{-1}}$:
\begin{align*}
    \textbf{IV Weighted Fed. ATE} & \quad \aVar(\hat\tau_{\text{IVW-agg}}) = \E\bigl(\aVar\big(\frac{\sum_{k=1}^K \left(\aVar(\hat{{\tau}}_k^\mathrm{fed})^{-1} \hat{\tau}_k^\mathrm{fed}\right)}{\sum_{k=1}^K \aVar(\hat{{\tau}}_k^\mathrm{fed})^{-1}}\big)\bigr)\\
    &= \aVar(\hat\tau_{\text{SW-agg}}) && \text{from \cref{prop:sw_ivw_same_weights}} \\
    &= \frac{\sigma^2}{n} \frac{1}{p(1-p)} + \frac{1}{n}\Vert \beta^{(1)} - \beta^{(0)} \Vert^2_\Sigma
\end{align*}

\subsubsection{Comparison of asymptotic variances - General Case}\label{proof:comparison_variances}
\textbf{Proof of \cref{prop:metas_larger_var_than_pool}:}
Under the graphical model in Figure~\ref{graph:homog}:
\begin{itemize}
    \item $\begin{aligned}[t]
        \aVar(\hat{\tau}_\mathrm{pool}) &= \aVar(\hat\tau_\mathrm{1S\text{-}SW}) = \aVar(\hat\tau_\mathrm{1S\text{-}IVW}) = \aVar(\hat\tau_\mathrm{GD})
        \end{aligned}$ (\cref{tab:taus_var})

    \item From \cref{prop:meta_ivw_min_var} we have: $\aVar(\hat\tau_\mathrm{meta\text{-}IVW}) \leq \aVar(\hat\tau_\mathrm{SW})$.
    \item Proving that $\aVar(\hat\tau_\mathrm{pool}) \leq \aVar(\hat\tau_\mathrm{Meta\text{-}IVW})$ is equivalent to proving the following inequality:
\begin{align*}
    \frac{1}{\sum_{k=1}^{K} \left(\frac{\sigma^2}{n_k p_k(1-p_k)} + \frac{1}{n_k} \Vert \beta^{(1)} - \beta^{(0)} \Vert^2_\Sigma\right)^{-1}} &\geq \frac{\sigma^2}{n p (1-p)} + \frac{1}{n} \Vert \beta^{(1)} - \beta^{(0)} \Vert^2_\Sigma \\
    \iff \frac{1}{D} &\geq \frac{\sigma^2}{p (1-p)} + a 
\end{align*}
with $D = \sum_{k=1}^{K} \frac{n_k}{n} \frac{x_k}{\sigma^2 + x_k a}$, $x_k = p_k(1-p_k)$ and $a = \Vert \beta^{(1)} - \beta^{(0)} \Vert^2_\Sigma$.

$x \mapsto \frac{x}{\sigma^2 + x a}$ is concave. Therefore, by Jensen's inequality $D\leq \frac{\overline{x}}{\sigma^2 + \overline{x} a}$, with $\overline{x} = \sum_{k=1}^{K} \frac{n_k}{n} x_k $. We then have:
\begin{align*}
    \frac{1}{D} \geq \frac{\sigma^2 + \overline{x} a}{\overline{x}} = \frac{\sigma^2}{\overline{x}} + a = \frac{\sigma^2}{\sum_{k=1}^{K} \frac{n_k}{n}p_k (1-p_k)} + a
\end{align*}

Since $p \mapsto p(1-p)$ is also concave, by Jensen's inequality again we have:
\begin{align*}
    \sum_{k=1}^{K} \frac{n_k}{n} p_k (1-p_k) \leq \left(\sum_{k=1}^{K} \frac{n_k}{n} p_k\right) \left(\sum_{k=1}^{K} \frac{n_k}{n} (1-p_k)\right) = p(1-p)
\end{align*}

So that $\frac{1}{D} \geq \frac{\sigma^2}{p(1-p)} + a$, which ends the proof.
    In conclusion, $$\aVar(\hat\tau_\mathrm{pool}) \leq \aVar(\hat\tau_\mathrm{Meta\text{-}IVW})$$
\end{itemize}
which concludes into $\aVar(\hat{\tau}_\mathrm{pool}) = \aVar(\hat{\tau}_\mathrm{GD}) = \aVar(\hat{\tau}_\mathrm{1S\text{-}SW}) = \aVar(\hat{\tau}_\mathrm{1S\text{-}IVW}) \leq \aVar(\hat{\tau}_\mathrm{Meta\text{-}IVW}) \leq \aVar(\hat{\tau}_{\mathrm{Meta\text{-}SW}})$.

Illustrating examples of this property:
\begin{enumerate}
    \item $\aVar(\hat{\tau}_\mathrm{Meta\text{-}IVW}) - \aVar(\hat{\tau}_\mathrm{pool}) \geq 0 $ increases as the treatment probabilities $\{p_k\}_k$ become more distinct. For example, with $K=2$ studies with balanced datasets ($n_1 = n_2 = n/2$), having $p_1 = 0.99$ and $p_2 = 0.01$ yields $\aVar(\hat{\tau}_\mathrm{Meta\text{-}IVW})$ to be 8 times larger than $\aVar(\hat{\tau}_\mathrm{pool})$ (where we chose $\sigma^2 = 1$ and $\Vert \beta^{(1)} - \beta^{(0)} \Vert^2_\Sigma = 10$).
    \item Similarly, $\aVar(\hat{\tau}_{\mathrm{Meta\text{-}SW}}) - \aVar(\hat{\tau}_\mathrm{Meta\text{-}IVW}) \geq 0 $ grows with the $\{p_k(1-p_k)\}_k$ terms being more and more different from one another. For $K=2$ and $n_1 = n_2 = n/2$, having $p_1 = 0.99$ and $p_2 = 0.5$ gives a $2.5$ larger asymptotic variance of the Meta-SW than that of the Meta-IVW estimator.
\end{enumerate}

\subsubsection{Comparison of asymptotic variances - Special Case}
\textbf{Proof that the estimators of the ATE have equal asymptotic variances when one RCT is conducted over $K$ studies:}
Under this setting, we modify the graphical model in Figure~\ref{graph:homog} and remove the edge between $H$ and $W$:
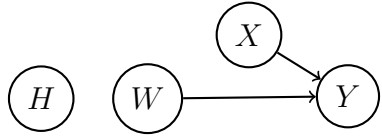
\begin{figure}[!h]
    \centering
    \begin{tikzpicture}[node distance={15mm}, thick, main/.style = {draw, circle}]
    \node[main] (1) {$W$};
    \node[main] (2) [above right=2mm and 7 mm of 1] {$X$};
    \node[main] (3) [below right=2mm and 7 mm of 2] {$Y$};
    \node[main] (4) [left =5mm of 1] {$H$};
    \draw[->] (2) -- (3);
    \draw[->] (1) -- (3);
    \coordinate (midpoint) at ($(1)!0.5!(3)$);
    \end{tikzpicture}
    \caption{Graphical Model of One RCT Conducted Over $K$ Studies}\label{graph:homog_one_rct}
\end{figure}

Under this graphical model (Figure~\ref{graph:homog_one_rct}), using the variances in \cref{tab:taus_var} and as $\forall k, p_k=p$, we have:\label{proof:metas_equal_var_to_pool}

First, $
    \aVar(\hat{\tau}_\mathrm{pool}) = \aVar(\hat{\tau}_\mathrm{1S\text{-}SW}) = \aVar(\hat{\tau}_\mathrm{1S\text{-}IVW}) = \aVar(\hat{\tau}_\mathrm{GD})
    $.

Then, \begin{align*}
        \aVar(\hat{\tau}_\mathrm{Meta\text{-}SW}) 
        &= \frac{\sigma^2}{n} \sum_{k=1}^{K} \frac{\E(n_k)}{n} \frac{1}{p_k(1-p_k)} + \frac{1}{n} \left\Vert \beta^{(1)} - \beta^{(0)} \right\Vert^2_\Sigma \\
        &= \frac{\sigma^2}{n} \frac{1}{p(1-p)} \sum_{k=1}^{K} \frac{\E(n_k)}{n} + \frac{1}{n} \left\Vert \beta^{(1)} - \beta^{(0)} \right\Vert^2_\Sigma \\
        &= \frac{\sigma^2}{n} \frac{1}{p(1-p)} + \frac{1}{n} \left\Vert \beta^{(1)} - \beta^{(0)} \right\Vert^2_\Sigma \\
        &= \aVar(\hat{\tau}_\mathrm{pool})
    \end{align*}

And \begin{align*}
        \aVar(\hat{\tau}_\mathrm{Meta\text{-}IVW}) 
        &= \frac{1}{n} \left(\sum_{k=1}^{K} \frac{\rho_k}{\frac{\sigma^2}{p_k(1-p_k)} + \Vert \beta^{(1)} - \beta^{(0)} \Vert^2_\Sigma}\right)^{-1} \\
        &= \left(\sum_{k=1}^{K} \left(\frac{\sigma^2}{\E(n_k)} \frac{1}{p_k(1-p_k)} + \frac{1}{\E(n_k)} \left\Vert \beta^{(1)} - \beta^{(0)} \right\Vert^2_\Sigma \right)^{-1} \right)^{-1} \\
        &= \left(\sum_{k=1}^{K} \left(\frac{\sigma^2}{\E(n_k)} \frac{1}{p(1-p)} + \frac{1}{\E(n_k)} \left\Vert \beta^{(1)} - \beta^{(0)} \right\Vert^2_\Sigma \right)^{-1} \right)^{-1} \\
        &= \left(\sum_{k=1}^{K} \left(\frac{1}{\E(n_k)} \left(\frac{\sigma^2}{p(1-p)} + \left\Vert \beta^{(1)} - \beta^{(0)} \right\Vert^2_\Sigma \right)\right)^{-1} \right)^{-1} \\
        &= \frac{\sigma^2}{n} \frac{1}{p(1-p)} + \frac{1}{n} \left\Vert \beta^{(1)} - \beta^{(0)} \right\Vert^2_\Sigma \\
        &= \aVar(\hat{\tau}_\mathrm{pool})
    \end{align*}

which gives $\aVar(\hat{\tau}_\mathrm{pool}) = \aVar(\hat{\tau}_\mathrm{GD}) = \aVar(\hat{\tau}_\mathrm{1S-IVW}) =
    \aVar(\hat{\tau}_\mathrm{1S-SW}) = \aVar(\hat{\tau}_\mathrm{Meta-IVW}) = \aVar(\hat{\tau}_{\mathrm{Meta-SW}})$.

\subsection{Heterogeneous Settings}
\subsubsection{Distributional Shift in Covariates}
In this part, we consider the graphical model in Figure~\ref{graph:diff_distribs}, and model~\ref{model:model1}.

\textbf{Unbiasedness and asymptotic variance of Meta-SW under covariate heterogeneity:}\label{proof:meta_sw_diff_means}
\begin{align*}
    \E(\hat \tau_\mathrm{Meta\text{-}SW}) &= \E_H\left(\E\left(\sum_{k=1}^K \frac{n_k}{n} \hat \tau_k |H\right)\right)\\
    &= \sum_k \E_H\left(\frac{n_k}{n}\right) \E(\hat \tau_k) \\
    &= \sum_k \rho_k \tau_k && \text{under \cref{cond:local_large_sample_size} and model \eqref{model:model1}}\\
    &= \tau
\end{align*}
\begin{align*}
    \aVar(\hat \tau_\mathrm{Meta\text{-}SW}) &= \E\left(\aVar\left(\hat\tau_\mathrm{Meta\text{-}SW}\right)\right) + \aVar\left(\E\left(\hat\tau_\mathrm{Meta\text{-}SW}\right)\right)\\
    &= \sum_k \E\left(\left(\frac{n_k}{n}\right)^2 \aVar(\hat \tau_k)\right) \\
    &= \sum_{k=1}^K \frac{\E(n_k)}{n^2}\left( \frac{\sigma^2}{p_k(1-p_k)} + \Vert \beta^{(1)} - \beta^{(0)} \Vert^2_{\Sigma_k}\right)  \\
    &= \frac{1}{n} \sum_{k=1}^K \rho_k \frac{\sigma^2}{p_k(1-p_k)} + \frac{1}{n} \sum_{k=1}^K \rho_k \left(\beta^{(1)} - \beta^{(0)}\right)^\top \Sigma_k \left(\beta^{(1)} - \beta^{(0)}\right) \\
    &=\frac{1}{n} \sum_{k=1}^K \rho_k  \frac{\sigma^2}{p_k(1-p_k)} + \frac{1}{n} \left(\beta^{(1)} - \beta^{(0)}\right)^\top \sum_{k=1}^K \rho_k \Sigma_k \left(\beta^{(1)} - \beta^{(0)}\right) \\
    &= \frac{\sigma^2}{n} \sum_{k=1}^K  \frac{\rho_k}{p_k(1-p_k)} + \frac{1}{n} \Vert \beta^{(1)} - \beta^{(0)} \Vert^2_{\Sigma} \\
\end{align*}

\textbf{Unbiasedness and asymptotic variance of GD under covariate heterogeneity:}\label{proof:gd_pool_diff_means_var}
The computation of the GD estimator's bias and asymptotic variance are direct using exactly the same proof as in \cref{proof:asymp_ols} since the assumptions over $X_\mathrm{pool}$ are still met in the distributional shift setting.

\textbf{Proof of \cref{prop:1s_diff_means_var}:}\label{proof:1s_diff_means_var}\\
We have $\hat c_k^{(w)} = \bar Y_k^{(w)} - \bar X_k^{(w)} \hat\beta^{(w)}$ 
so that:
\begin{align*}
    \V(\hat c_k^{(w)}\mid X_k^{(w)}) &= \V(\bar Y_k^{(w)} - \bar X_k^{(w)} \hat\beta^{(w)}\mid X_k^{(w)})\\
    &= \V(\bar Y_k^{(w)}) + \V(\bar X_k^{(w)} \hat\beta^{(w)}\mid X_k^{(w)}) - 2\cov{\bar Y_k^{(w)}\mid X_k^{(w)}}{\bar X_k^{(w)}\hat\beta^{(w)}\mid X_k^{(w)}}\\
    &= \frac{\sigma^2}{n_k^{(w)}} + \bar {X_k^{(w)}}^\top \V(\hat\beta^{(w)}\mid X_k^{(w)})\bar X_k^{(w)} - 2\cov{\bar Y_k^{(w)}\mid X_k^{(w)}}{\bar X_k^{(w)}\hat\beta^{(w)}\mid X_k^{(w)}}
\end{align*}

Given $X_k^{(w)}$ we have:
\begin{align*}
    \cov{\bar{Y}_k^{(w)}}{\hat{\beta_k}} &= \cov{\frac{1}{n_k^{(w)}}\sum_{i=1}^{n_k^{(w)}} Y_{k,i}}{ \frac{\sum_{j=1}^{n_k^{(w)}}(X_{k,j}^{(w)} - \bar{X}_k^{(w)})Y_{k,j}}{\sum_{i=1}^{n_k^{(w)}}(X_{k,i} - \bar{X}_k^{(w)})^2}} \\
    &= \frac{1}{n_k^{(w)} \sum_{i=1}^{n_k^{(w)}}(X_{k,i}^{(w)} - \bar{X}_k^{(w)})^2} \cov{\sum_{i=1}^{n_k^{(w)}} Y_{k,i}}{\sum_{j=1}^{n_k^{(w)}}(X_{k,j} - \bar{X}_k^{(w)})Y_j} \\
    &= \frac{1}{n_k^{(w)} \sum_{i=1}^{n_k^{(w)}}(X_{k,i} - \bar{X}_k^{(w)})^2} \sum_{i=1}^{n_k^{(w)}} \sum_{j=1}^{n_k^{(w)}}(X_{k,j}^{(w)} - \bar{X}_k^{(w)}) \cov{Y_{k,i}}{Y_{k,j}} \\
    &= \frac{1}{n_k^{(w)} \sum_{i=1}^{n_k^{(w)}}(X_{k,i}^{(w)} - \bar{X}_k^{(w)})^2} \sum_{j=1}^{n_k^{(w)}} (X_{k,
    j}^{(w)} - \bar{X}_k^{(w)}) \sigma^2 \\
    &= 0
\end{align*}

Therefore, $\V(\hat c_k^{(w)}\mid X_k^{(w)}) = \sigma^2\left(\frac{1}{n_k^{(w)}} +(\bar{X}_k^{(w)})^\top
({X_k^{(w)}}^\top X_k^{(w)})^{-1}\bar{X}_k^{(w)}\right)$, which yields in a random design over the $X_k^{(w)}$: $$\aVar(\hat c_k^{(w)}) = \sigma^2\left(\frac{1}{n_k^{(w)}} + \E\left((\bar{X}_k^{(w)})^\top ({X_k^{(w)}}^\top X_k^{(w)})^{-1} \bar{X}_k^{(w)}\right)\right)$$
where 
\begin{align*}
    \E\left((\bar{X}_k^{(w)})^\top ({X_k^{(w)}}^\top X_k^{(w)})^{-1} \bar{X}_k^{(w)}\right) &= \mu_k^\top \Sigma_k^{-1} \mu_k + \E\left(\frac{1}{n_k^{(w)}} (\bar{X}_k^{(w)} - \mu_k)^\top \Sigma_k^{-1} (\bar{X}_k^{(w)} - \mu_k)\right) \\
    &= \mu_k^\top \Sigma_k^{-1} \mu_k + \frac{1}{n_k^{(w)}} \mathrm{Tr}\left(\Sigma_k^{-1} \sum_{i=1}^{n_k^{(w)}} \E\left((X_{ki}^{(w)} - \mu_k) (X_{ki}^{(w)} - \mu_k)^\top\right)\right) \\
    &= \mu_k^\top \Sigma_k^{-1} \mu_k + \frac{1}{n_k^{(w)}} \mathrm{Tr}\left(\Sigma_k^{-1} \sum_{i=1}^{n_k^{(w)}} \Sigma_k\right) \\
    &= \mu_k^\top \Sigma_k^{-1} \mu_k + \frac{1}{n_k^{(w)}} \mathrm{Tr}\left(\Sigma_k^{-1}\right)
\end{align*}
so that $\aVar(\hat c_k^{(w)}) = \sigma^2\left(\frac{1}{n_k^{(w)}} + \mu_k^\top \Sigma_k^{-1} \mu_k + \frac{1}{n_k^{(w)}} \mathrm{Tr}\left(\Sigma_k^{-1}\right)\right)$.

Then, we get $\aVar(\hat c_\mathrm{pool}^{(w)}) = \sigma^2\left(\frac{1}{n^{(w)}} + \mu^\top \Sigma^{-1} \mu + \frac{1}{n^{(w)}} \mathrm{Tr}\left(\Sigma^{-1}\right)\right)$ and
\begin{align*}
    \aVar(\hat c_\mathrm{1S\text{-}SW}^{(w)}) &= \sigma^2\sum_{k=1}^{K}\left(\frac{n_k^{(w)}}{n^{(w)}}\right)^2 \left(\frac{1}{n_k^{(w)}} + \mu_k^\top \Sigma_k^{-1} \mu_k + \frac{1}{n_k^{(w)}} \mathrm{Tr}\left(\Sigma_k^{-1}\right)\right)\\
    &= \sigma^2\left(\frac{1}{n^{(w)}} + \sum_{k=1}^{K} \left(\frac{n_k^{(w)}}{n^{(w)}}\right)^2 \left(\mu_k^\top \Sigma_k^{-1} \mu_k + \frac{1}{n_k^{(w)}} \mathrm{Tr}\left(\Sigma_k^{-1}\right)\right)\right)
\end{align*}

Finally,
\begin{align*}
    \frac{1}{\sigma^2}\left(\aVar(\hat c_\mathrm{pool}^{(w)}) - \aVar(\hat c_\mathrm{1S\text{-}SW}^{(w)})\right) &= \mu^\top \Sigma^{-1} \mu - \sum_{k=1}^{K} \left(\frac{n_k^{(w)}}{n^{(w)}}\right)^2 \mu_k^\top \Sigma_k^{-1} \mu_k\\
    &= \left(\sum_{k=1}^{K} \frac{n_k^{(w)}}{n^{(w)}} \mu_k^\top\right) \left(\sum_{k=1}^{K}\frac{n_k^{(w)}}{n^{(w)}}\left(\Sigma_k +(\mu_k- \mu)(\mu_k - \mu)^\top\right)\right)^{-1}\\
    &\quad\quad \times \left(\sum_{k=1}^{K} \frac{n_k^{(w)}}{n^{(w)}} \mu_k\right) - \sum_{k=1}^{K} \left(\frac{n_k^{(w)}}{n^{(w)}}\right)^2 \mu_k^\top \Sigma_k^{-1} \mu_k
\end{align*}
With Jensen's inequality applied to the convex function $f(x) = x^\top \Sigma_k^{-1} x$ for any positive definite matrix $\Sigma_k$ with positive and summing to 1 weights, we have:

$$\left(\sum_{k=1}^{K} \frac{n_k^{(w)}}{n^{(w)}} \mu_k\right)^\top \left(\sum_{k=1}^{K} \frac{n_k^{(w)}}{n^{(w)}} \Sigma_k\right)^{-1} \left(\sum_{k=1}^{K} \frac{n_k^{(w)}}{n^{(w)}} \mu_k\right) \le \sum_{k=1}^{K} \frac{n_k^{(w)}}{n^{(w)}} \mu_k^\top \Sigma_k^{-1} \mu_k.$$

Therefore,
$$\mu^\top \Sigma^{-1} \mu - \sum_{k=1}^{K} \left(\frac{n_k^{(w)}}{n^{(w)}}\right)^2 \mu_k^\top \Sigma_k^{-1} \mu_k \le \sum_{k=1}^{K} \frac{n_k^{(w)}}{n^{(w)}} \left(1 - \frac{n_k^{(w)}}{n^{(w)}}\right) \mu_k^\top \Sigma_k^{-1} \mu_k \le 0$$
After applying the law of total variance over $H$, we get $\aVar(\hat c_\mathrm{pool}^{(w)}) \le \aVar(\hat c_\mathrm{1S\text{-}SW}^{(w)})$, implying $\aVar(\hat \tau_\mathrm{pool}) \le \aVar(\hat \tau_\mathrm{1S\text{-}SW})$.

In particular, notice that:
\vspace*{-\baselineskip}
\begin{itemize}
    \item If the studies have equal means ($\forall k, \mu_k = \mu$), then $\aVar(\hat c_\mathrm{pool}^{(w)}) = \aVar(\hat c_\mathrm{1S\text{-}SW}^{(w)})$, even if the $\{\Sigma_k\}_k$ are different.
    \item If a small number of studies have very distinct means from the rest of the studies, the difference in variances between the Pool and 1S-SW estimators will be large.
\end{itemize}

\subsection{Study-Effects}
In this part, we consider the graphical model in Figure~\ref{graph:diff_intercepts} and model~(\ref{model:model2}).

\textbf{Proof of meta estimators' unbiasedness and asymptotic variance (\cref{prop:meta_study_effects_intercepts})}\label{proof:meta_study_effects_intercepts}:\\
First, let us compute the intercepts for the treated and control groups in study $k$ under model \eqref{model:model2}. 

First, $a_k^{(w)}=c^{(w)} + h_k$ is the intercept in model \eqref{model:model2}, so that $a_k^{(w)} = \E(Y_k^{(w)} - X_k \beta^{(w)} - \varepsilon_k^{(w)}) = \E(Y_k^{(w)}) - \E(X_k)\beta^{(w)}$, which yields a \textbf{locally estimated} intercept $\hat a_k^{(w)} = \overline{{Y_k}^{(w)}} - \overline{X_k}^{(w)} \hat{\beta}_k^{(w)}$ in study $k$ for treatment group $w$.

Then, we can compute the Meta estimator with weights $\omega_k$:
\begin{align*}
    \hat{\tau}_{\mathrm{Meta}} &= \sum_{k=1}^K \omega_k \hat{\tau}_k\\
    &= \sum_{k=1}^K \omega_k \left(\hat a_k^{(1)} - \hat a_k^{(0)} + \overline{X_k}(\hat\beta^{(1)} - \hat\beta^{(0)})\right) \\
    &= \sum_{k=1}^K \omega_k \left((a_k^{(1)}+\hat \epsilon^{(1)}) -  (a_k^{(0)}+\hat \epsilon^{(0)}) + \overline{X_k}(\hat\beta^{(1)} - \hat\beta^{(0)})\right) \\
    &= \sum_{k=1}^K \omega_k \left((c^{(1)} + h_k +\hat \epsilon^{(1)}) - (c^{(0)} + h_k +\hat \epsilon^{(0)})+ \overline{X_k}(\hat\beta^{(1)} - \hat\beta^{(0)})\right) \\
    &= \sum_{k=1}^K \omega_k \left(\hat c^{(1)} - \hat c^{(0)} + \overline{X_k}(\hat\beta^{(1)} - \hat\beta^{(0)})\right) \\
\end{align*} where $\hat c^{(w)} = c^{(w)} + \hat \epsilon^{(w)}$ and $\hat \epsilon^{(w)}$ is the estimation bias of the OLS, with expectancy 0.

Therefore,
\begin{align*}
    \E(\hat{\tau}_{\mathrm{Meta}}) &= \sum_{k=1}^K \omega_k \E\left(\hat a_k^{(1)} - \hat a_k^{(0)} + \overline{X_k}(\hat\beta^{(1)} - \hat\beta^{(0)})\right) \\
    &= \sum_{k=1}^K \omega_k \E\left(\hat c^{(1)} - \hat c^{(0)} + \overline{X_k}(\hat\beta^{(1)} - \hat\beta^{(0)})\right)\\
    &= \sum_{k=1}^K \omega_k \E(\hat \tau_k) \\
    &= \sum_{k=1}^K \omega_k \tau_k 
\end{align*}



which leads the Meta estimators to be unbiased and have the same asymptotic variance as in \cref{tab:taus_var}.
\subsubsection{Unadjusted Federated Estimators}
We define the pooled outcome model parameters estimator as $\hat{\theta}_\mathrm{pool}^{(w)} = \argmin_{\theta} \sum_{i=1}^{n} \left(Y_i^{(w)} - X'_i \theta\right)^2$, and define $\hat a_\mathrm{pool}^{(w)}$ the estimated intercept in the pooled data without the membership variable for group $w$, \textit{i.e.} we have $\hat\theta^{(w)} = \{\hat a_\mathrm{pool}^{(w)}, \hat\beta_\mathrm{pool}^{(w)}\}$ and the estimated intercept in the pooled data is:
\begin{align*}
    \hat{a}_\mathrm{pool}^{(w)} &= \overline{Y^{(w)}} - \overline{X'^{(w)}} \hat{\theta}_\mathrm{pool}^{(w)} \\
    &= \frac{1}{n^{(w)}}\sum_{i=1}^{n^{(w)}} Y_i(w) - \frac{1}{n^{(w)}}\sum_{i=1}^{n^{(w)}} {X'_i}^{(w)} \hat{\theta}_\mathrm{pool}^{(w)} \\
    &= \frac{1}{n^{(w)}}\sum_{i=1}^{n^{(w)}} \left(c^{(w)} + h_{k,i} + X'_{k,i} \theta^{(w)} + \varepsilon_i(w)\right) - \frac{1}{n^{(w)}}\sum_{i=1}^{n^{(w)}} {X'_i}^{(w)} \hat{\theta}_\mathrm{pool}^{(w)} && \text{with } h_{k,i} = h_k \mathds{1}_{\{H_i = k\}} \\
    &= c^{(w)} + \overline{h_k^{(w)}} + \overline{\varepsilon_i(w)} + \overline{X^{(w)}} \left(\beta^{(w)} - \hat{\beta}_\mathrm{pool}^{(w)}\right) \\
\end{align*}
with:
\begin{itemize}
    \item $\overline{h_k^{(w)}} = \frac{1}{n^{(w)}}\sum_{i=1}^{n^{(w)}} h_{k,i} = \frac{1}{n^{(w)}} \sum_{k=1}^{n^{(w)}} n_k^{(w)} h_k$ is the average effect of study $k$ in group $w$. 
    \item $\overline{\varepsilon_i(w)} = \frac{1}{n^{(w)}}\sum_{i=1}^{n^{(w)}} \varepsilon_i(w)$ is the average error in group $w$.
    \item $\overline{X^{(w)}} = \frac{1}{n^{(w)}}\sum_{i=1}^{n^{(w)}} X_i^{(w)}$ is the average covariate in group $w$.
\end{itemize} 

Therefore, the estimate of the intercept in the pooled dataset in presence of study-effects has expectancy:
\begin{align*}
    \E(\hat{a}_\mathrm{pool}^{(w)}) &= c^{(w)} + \E\left(\overline{h_k^{(w)}}\right) \\
    &= c^{(w)} + \E\left(\frac{1}{n^{(w)}} \sum_{k=1}^{K} n_k^{(w)} h_k\right) \\
    &= c^{(w)} + \sum_{k=1}^{K} \E\left(\frac{n_k^{(w)}}{n^{(w)}}\right) h_k 
\end{align*}

Then the unadjusted pooled estimator is:
\begin{align*}
    \hat{\tau}_{\mathrm{pool}} &= \frac{1}{n}\sum_{i=1}^{n} \left(\hat{a}_\mathrm{pool}^{(1)} - \hat{a}_\mathrm{pool}^{(0)} + X_i(\hat\beta_\mathrm{pool}^{(1)} - \hat\beta_\mathrm{pool}^{(0)})\right)
\end{align*}

We now prove that this unadjusted estimator is biased when the $p_k$ probabilities are not equal across studies. Then, we have:
\label{proof:bias_ols_study_effect_intercepts} 
\begin{align*}
    \E(\hat{\tau}_{\mathrm{pool}}) &= \E\left(\hat a_\mathrm{pool}^{(1)} - \hat a_\mathrm{pool}^{(0)} + \overline{X}(\hat\beta_\mathrm{pool}^{(1)} - \hat\beta_\mathrm{pool}^{(0)})\right) \\
    &= c^{(1)} - c^{(0)} + \sum_{k=1}^{K} \E\left(\frac{n_k^{(1)}}{n^{(1)}}\right) h_k - \sum_{k=1}^{K} \E\left(\frac{n_k^{(0)}}{n^{(0)}}\right) h_k \\
    &= \tau + \sum_{k=1}^{K} \E\left(\frac{n_k^{(1)}}{n^{(1)}} - \frac{n_k^{(0)}}{n^{(0)}}\right) h_k \\
    &= \tau + \underbrace{\sum_{k=1}^{K} \left(p_k \E\left(\frac{n_k}{\sum_{k=1}^{K} n_k p_k } \right) - (1-p_k) \E\left(\frac{n_k}{\sum_{k=1}^{K} n_k (1-p_k)} \right)\right) h_k}_\mathrm{bias} 
\end{align*}

Then consider the two following cases:
\begin{itemize}
    \item If equal treatment assignment ($p_k = p$ for all $k$): \\
    Then, the bias is equal to $\sum_{k=1}^{K} p \E\left(\frac{n_k}{p n} \right) h_k - (1-p) \E\left(\frac{n_k}{(1-p) n} \right) h_k = 0$ so that $\E(\hat{\tau}_{\mathrm{pool}}) = \tau$.
    \item If unequal treatment assignment ($p_k \neq p$ for all $k$): \\
    Then, the bias is $\neq 0$ so that $\E(\hat{\tau}_{\mathrm{pool}}) \neq \tau$.
\end{itemize}

Therefore, the Pool estimator is biased when the treatment probabilities are not equal among studies. This happens because the variable $H$ acts as a confounder between the treatment and the outcome in cases of study-effects. 
We need to account for it by adding a membership variable $H$ in the dataset, which will allow the model to estimate the study-effects.

\subsubsection{With membership variable in the dataset}
\textbf{Proof of unbiasedness of the adjusted GD estimator:}\label{proof:gd_intercepts}
Denoting $\beta_H = (h_1, \dots, h_K)^\top$ the coefficients of the membership variables, model~(\ref{model:model2}) can be written as:
\begin{align}\label{model:model_w_membership}
    Y_{k,i}(w) &= c^{(w)} + h_k + X_{k,i} \beta^{(w)} + \varepsilon_i (w)\nonumber\\
    &= c^{(w)} + H_{k,i} \beta_H + X_{k,i} \beta^{(w)} + \varepsilon_i (w) 
\end{align}


Under \cref{model:model_w_membership}, the (adjusted) Pooled estimator then estimates the coefficients of the variables $H$ as a substitute for the study-effects. This technique does not allow to estimate distinctly the intercepts $c^{(w)}$ and the effects of the studies $\{h_k\}_k$, as it relies on a relative rescaling of the intercepts of each study with respect to a choosen study of reference. In practice, choosing study 1 as the reference study by not including $H_1$ in the dataset offers the advantages of avoiding the underdeterminancy of the solutions of $\{(c^{(w)}, h_k)\}_k$ in \cref{model:model_w_membership}, and is easy to implement. In any case, this is without loss of generality.

Results from OLS estimation yield that the estimated intercept is the mean of the outcomes in the reference study (we arbitrarily choose study 1 to be the reference study), and the estimated coefficients of the membership variables are the differences between the means of the outcomes in the reference study and the other studies, which writes as:
\begin{align*}
    \hat{c}_\mathrm{pool}^{(w)} &= \frac{1}{n_1^{(w)}} \sum_{i=1}^{n_1^{(w)}} Y_{1,i}(w) \\
    \hat h_k &= \frac{1}{n_k^{(w)}} \sum_{i=1}^{n_k^{(w)}} Y_{k,i}(w) - \hat{c}_\mathrm{pool}^{(w)} \quad \forall k \in \llbracket 2, K \rrbracket
\end{align*}

where $\hat h_k^{(w)}$ is the estimated effect of study $k$ on the outcomes in group $w$. In our model, we have that $\E(\hat{c}_\mathrm{pool}^{(w)}) = Y_1(w)$ so that:
\begin{align*}
    \E\left(\hat h_k^{(w)}\right) &= \E\left(Y_k(w)\right) - \E\left(Y_1(w)\right) \\
    &= \E\left(c^{(w)} + X_{k,i}\beta^{(w)} + h_k + \varepsilon_i(w)\right) - \E\left(c^{(w)} - X_{1,i}\beta^{(w)} - h_1 - \varepsilon_i(w)\right)\\
    & = h_k
\end{align*}

Finally, the adjusted pool estimator is:
\begin{align*}
    \hat{\tau}_{\mathrm{pool\circ}} &= \frac{1}{n}\sum_{i=1}^{n} \left(\hat{c}_\mathrm{pool}^{(1)} - \hat{c}_\mathrm{pool}^{(0)} + X_i(\hat\beta_\mathrm{pool}^{(1)} - \hat\beta_\mathrm{pool}^{(0)}) + \sum_{k=1}^{K} H_{k,i}(\hat{\beta}_H^{(1)} - \hat{\beta}_H^{(0)})\right)\\
    &= \frac{1}{n}\sum_{k=1}^{K} \sum_{i=1}^{n_k} \big(\tilde X'_{k,i}(\hat \theta_\mathrm{pool\circ}^{(1)} - \hat \theta_\mathrm{pool\circ}^{(0)})\big)
\end{align*}
with the augmented design matrix $\tilde X'_{k,i} = (1, X_{k,i}, H_{2,i}, \ldots, H_{K,i})$, $\hat{\beta}_H^{(w)} = (\hat{h}_1^{(w)}, \dots, \hat{h}_K^{(w)})^\top$ the estimated effects of the studies on the outcomes in group $w$ and $\hat\theta_\mathrm{pool\circ}^{(w)} = (\hat c_\mathrm{pool}^{(w)}, \hat \beta_\mathrm{pool}^{(w)}, \hat \beta_H^{(w)})$. We then have $\E(\hat{\beta}_H^{(1)}) = \E(\hat{\beta}_H^{(0)}) = \beta_H$.

Finally, we get an unbiased estimator:
    \begin{align*}
        \E(\hat{\tau}_{\mathrm{pool\circ}}) &= \E\left(\hat{c}_\mathrm{pool}^{(1)} - \hat{c}_\mathrm{pool}^{(0)} + \overline{X}(\hat\beta_\mathrm{pool}^{(1)} - \hat\beta_\mathrm{pool}^{(0)}) + \sum_{k=1}^{K} H_k(\hat{\beta}_H^{(1)} - \hat{\beta}_H^{(0)})\right) \\
        &= \tau + \sum_{k=1}^K \E\left(H_k(\hat{\beta}_H^{(1)} - \hat{\beta}_H^{(0)})\right)\\
        &= \tau + \sum_{k=1}^K \E\left(H_k \E(\hat{\beta}_H^{(1)} - \hat{\beta}_H^{(0)} \mid H_k)\right)\\
        &= \tau + \sum_{k=1}^K \E\left(H_k (\beta_H - \beta_H)\right)\\
        &= \tau
    \end{align*}
    
The asymptotic variance of the adjusted Pool estimator is given by (using \cref{proof:asymp_ols}): $\aVar(\hat{\tau}_{\mathrm{pool\circ}}) = \frac{\sigma^2}{p(1-p)} + \frac{1}{n}\Vert \tilde\beta^{(1)} - \tilde\beta^{(0)} \Vert^2_{\tilde\Sigma}$ with $\tilde\beta^{(w)} = (\beta^{(w)}, \beta_H)$ and $\tilde\Sigma = \V(\tilde X)$. Furthermore, remark that the block covariance-variance matrix is of the form $\tilde\Sigma=\begin{pmatrix}
    \Sigma & 0 \\
    0 & \mathrm{diag}(p_1, \dots, p_K)
    \end{pmatrix}$ because the study-effects are independent of the covariates $X$. Therefore, the asymptotic variance of the adjusted Pooled estimator is:
\begin{align*}
    \aVar(\hat{\tau}_{\mathrm{pool\circ}}) = \frac{\sigma^2}{p(1-p)} + \frac{1}{n}\Vert \beta^{(1)} - \beta^{(0)} \Vert^2_{\Sigma}
\end{align*}
Note that the asymptotic variance of the adjusted Pool estimator in this study-effect model is insensitive to these effects, which means that the heterogeneity among the studies does not affect this estimator. 

Therefore, we can build the federated Gradient Descent with $H$ variables (‘‘$\hat\tau_\mathrm{GD\circ}$'') by allowing the studies to add $K-1$ columns to their local datasets, with one of them containing only ones, unique to their dataset, and under convergence of Alg.~\ref{alg:FedAvg}, we get $\hat\tau_\mathrm{GD\circ} = \hat\tau_\mathrm{pool\circ}$.

\section{SIMULATIONS}\label{sec:app_simulations}
\subsection{Simulation parameters}\label{subsec:simulation_parameters}
We generate data as follows:
\begin{itemize}
    \item[1.] Generate $X_{k,i} \sim \mathcal{N}(\mu_k, \Sigma_k)$ for individual $i\in \llbracket 1, n_k \rrbracket$, study $k\in \llbracket 1, K \rrbracket$ and $d$ covariates. Add a constant covariate $1$ to each $X_{k,i}$ to account for the intercept.
    \item[2.] Generate $W_{k,i} \sim \mathcal{B}(p_k)$ .
    \item[3.] Generate $\varepsilon_{k,i} \sim \mathcal{N}(0, \sigma^2)$ (homoscedasticity).
    \item[4.] Build $Y_{k,i}(w) =  c^{(w)} + h_k + X_{k,i} \beta_k^{(w)} + \varepsilon_{k,i}$, with $W_{k,i} = w$.
\end{itemize}
\begin{itemize}
    \item Studies:
    \begin{itemize}
        \item $K=5$ studies 
        \item Sample sizes: \textit{Large} settings: $\{n_k = 20d\}_k$ so $n=20Kd$; \textit{Small} settings: $\{n_k = 5d\}_k$ so that $n=4Kd$
    \end{itemize}
    \item Model:
    \begin{itemize}
        \item $\sigma^2 = 1$ 
        \item $\{\theta_k^{(w)}\}_k = \{(c^{(w)} + h_k, \beta_1^{(w)}, \dots, \beta_d^{(w)})\}$ with $c^{(1)}=-1.85, c^{(0)}= -2$, \\$\beta^{(1)} = (-1.75, -1.5, -1.25, -1.0, -0.75, -0.5, -0.25, 0.0, 0.25, 0.5)$, \\$\beta^{(0)} = (-1.8, -1.6, -1.4, -1.2, -1, -0.8, -0.6, -0.4, -0.2, 0)$,\\
        $h_k = 0$ (no study-effect by default for model \eqref{model:model1})
    \end{itemize}
    \item Covariates:
    \begin{itemize}
        \item Dimension $d = 10$
        \item $\bigg\{\bigl(\mu_k = (\underbrace{1,\dots,1}_{\lfloor d/2\rfloor},\underbrace{-1,\dots,-1}_{\lfloor d/2\rfloor}), \Sigma_k = 0.5 I_{d} + 0.5 J_{d}\bigr)\bigg\}_k$, $J\in \mathds{1}^{d,d}$ the matrix of ones
    \end{itemize}
    which yields $\tau = c^{(1)} - c^{(0)} + \mu^\top \left(\beta^{(1)} - \beta^{(0)}\right) = -1.1$
    \item Treatment assignment:
    \begin{itemize}
        \item RCT: $W_{k,i} \sim \mathcal{B}(p_k)$ with $\{p_k = 0.5\}_k$; 
    \end{itemize}
\end{itemize} 

For the Meta-IVW estimator, we estimate the asymptotic variance of the local ATE estimates, \textit{i.e.} $\hat{\tau}_{\mathrm{Meta\text{-}IVW}} = \frac{\sum_{k=1}^{K}\widehat{\aVar(\hat{\tau}_k)}^{-1} \hat{\tau}_k}{\sum_{k=1}^{K}\widehat{\aVar(\hat{\tau}_k)}^{-1}}$ with:
\begin{itemize}\label{params_simus:meta_ivw}
    \item $\widehat{\aVar(\hat{\tau}_k)} = \frac{\hat\sigma_k^2}{n_k} \Big(\frac{1}{\widehat{p_k}(1-\widehat{p_k})}\Big) + \frac{1}{n_k}\Vert \hat\beta_k^{(1)} - \hat\beta_k^{(0)} \Vert^2_{\hat\Sigma_k}$
    \item Sample variance of the residuals\\ $\hat\sigma_k^2=\frac{1}{n_k - d - 1} \sum_{i=1}^{n_k}\bigl(Y_{k,i} - X_{k,i} (\hat\beta_k^{(1)} \mathds{1}_{[W_i=1]} + \hat\beta_k^{(0)} \mathds{1}_{[W_i=0]})\bigr)^2$
    \item $\hat p_k=\frac{n_k^{(1)}}{n_k}$ and the sample covariance matrix $\hat\Sigma_k = \frac{1}{n_k-1} \sum_{i=1}^{n_k} (X_{k,i} - \overline{X}_k)(X_{k,i} - \overline{X}_k)^\top$
\end{itemize}

\subsection{Additional scenarios}\label{sec:simulation_figures_appendix}
\subsubsection{Balanced datasets}\label{par:homog}
In this scenario, we generate data in the \textit{homogeneous} setting according to the graphical model in Figure~\ref{graph:homog} and outcome model~\ref{model:model1} with $K=5$ studies and $d=10$ covariates dimension. We consider a balanced setting where all studies have equal sample size $n_k = 100d$ in the \textit{Large} regime and  $n_k = 5d$ in the \textit{Small} sample size regime. The treatment assignment is the same for all studies in this first scenario, with $\forall k, W_{k,i} \sim \mathcal{B}(p)$ and $p=0.5$ for all $k$. With the theoretical considerations on the hyperparameters discussed in \cref{app:choice_of_learning_rate}, we set them to the following values for the full batch Gradient Descent algorithm: $T=1000, E=1, B=n, \eta=0.001$ for the \textit{Small} setting.

\begin{figure}[H]
    \centering
    \begin{minipage}[t]{.9\textwidth}
        \begin{tabular}{>{\centering\arraybackslash}m{6cm}>{\centering\arraybackslash}m{6cm}>{\centering\arraybackslash}m{1cm}}
            \textbf{Large} & \textbf{Small} \\
            \begin{minipage}[t]{.45\textwidth}
                \centering
                \includegraphics[width=\linewidth, trim={0.2cm 1cm 0 1.1cm}, clip]{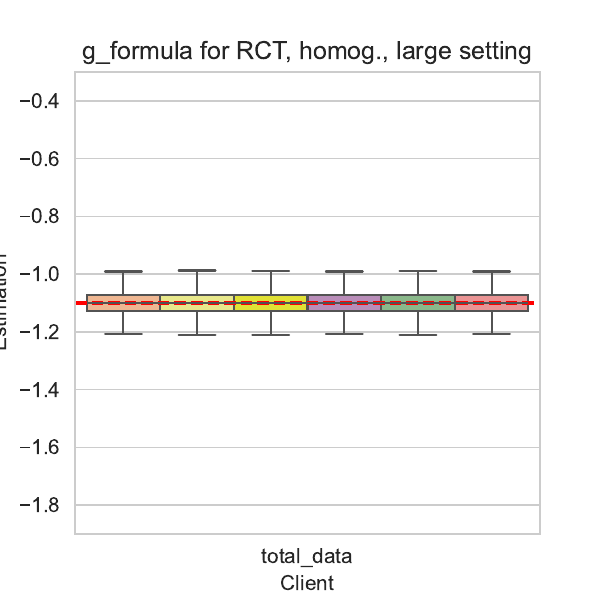}
            \end{minipage} & 
            \begin{minipage}[t]{.45\textwidth}
                \centering
                \includegraphics[width=\linewidth, trim={0.2cm 1cm 0 1.1cm}, clip]{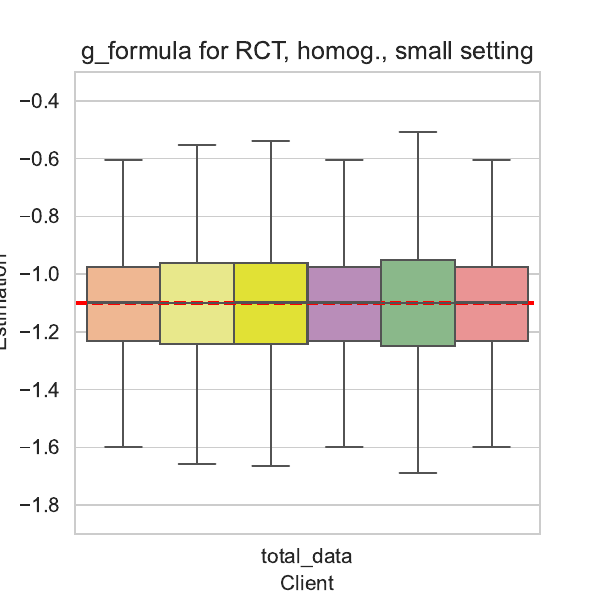}
            \end{minipage} &
            \begin{minipage}[t]{.13\textwidth}
                \centering
              \includegraphics[width=\linewidth]{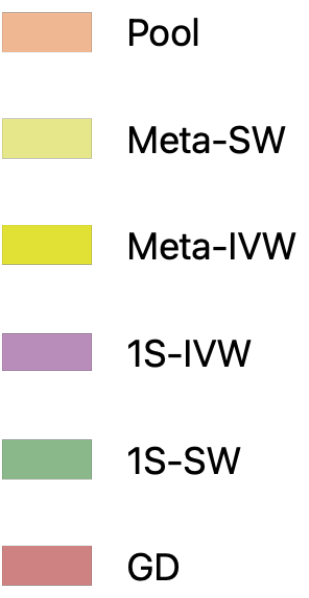}
    \end{minipage}
        \end{tabular}
    \end{minipage}
    \caption{Estimation of the ATE in the balanced and homogeneous setting for RCT studies.}\label{fig:homog}
\end{figure}

Figure \ref{fig:homog} represents the distribution over 2000 simulations of the estimated ATEs using the Pool estimator on the concatenated data as well as the Meta estimators with both weighting strategies (sample weighting and inverse variance weighting), the One-Shot estimators (IVW and SW) and Gradient Descent estimator (GD). 

In large sample size regime, it highlights that in the balanced and homogeneous setting, (left panel \textit{Large}), all estimators are unbiased and have the same variance, as expected in the special case of one RCT conducted over $K$ studies (\cref{subsec:one_rct_special_case}). 

In the small sample size regime (right panel \textit{Small}), the Metas and One-Shot SW have larger variances than the One-Shot IVW and GD estimators which are both equal to the pooled data one as expected given \cref{prop:1S_ivw_thetas_equal_pool}. 

\subsubsection{Imbalanced datasets}\label{par:imbal}
We consider a case of imbalance in the sample sizes of the studies, where one study has more observations than the others. For the \textit{Large} setting,  $n_1=400d$ and $n_2=...=n_5=25d $, leading to the same total number of observations of  $n=5000$ as in \cref{fig:homog}, whereas in the \textit{Small} case $n_1=13d$ and  $n_2=...=n_5=3d$  resulting in  $n=250$, similarly to the balanced scenario.

\begin{figure}[h!]
    \centering
    \begin{minipage}[t]{.9\textwidth}
        \begin{tabular}{>{\centering\arraybackslash}m{6cm}>{\centering\arraybackslash}m{6cm}>{\centering\arraybackslash}m{1cm}}
            \textbf{Large} & \textbf{Small} \\
            \begin{minipage}[t]{.45\textwidth}
                \centering
                \includegraphics[width=\linewidth, trim={0.2cm 1cm 0 1.1cm}, clip]{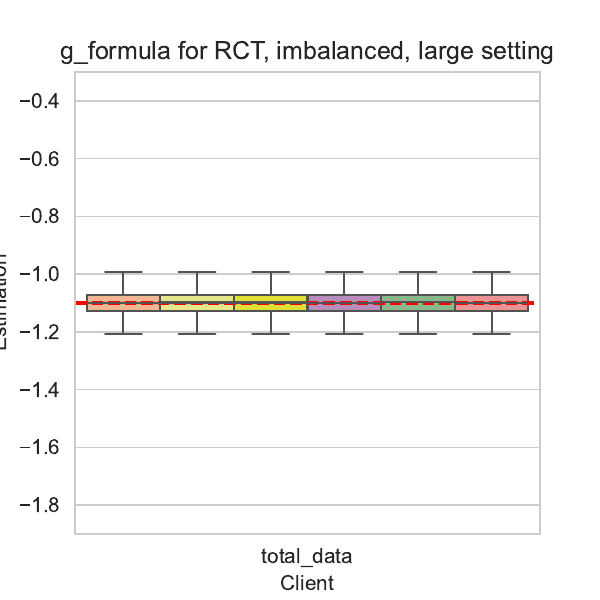}
            \end{minipage} & 
            \begin{minipage}[t]{.45\textwidth}
                \centering
                \includegraphics[width=\linewidth, trim={0.2cm 1cm 0 1.1cm}, clip]{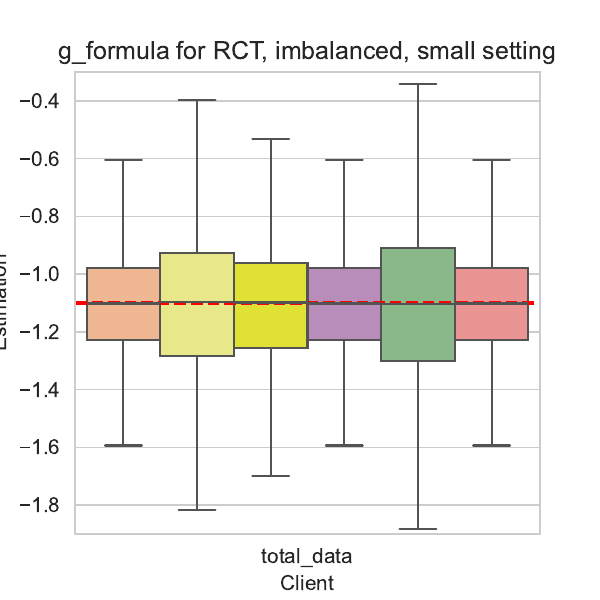}
            \end{minipage} &
            \begin{minipage}[t]{.13\textwidth}
                \centering
                \includegraphics[width=\linewidth]{graphics/legend_vertical_aistats.pdf}
            \end{minipage}
        \end{tabular}
    \end{minipage}
    \caption{Estimation of the ATE in the imbalanced and homogeneous setting for RCT studies.}\label{fig:homog_imbalanced}
\end{figure}

\cref{fig:homog_imbalanced} shows that in the \textit{Large} case, the partition of the pooled data has no impact on the estimation of the ATE for all estimators as long as \cref{cond:local_large_sample_size} holds. The boxplots are similar to the ones obtained in the balanced case (\cref{fig:homog}), as expected again with the ``One RCT'' scenario. 

However, in the \textit{Small} case, the variances of the Meta estimators and 1S-SW are greater than in the balanced setting due to the local estimates $\hat\tau_2, \dots \hat\tau_5$ being obtained after performing two local OLS regressions on each of their treatment arms on very small datasets.

\subsection{Shift in covariates distributions}\label{subsec:simus_diff_X}
We now consider an \textit{heterogeneous} setting where the data are generated according to the graphical model in Figure~\ref{graph:diff_distribs} under model~(\ref{model:model1}), where individuals follow different distributions according to the study they belong to. We consider the case where the means and covariance matrices $\{(\mu_k, \Sigma_k)\}_k$ are different from one study to another, with values:
\begin{table}[h]\label{tab:params_simu_diff_X}
    \centering
    \begin{tabular}{>{\centering\arraybackslash}m{.35\textwidth} >{\centering\arraybackslash}m{.35\textwidth}}
        \textbf{Means} & \textbf{Covariances}\\
        $
        \begin{aligned}
            \mu_1 &= (\underbrace{1,\dots,1}_{\lfloor d/2 \rfloor},\underbrace{-1,\dots,-1}_{\lfloor d/2 \rfloor})^\top \\
        \end{aligned}
        $ &
        $
        \begin{aligned}
            \Sigma_1 &= I_d + 0.5 - 0.5 I_d \\
        \end{aligned}
        $ \\
        $
        \begin{aligned}
            \mu_2 &= (\underbrace{-1,\dots,-1}_{\lfloor d/2 \rfloor},\underbrace{1,\dots,1}_{\lfloor d/2 \rfloor})^\top \\
        \end{aligned}
        $ &
        $
        \begin{aligned}
            \Sigma_2 &= 20 \cdot \Sigma_1 + 0.5 I_d - 0.5 \\
        \end{aligned}
        $ \\
        $
        \begin{aligned}
            \mu_3 &= (0, \dots, 0)^\top \\
        \end{aligned}
        $ &
        $
        \begin{aligned}
            \Sigma_3 &= 0.02 \cdot \Sigma_1 + 0.7 I_d \\
        \end{aligned}
        $ \\
        $
        \begin{aligned}
            \mu_4 &= (\underbrace{0.5,\dots,0.5}_{\lfloor d/2 \rfloor},\underbrace{-1,\dots,-1}_{\lfloor d/2 \rfloor})^\top \\
        \end{aligned}
        $ &
        $
        \begin{aligned}
            \Sigma_4 &= 1 \cdot \Sigma_1 + 0.5 I_d - 0.15 \\
        \end{aligned}
        $ \\
        $
        \begin{aligned}
            \mu_5 &= (\underbrace{1.2,\dots,1.2}_{\lfloor d/2 \rfloor},\underbrace{0.8,\dots,0.8}_{\lfloor d/2 \rfloor})^\top \\
        \end{aligned}
        $ &
        $
        \begin{aligned}
            \Sigma_5 &= 1.5 \cdot \Sigma_1 + 0.5 I_d - 0.15 \\
        \end{aligned}
        $ \\
    \end{tabular}
\end{table}

Note that in this scenario, the Meta-IVW is not relevant as explained in \cref{sec:hetero}.
In this setting, the targeted ATE is $\tau= \sum_{k=1}^{K} \rho_k \left(c^{(1)} - c^{(0)} + \mu_k^\top \left(\beta^{(1)} - \beta^{(0)}\right)\right) \approx 0.45$.

\begin{figure}[!h]
    \centering
    \begin{minipage}[t]{.9\textwidth}
        \begin{tabular}{>{\centering\arraybackslash}m{6cm}>{\centering\arraybackslash}m{6cm}>{\centering\arraybackslash}m{1cm}}
            \textbf{Large} & \textbf{Small} \\
            \begin{minipage}[t]{.45\textwidth}
                \centering
                \includegraphics[width=\linewidth, trim={0.2cm 1cm 0 1.1cm}, clip]{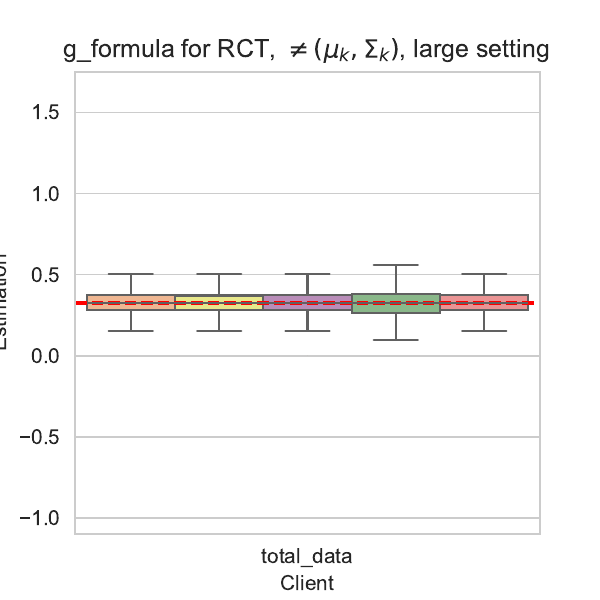}
            \end{minipage} & 
            \begin{minipage}[t]{.45\textwidth}
                \centering
                \includegraphics[width=\linewidth, trim={0.2cm 1cm 0 1.1cm}, clip]{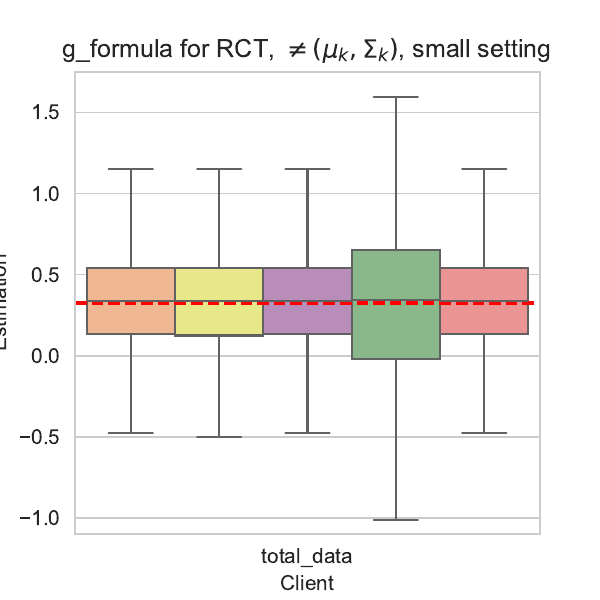}
            \end{minipage} &
            \begin{minipage}[t]{.13\textwidth}
                \centering
                \includegraphics[width=\linewidth]{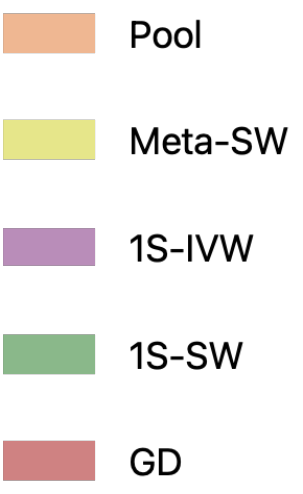}
            \end{minipage}
        \end{tabular}
    \end{minipage}
    \caption{Estimation of the ATE in the heterogeneous setting for RCT studies.}\label{fig:diff_means}
\end{figure}

\cref{fig:diff_means} illustrates the fact that when studies have different covariate means, the variance of the One-Shot SW estimator is enlarged compared to homogeneous settings (\cref{prop:1s_diff_means_var}), both in \textit{Small} and \textit{Large} sample sizes regimes. Notice that the Meta-SW estimator achieves better performance than in the homogeneous setting since its weights $n_k/n$ are good estimates of $\rho_k$ under Condition~\ref{cond:local_large_sample_size} of the true weights of the local ATEs, as $\tau=\sum_{k=1}^K \rho_k \tau_k$.

\subsection{Study-effects}\label{subsec:simus_diff_hk}
We now generate data according to model~(\ref{model:model2}) under the graphical model in Figure~\ref{graph:diff_intercepts}, with study-effects equal to $(h_1, h_2, h_3, h_4, h_5) = (1, .2, -1, 30, 2)$, and with the other parameters set to their default values (\cref{subsec:simulation_parameters}).

\subsubsection{No adjustment}

\begin{figure}[H]
    \centering
    \begin{minipage}[t]{.9\textwidth}
        \begin{tabular}{>{\centering\arraybackslash}m{6cm}>{\centering\arraybackslash}m{6cm}>{\centering\arraybackslash}m{1cm}}
            \textbf{Large} & \textbf{Small} \\
            \begin{minipage}[t]{.45\textwidth}
                \centering
                \includegraphics[width=\linewidth, trim={0.2cm 1cm 0 1.1cm}, clip]{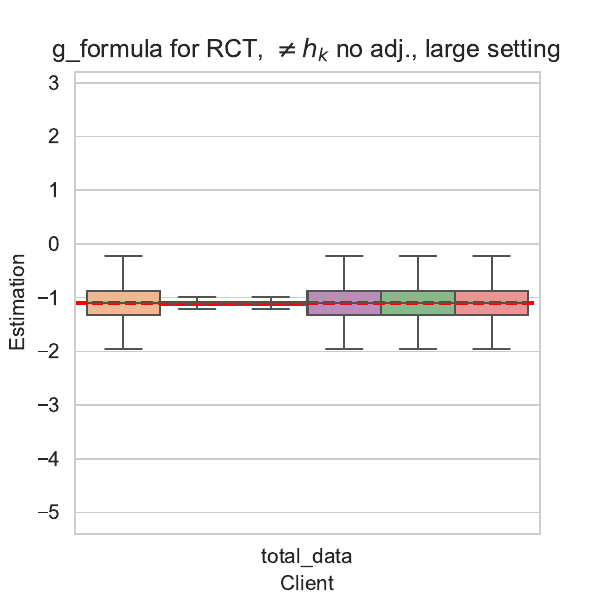}
            \end{minipage} & 
            \begin{minipage}[t]{.45\textwidth}
                \centering
                \includegraphics[width=\linewidth, trim={0.2cm 1cm 0 1.1cm}, clip]{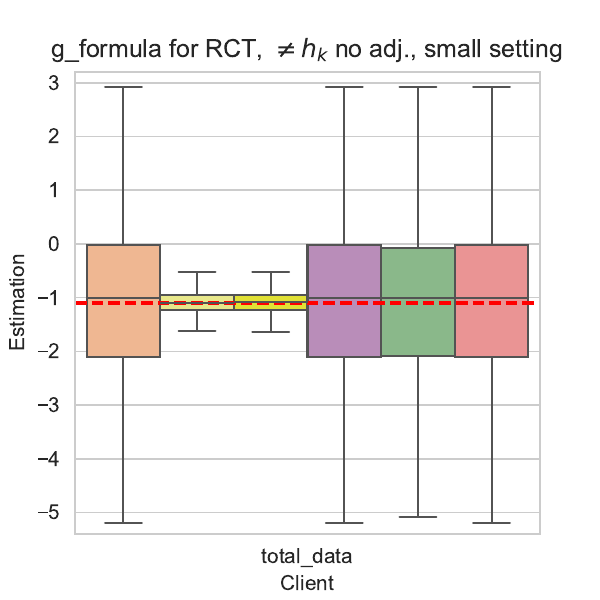}
            \end{minipage} &
            \begin{minipage}[t]{.13\textwidth}
        \centering
        \includegraphics[width=\linewidth]{graphics/legend_vertical_aistats.pdf}
    \end{minipage}
        \end{tabular}
    \end{minipage}
    \caption{Estimation of the ATE in a homogeneous balanced setting with study-effects for RCTs.}\label{fig:diff_intercepts_no_adj}
\end{figure}

\begin{figure}[H]
    \centering
    \begin{minipage}[t]{.9\textwidth}
        \begin{tabular}{>{\centering\arraybackslash}m{6cm}>{\centering\arraybackslash}m{6cm}>{\centering\arraybackslash}m{1cm}}
            \textbf{Large} & \textbf{Small} \\
            \begin{minipage}[t]{.45\textwidth}
                \centering
                \includegraphics[width=\linewidth, trim={0.2cm 1cm 0 1.1cm}, clip]{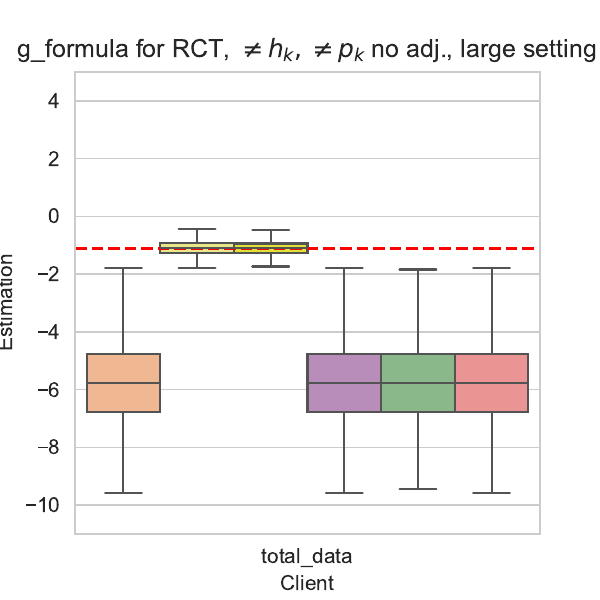}
            \end{minipage} & 
            \begin{minipage}[t]{.45\textwidth}
                \centering
                \includegraphics[width=\linewidth, trim={0.2cm 1cm 0 1.1cm}, clip]{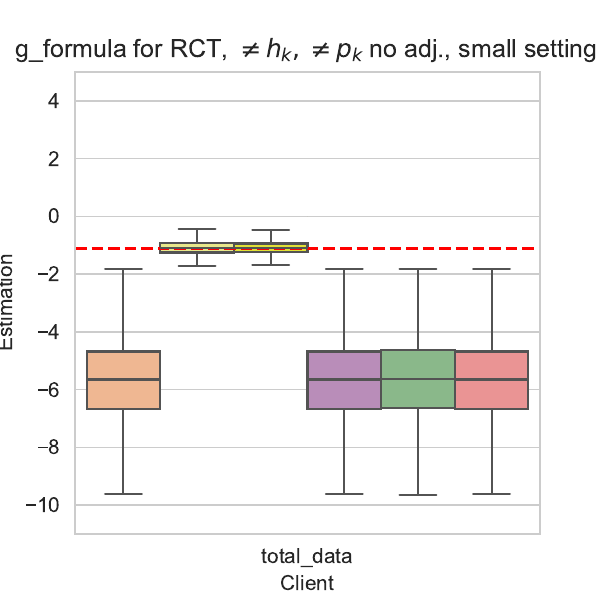}
            \end{minipage}&
    \begin{minipage}[t]{.13\textwidth}
        \centering
        \includegraphics[width=\linewidth]{graphics/legend_vertical_aistats.pdf}
    \end{minipage}
        \end{tabular}
    \end{minipage}
    \caption{Estimation of the ATE in a homogeneous setting with study-effects and different treatment allocation schemes for RCTs.}\label{fig:diff_intercepts_diff_pk_no_adj}
\end{figure}

Figures \ref{fig:diff_intercepts_no_adj} and \ref{fig:diff_intercepts_diff_pk_no_adj} present the distribution of all estimators without any modification, respectively in the ``One RCT'' scenario, and in the different treatment probabilities $p_k$ setting. The estimators are not adjusted to take into account the presence of study-effects, leading to the presence of a confounder between the treatment and the outcome variables (Assumption \hyperlink{as:unconfoundedness}{c}), violating the identifiability assumption.

This case illustrates the advantage of Meta estimators over other estimators, which require in this setting less a priori knowledge on the underlying model in order to be used. 

\subsubsection{Adjusted}
We now consider the same study-effects scenario when adjusting the estimators with the procedures described in \cref{subsec:study_effects_intercepts}: the Pool and GD estimators are computed with access to the membership variables, and the One-Shot estimators do not federate local intercepts. The computation of the Meta estimators remains unchanged.

\begin{figure}[H]
    \centering
    \begin{minipage}[t]{.9\textwidth}
        \begin{tabular}{>{\centering\arraybackslash}m{6cm}>{\centering\arraybackslash}m{6cm}>{\centering\arraybackslash}m{1cm}}
            \textbf{Large} & \textbf{Small} \\
            \begin{minipage}[t]{.4\textwidth}
                \centering
                \includegraphics[width=\linewidth, trim={0cm 0cm 0 0}, clip]{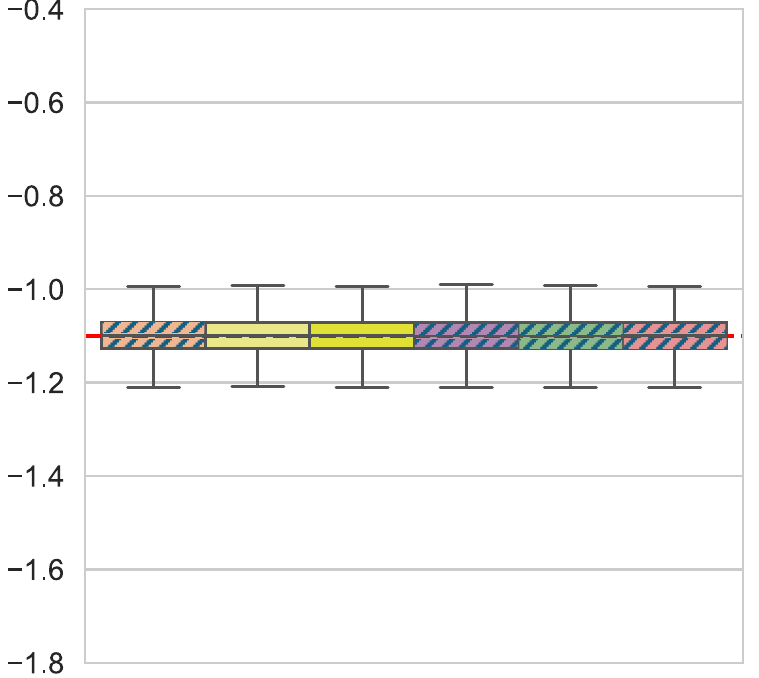}
            \end{minipage} & 
            \begin{minipage}[t]{.4\textwidth}
                \centering
                \includegraphics[width=\linewidth, trim={0cm 0cm 0 0}, clip]{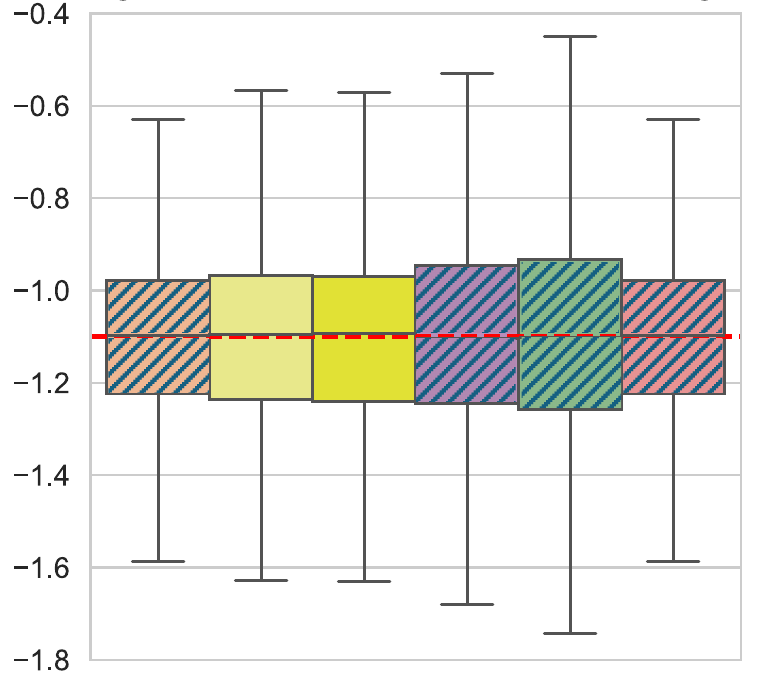}
            \end{minipage} &
    \begin{minipage}[t]{.13\textwidth}
        \centering
        \includegraphics[width=\linewidth]{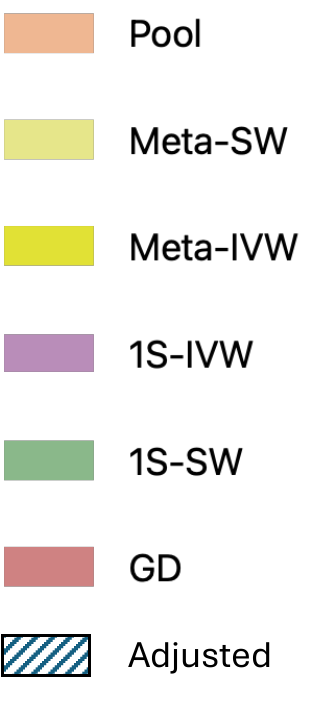}
    \end{minipage}
        \end{tabular}
    \end{minipage}
    \caption{Estimation of the ATE in a homogeneous setting with study-effects for RCTs.}\label{fig:diff_intercepts_adjusted}
\end{figure}

\cref{fig:diff_intercepts_adjusted} shows that all the estimators are now unbiased, and their variance are equal to the pooled data one in the \textit{Large} sample size regime.  

However, in the \textit{Small} regime, the adjusted One-Shot IVW and SW are less efficient than others, as they suffer from the unshared intercepts' variances. The adjusted One-Shot IVW does not benefit from \cref{prop:1S_ivw_thetas_equal_pool}. Their variances are highly dependent on the partition of the data into $K$ splits, unlike the Pool and GD estimators whose variances solely depend on the total amount of data. The meta estimators naturally handle the study-effects and their variances do not suffer much from their magnitude. 

This scenario still highlights the advantage of the Meta estimators, which do not require a specific modelling of the study-effects. However, in this setting they have a higher variance than the (adjusted) GD and (adjusted) pool estimators, as in the simple homogeneous balanced setting (\ref{par:homog}).

\subsection{Full heterogeneity: shifts in covariates distributions, study-effects and in treatment allocation}\label{par:full_heterogeneity}
We now combine the different scenarios of \cref{subsec:simus_diff_X}, \cref{subsec:simus_diff_hk} with study-effects (specific intercept per study), distribution of the covariates that are different from one study to the other (different means and covariance matrices) and different probabilities of being treated by study, corresponding to the graphical model in Figure~\ref{graph:full_hetero}. The simulation parameters used in Figures~\ref{fig:simu_full_hetero} are displayed in \cref{tab:params_simu_full_hetero}:
\begin{table}[!h]\label{tab:params_full_hetero}
    \centering
    \begin{tabular}{>{\centering\arraybackslash}m{.3\textwidth} >{\centering\arraybackslash}m{.25\textwidth}>{\centering\arraybackslash}m{.2\textwidth}>{\centering\arraybackslash}m{.15\textwidth}}
        \textbf{Means} & \textbf{Covariances} & \textbf{Study-Effects} & \textbf{Treatment probabilities}\\
        $
        \begin{aligned}
            \mu_1 &= (\underbrace{1,\dots,1}_{\lfloor d/2 \rfloor},\underbrace{-1,\dots,-1}_{\lfloor d/2 \rfloor})^\top \\
        \end{aligned}
        $ &
        $
        \begin{aligned}
            \Sigma_1 &= I_d + 0.5 - \frac{1}{2}I_d \\
        \end{aligned}
        $ &
        $h_1 = 1 $ &
        $
        \begin{aligned}
            p_1 &= 0.75 
        \end{aligned}
        $ \\
        $
        \begin{aligned}
            \mu_2 &= (\underbrace{-1,\dots,-1}_{\lfloor d/2 \rfloor},\underbrace{1,\dots,1}_{\lfloor d/2 \rfloor})^\top \\
        \end{aligned}
        $ &
        $
        \begin{aligned}
            \Sigma_2 &= 20 \cdot \Sigma_1 + 0.5 I_d - 0.5 \\
        \end{aligned}
        $ &
        $h_2 = 0.2 $ &
        $
        \begin{aligned}
            p_2 &= 0.75 
        \end{aligned}
        $ \\
        $
        \begin{aligned}
            \mu_3 &= (0, \dots, 0)^\top \\
        \end{aligned}
        $ &
        $
        \begin{aligned}
            \Sigma_3 &= 0.02 \cdot \Sigma_1 + 0.7 I_d \\
        \end{aligned}
        $ &
        $h_3 = -1 $ &
        $
        \begin{aligned}
            p_3 &= 0.75 
        \end{aligned}
        $ \\
        $
        \begin{aligned}
            \mu_4 &= (\underbrace{0.5,\dots,0.5}_{\lfloor d/2 \rfloor},\underbrace{-1,\dots,-1}_{\lfloor d/2 \rfloor})^\top \\
        \end{aligned}
        $ &
        $
        \begin{aligned}
            \Sigma_4 &= 1 \cdot \Sigma_1 + 0.5 I_d - 0.15 \\
        \end{aligned}
        $ &
        $h_4 = 30 $ &
        $
        \begin{aligned}
            p_4 &= 0.25 
        \end{aligned}
        $ \\
        $
        \begin{aligned}
            \mu_5 &= (\underbrace{1.2,\dots,1.2}_{\lfloor d/2 \rfloor},\underbrace{0.8,\dots,0.8}_{\lfloor d/2 \rfloor})^\top \\
        \end{aligned}
        $ &
        $
        \begin{aligned}
            \Sigma_5 &= 1.5 \cdot \Sigma_1 + 0.5 I_d - 0.15 \\
        \end{aligned}
        $ &
        $h_5 = 2 $ &
        $
        \begin{aligned}
            p_4 &= 0.25 
        \end{aligned}
        $ \\
    \end{tabular}
\end{table}

\subsection{G-Formula OLS covariate adjustment in non-linearity}\label{simu:dm_vs_gf_var_reduction}
In this setting, we simulate one dataset where $Y_i(1) = 0 x_1^2 + \frac{-1}{2} x_2^2 + \frac{1}{2} x_3^2 + \frac{3}{2} x_4^2 + x_3 * x_4$ and $Y_i(0)= -0.35 x_1^2 + 0 x_2^2 + \frac{1}{2} x_3^2 + \frac{3}{2} x_4^2 + x_1 * x_2$. 

\begin{table}[!h]\label{tab:params_simu_full_hetero}
    \centering
    \begin{minipage}[t]{.45\textwidth}
        \centering
            \includegraphics[width=\linewidth, trim={.3cm .5cm 1cm 1.5cm}, clip]{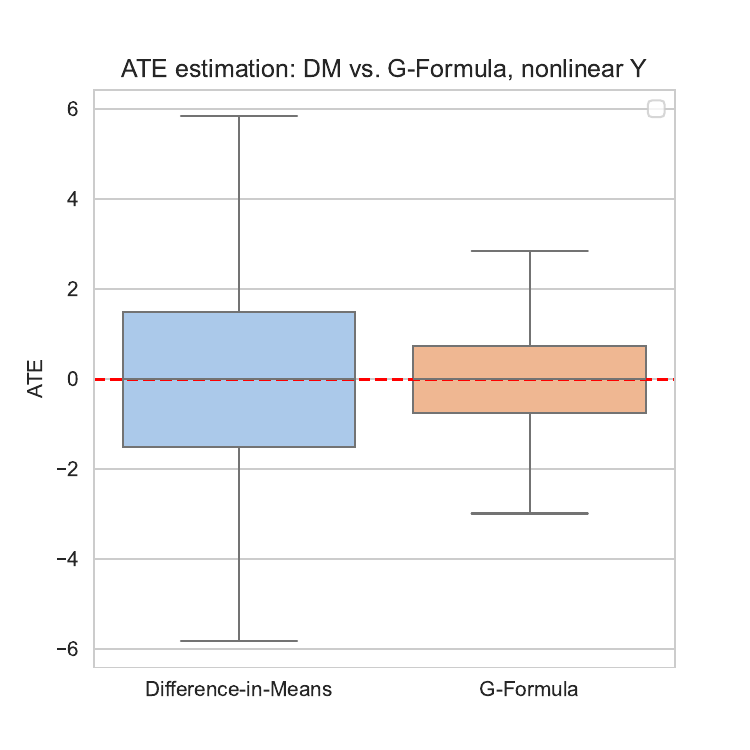}
    \end{minipage}
    \caption{ATE Estimation: DM vs. G-Formula with OLS adjustment on covariates under non-linear outcome modeling}
\end{table}

The pooled-data OLS G-Formula has reduced variance compared to the simple Difference-in-Means, even if the outcome model is non-linear, illustrating \citep{FDA2023, EMA2024, tsiatis2008covariate, benkeser2021improving,lin2013agnostic, wager2020stats, lei2021regression, van2024covariate}.

\section{Algorithms}\label{alg:FedAvg}
We describe in Algorithm~\ref{alg:fedavg} the procedure to obtain the Gradient Descent estimator of outcome model parameters $\hat\theta_\mathrm{GD}$:
\begin{algorithm}[h]
    \caption{Federated Averaging (FedAvg) algorithm to learn $\hat\theta_\mathrm{GD}^{(w)}$}
    \begin{algorithmic}[1]
    \State \textbf{Input:} $K$ studies, $E$ local steps, $B$ batch size, $\eta$ learning rate, $T$ rounds of communication
    \State \textbf{Server executes:}
    \State \quad Initialize $\theta_0^{(w)}$
    \For{each round $t = 0, 1, \dots T$} 
    \For{each study $k \in \llbracket1,K\rrbracket$ \textbf{in parallel}} 
    \State $\theta_{t+1}^{k^{(w)}} \gets \text{LocalUpdate}(k, \theta^{(w)}_t)$
    \EndFor
    \State $\theta^{(w)}_{t+1} \gets \sum_{k=1}^K \frac{n_k^{(w)}}{n^{(w)}} \theta_{t+1}^{k^{(w)}}$\hfill \texttt{// Federated Averaging}
    \EndFor
    \State \textbf{LocalUpdate}($k$, $\theta_k^{(w)}$):
    \For{each local step $e = 0, 1, \dots, E-1$}
    \State $\mathcal{B}_k \gets$ a random batch of $B$ samples from $\mathcal{Z}_k$
    \State $\nabla \ell(\theta_k^{(w)}, \mathcal{B}_k) \gets -\frac{2}{B} X_{\mathcal{B}_k}^\top (Y_{\mathcal{B}_k} - X_{\mathcal{B}_k}\ \theta_k^{(w)}) $\hfill \texttt{// Compute gradient on the mini-batch}
    \State $\theta_k^{(w)} \gets \theta_k^{(w)} - \eta \nabla \ell(\theta_k^{(w)}, \mathcal{B}_k)$\hfill \texttt{// Gradient descent step}
    \EndFor
    \State \textbf{return} $\hat\theta_\mathrm{GD}^{(w)}\leftarrow \theta_k^{(w)}$
    \end{algorithmic}\label{alg:fedavg}
\end{algorithm}

\label{app:choice_of_learning_rate}
We provide some details on the choices of learning rates discussed in \cref{par:eta_gd}:
\begin{itemize}
\item  In the setting where $T=1$ and $E\to\infty$, each study can choose a local learning rate $\eta_k\leq\frac{2}{L_k}$, where $L_k$ is the smoothness constant of the local problem. This quantity is equal to the largest eigenvalue of the covariance matrix of study $k$, $(X_k^{(w)})^\top X_k^{(w)}$, so choosing a learning rate $\eta_k=\frac{2}{\lambda_{\max,k}}$ is a (conservative) choice which ensures the convergence to the local solution $\hat\theta_k^{(w)}$.

\item When performing one local step per round ($E=1$), one can ensure that $\hat\theta_\mathrm{GD}^{(w)}\to\hat\theta_\mathrm{pool}^{(w)}$ as $T\rightarrow\infty$ by choosing a learning rate $\eta < \frac{2}{L}$, where $L$ is the smoothness constant of the global problem. Here, $L$ corresponds to the highest eigenvalue $\lambda_\mathrm{max}$ of the pooled data $(X^{(w)})^\top X^{(w)}$. Its computation in the federated setting is not straightforward (although there exists some methods, e.g., based on distributed power method). A simple alternative that requires a single round of communication consists in noticing that $\lambda_\mathrm{max} \leq \sum_{k=1}^K \lambda_{\mathrm{max},k}$, where $\lambda_{\mathrm{max},k}$ is the highest eigenvalue of study $k$, to set the learning rate smaller than $\frac{2}{\sum_{k=1}^K \lambda_{\mathrm{max},k}}$. In the homogeneous setting, since all sites have equal covariance matrices, $\lambda_\mathrm{max}$ can be approximated by a mere averaging of the local estimates of the highest eigenvalues, i.e. $\hat\lambda_\mathrm{max} = \sum_{k=1}^K \frac{n_k}{n} \hat \lambda_{\mathrm{max},k}$ where $\hat \lambda_{\mathrm{max},k}$
 is learned with any eigenvalues estimation method locally. \end{itemize}

For a choice of a learning rate as discussed above, setting:\label{app:choice_of_T}
\begin{equation*}
    T=\tilde \Omega\left(\frac{\lambda_\mathrm{max}E}{\lambda_\mathrm{min}}\log(1/\varepsilon)+ \frac{\sqrt{\lambda_\mathrm{max}}\zeta E}{\lambda_\mathrm{min}\sqrt{\varepsilon}} \right)\,,
\end{equation*}
gives $\|\theta_T^{(w)}-\hat\theta_\mathrm{pool}^{(w)}\|^2\leq \varepsilon$, where $\theta_T^{(w)}$ is the output of FedAvg after $T$ communication rounds, $\lambda_\mathrm{max},\lambda_\mathrm{min}$ are respectively the maximum and minimum eigenvalues of ${X_\mathrm{pool}^{(w)}}^\top X_\mathrm{pool}^{(w)}$ ($\lambda_\mathrm{min}>0$ under \cref{cond:federated_large_sample_size}), and $\zeta^2=\frac{1}{K}\sum_{k=1}^K\left\|\frac{1}{n_k}\sum_{i=1}^{n_k}X_{k,i}({X_{k,i}^{(w)}}^\top \hat\theta_\mathrm{pool}^{(w)}-Y_{k,i}^{(w)})\right\|^2$, and the $\tilde \Omega$ notation hides constants and polylogarithmic factors~\citep{pmlr-v119-koloskova20a}.

Applying these hyperparameters choices to the simulation setting described in \cref{sec:app_simulations}, we get in the Homogeneous scenario (\ref{par:homog}) $\hat\lambda_\mathrm{max}^{(w)}=\sum_{k=1}^K \frac{n_k}{n}\hat\lambda_\mathrm{max,k}^{(w)} \approx 10.5$ for both treated and control groups, which yields a choice of learning rate $\displaystyle  \eta<\min_{w\in\{0,1\}}(2/\hat\lambda_\mathrm{max}^{(w)}) \approx 0.19$. We chose to divide the upper bound by 10 to ensure stability of the algorithm, which is a common practice, and thus set $\eta$ to $10^{-2}$.

In the simulation, we most often used the following values of parameters for the full batch Gradient Descent algorithm: $T$ ranging from 1000 to 4000, with increased values for heterogeneous settings, $E=1, B=n, \eta=10^{-2}$, and increased $T$ for heterogeneous settings.

\end{appendices}

\end{document}